\documentclass[a4paper,fleqn]{cas-dc}

\usepackage[numbers]{natbib}

\usepackage{graphicx}
\usepackage{array}
\usepackage{xcolor}
\usepackage{tabularx} 
\usepackage{ragged2e}
\usepackage{float}

\colorlet{red}{black}

\definecolor{organColor}{RGB}{176, 154, 255}     % purple
\definecolor{procedureColor}{RGB}{255, 216, 154} % orange
\definecolor{correctColor}{RGB}{154, 252, 255}   % blue
\definecolor{incorrectColor}{RGB}{255, 154, 201} % red
\definecolor{IndeterminateColor}{RGB}{201, 255, 154} % green

\newcommand{\hl}[2]{\colorbox{#1}{#2}}
\newcommand{\tab}{\hspace*{3mm}} 
\newcolumntype{Y}{>{\RaggedRight\arraybackslash}X}

\def\tsc#1{\csdef{#1}{\textsc{\lowercase{#1}}\xspace}}
\tsc{WGM}
\tsc{QE}
\begin{document}
\let\WriteBookmarks\relax
\def\floatpagepagefraction{1}
\def\textpagefraction{.001}

\shorttitle{REG2025}    

\shortauthors{Yumi Lee, Harim Oh et al.}  

\title [mode = title]{Benchmarking Vision-Language Models for Automated Pathology Diagnosis and Report Generation}  

\tnotemark[1] 

\tnotetext[1]{Dataset is available at \href{https://reg2025.grand-challenge.org/reg2025/}{https://reg2025.grand-challenge.org/reg2025/}}
% \tnotetext[1]{Code are available at \href{https://github.com/hrb0/reg/}{https://github.com/hrb0/reg/}} 


\author[1]{Yumi Lee}
% [orcid=0009-0001-7552-0831]%[<options>]   

\fnmark[1]

% % Email id of the first author
% \ead{eumi@ewha.ac.kr}

% % URL of the first author
% \ead[url]{www.linkedin.com/in/eumi}

\credit{Data curation, Formal analysis, Investigation, Project administration, Writing - original draft, Writing - review \& editing}

\author[2]{Harim Oh}
% [orcid=0000-0003-4904-4015]   

\fnmark[1]

% % Email id of the first author
% \ead{harim1028@gmail.com}


\credit{Resources, Data curation, Investigation, Writing - original draft}

\author[1]{Hyoryung Kim}    
\credit{Data curation, Writing - review}
\author[1]{Minji Kim}   
\credit{Data curation, Writing - review}
\author[3]{Eunsu Kim}   
\credit{Data curation}
\author[3]{Hyeseong Lee}   
\credit{Data curation}

%%% Data Offering Institutes  
\author[5,6]{Junya Fukuoka}%[<options>]   
\credit{Resources}
\author[5]{Andrey Bychkov}%[<options>]   
\credit{Resources}
\author[5]{Jijgee Munkhdelger}%[<options>]   
\credit{Resources}
\author[7]{Rajiv Kumar Kaushal}%[<options>]   
\credit{Resources}
\author[7]{Ayushi Sahay}%[<options>]   
\credit{Resources}
\author[8]{Rajni Yadav}%[<options>]   
\credit{Resources}
\author[8]{Bharathi Prabakaran}%[<options>]   
\credit{Resources}
\author[4]{Sulen Sarioglu}%[<options>]    
\credit{Resources}
\author[4]{Serdar Balcı}%[<options>]   
\credit{Resources}
\author[4]{Ilknur Turkmen}%[<options>]   
\credit{Resources}
\author[9]{Yuri Tolkach}%[<options>]   
\credit{Resources}
\author[9]{Christian Harder}%[<options>]   
\credit{Resources}
\author[9]{Julian Westerdorf}%[<options>]   
\credit{Resources}
\author[9]{Reinhard Buettner}%[<options>]   
\credit{Resources}

%%% ICGI
\author[10]{Audun Ljone Henriksen}%[<options>]   
\credit{Investigation}
\author[10]{Sepp De Raedt}%[<options>]  
\credit{Investigation}

%%% ICL_PathReport
\author[11]{Byung Hyun Lee}%[<options>]  
\credit{Investigation}
\author[11]{Sungjin Lim}%[<options>]  
\credit{Investigation}
\author[11]{Joohoon Lee}%[<options>]  
\credit{Investigation}
\author[11]{Gwanghyun Kim}%[<options>]   
\credit{Investigation}
\author[11]{Se Young Chun} %[<options>]   
\credit{Investigation}

%%% IMAGINE Lab
\author[12]{Suryakant Singh}%[<options>]   
\credit{Investigation}
\author[12]{Saarthak Kapse}%[<options>]  
\credit{Investigation}
\author[12]{Prateek Prasanna}%[<options>]  
\credit{Investigation}

%%% ADCT
\author[13]{Kyung A Kim}%[<options>]   
\credit{Investigation}
\author[14]{Yousun Kang}%[<options>]   
\credit{Investigation}
\author[15]{Sehwan Yoo}%[<options>]   
\credit{Investigation}
\author[16]{Sungman Hong}%[<options>]    
\credit{Investigation}

%%% IUCompPath
\author[17]{Shubham Innani}%[<options>]     
\credit{Investigation}
\author[17]{Michael Feldman}%[<options>]    
\credit{Investigation}
\author[17]{Spyridon Bakas}%[<options>]    
\credit{Investigation}

%%% TeamTiger@REG2025
\author[18]{Ujjwal Baid}%[<options>]   
\credit{Investigation}
\author[19]{Prasad Dutande}%[<options>]   
\credit{Investigation}
\author[19]{Suhas Gajare}%[<options>]   
\credit{Investigation}
\author[18]{Bhakti Baheti}%[<options>]   
\credit{Investigation}

%%% MedInsight-ViseurAI
\author[20]{Serkan Sökmen}%[<options>]   
\credit{Investigation}
\author[20]{Ece Tuğba Cebeci}%[<options>]   
\credit{Investigation}
\author[20]{Ahmet Halıcı}%[<options>]   
\credit{Investigation}
\author[20]{Musa Balcı}%[<options>]   
\credit{Investigation}
\author[20]{Kardelen Peçenek} %[<options>]   
\credit{Investigation}

%%% PathoMozhi
\author[21]{Srividhya Sainath}%[<options>]   
\credit{Investigation}

%%% MTS_REG_2025
\author[22]{Kyongseok Jang}%[<options>]   
\credit{Investigation}
\author[22]{Messi H.J. Lee}%[<options>]   
\credit{Investigation}

%%%  NW-TIA
\author[23]{Noorul Wahab}%[<options>]    
\credit{Investigation}

%%% REG_Path 
\author[24]{Bodong Du}%[<options>]
\credit{Investigation}
\author[25]{Jiaming Zhang}%[<options>]
\credit{Investigation}
\author[24]{Qixiang Zhang}%[<options>]
\credit{Investigation}

\author[1]{Jang-Hwan Choi}[orcid=0000-0001-9273-034X]
\cormark[1]
\ead{choij@ewha.ac.kr}
\credit{Conceptualization, Supervision, Writing - review \& editing}
\author[2,3]{Sangjeong Ahn}[orcid=0000-0003-4338-3235]
\cormark[1]
\ead{vanitas80@korea.ac.kr}
\credit{Conceptualization, Supervision, Writing - review \& editing}

% % Corresponding author indication
% \cormark[1]

% % Footnote of the first author

% % Email id of the first author
% \ead{}

% % URL of the author

% % Credit authorship
% % eg: \credit{Conceptualization of this study, Methodology, Software}

%%%%%%%%%%% Address/affiliation %%%%%%%%%%%
\affiliation[1]{organization={Department of Artificial Intelligence, Ewha Womans University},
            addressline={52 Ewhayeodae-gil, Seodaemun-gu}, 
            city={Seoul},
%          citysep={}, % Uncomment if no comma needed between city and postcode
            postcode={03760}, 
            country={Republic of Korea}}

\affiliation[2]{organization={Department of Pathology, Korea University Anam Hospital},
            addressline={73 Goryeodae-ro, Seongbuk-gu}, 
            city={Seoul},
%          citysep={}, % Uncomment if no comma needed between city and postcode
            postcode={02841}, 
            country={Republic of Korea}}

% NEW
\affiliation[3]{organization={Department of Biomedical Informatics, Korea University College of Medicine},
            addressline={73, Goryeodae-ro, Seongbuk-gu}, 
            city={Seoul},
%          citysep={}, % Uncomment if no comma needed between city and postcode
            postcode={02841}, 
            country={Republic of Korea}}

\affiliation[4]{organization={Department of Pathology, Memorial Health Group},
            addressline={Burhaniye, Nagehan Soka{\u{g}}{\i} No:4/A D:1}, 
            city={{\"{U}}sk{\"{u}}dar},
%          citysep={}, % Uncomment if no comma needed between city and postcode
            postcode={34676}, 
            state={{\.{I}}stanbul},
            country={T{\"{u}}rkiye}}

% Data Offering Institutes
\affiliation[5]{organization={Department of Pathology, Kameda Medical Center},
            addressline={929 Higashicho}, 
            city={Kamogawa},
%          citysep={}, % Uncomment if no comma needed between city and postcode
            postcode={296-0041}, 
            state={Chiba},
            country={Japan}}

\affiliation[6]{organization={Department of Pathology Informatics, Nagasaki University Graduate School of Biomedical Sciences},
            addressline={1-7-1 Sakamoto}, 
            city={Nagasaki},
            % citysep={}, % Uncomment if no comma needed between city and postcode
            postcode={852-8523}, 
            country={Japan}}

\affiliation[7]{organization={Department of Pathology, Tata Memorial Hospital},
            addressline={Homi Bhabha National Institute (HBNI)}, 
            city={Mumbai},
%          citysep={}, % Uncomment if no comma needed between city and postcode
            postcode={400012}, 
            country={India}}

\affiliation[8]{organization={Department of Pathology, All India Institute Of Medical Sciences Delhi},
            addressline={Sri Aurobindo Marg, Ansari Nagar East}, 
            city={New Delhi},
%          citysep={}, % Uncomment if no comma needed between city and postcode
            postcode={110029}, 
            state={Delhi},
            country={India}}

\affiliation[9]{organization={Institute of Pathology, University Hospital Cologne, Medical Faculty, University of Cologne},
            addressline={Kerpener Str. 62}, 
            city={K{\"{o}}ln},
%          citysep={}, % Uncomment if no comma needed between city and postcode
            postcode={50937}, 
            country={Germany}}

% ICGI
\affiliation[10]{organization={Institute for Cancer Genetics and Informatics},
            addressline={Ullernchausseen 70}, 
            city={Oslo},
%          citysep={}, % Uncomment if no comma needed between city and postcode
            postcode={0372}, 
            country={Norway}}

% ICL_PathReport
\affiliation[11]{organization={Seoul National University},
            addressline={1 Gwanak-ro, Gwanak-gu}, 
            city={Seoul},
%          citysep={}, % Uncomment if no comma needed between city and postcode
            postcode={08826}, 
            country={Republic of Korea}}

% IMAGINE Lab
\affiliation[12]{organization={Stony Brook University},
            addressline={100 Nicolls Rd}, 
            city={Stony Brook},
%          citysep={}, % Uncomment if no comma needed between city and postcode
            postcode={NY 11794}, 
            state={New York},
            country={United States}}

% ADCT
\affiliation[13]{organization={Department of Pathology, Yonsei University College of Medicine},
            % addressline={50-1 Yonsei-ro, Seodaemun-gu}, 
            city={Seoul},
%          citysep={}, % Uncomment if no comma needed between city and postcode
            % postcode={03722}, 
            country={Republic of Korea}}

\affiliation[14]{organization={Tokyo Polytechnic University},
            addressline={2 Chome-9-5 Honcho}, 
            city={Nakano City},
            % citysep={}, % Uncomment if no comma needed between city and postcode
            postcode={Tokyo 164-8678}, 
            country={Japan}}

\affiliation[15]{organization={Nanyang Technological University},
            addressline={50 Nanyang Ave}, 
            % city={},
%          citysep={}, % Uncomment if no comma needed between city and postcode
            postcode={639798}, 
            country={Singapore}}

\affiliation[16]{organization={Graduate School of Software and Artificial Intelligence Convergence, Korea University},
            addressline={145 Anam-ro, Seongbuk-gu}, 
            city={Seoul},
%          citysep={}, % Uncomment if no comma needed between city and postcode
            postcode={02841}, 
            country={Republic of Korea}}

% \affiliation[17]{organization={LG Electronics Inc.},
%             addressline={10, Magokjungnang 10-ro , Gangseo-gu}, 
%             city={Seoul},
% %          citysep={}, % Uncomment if no comma needed between city and postcode
%             postcode={07796}, 
%             % state={},
%             country={Republic of Korea}}
            
% IUCompPath
\affiliation[17]{organization={Division of Computational Pathology, Department of Pathology and Laboratory Medicine, Indiana University School of Medicine},
            addressline={410 W 10th St}, 
            city={Indianapolis},
%          citysep={}, % Uncomment if no comma needed between city and postcode
            postcode={IN, 46202}, 
            state={Indiana},
            country={United States}}

% TeamTiger@REG2025
\affiliation[18]{organization={Emory University},
            addressline={201 Dowman Dr}, 
            city={Atlanta},
%          citysep={}, % Uncomment if no comma needed between city and postcode
            postcode={GA 30322}, 
            state={Georgia},
            country={United States}}

\affiliation[19]{organization={Shri Guru Gobind Singhji Institute of Engineering and Technology},
            addressline={Guru Tegh Bahadurji Marg, Vishnupuri}, 
            city={Nanded},
%          citysep={}, % Uncomment if no comma needed between city and postcode
            postcode={431606}, 
            state={Maharashtra},
            country={India}}
            
% MedInsight-ViseurAI
\affiliation[20]{organization={Viseur AI},
            addressline={Söğütözü, 9 Eylül Cad No: 4 İç Kapı No:1}, 
            city={Çankaya},
%          citysep={}, % Uncomment if no comma needed between city and postcode
            postcode={06510}, 
            state={Ankara},
            country={T{\"{u}}rkiye}}

% PathoMozhi
\affiliation[21]{organization={EKFZ TU Dresden (KatherLab)},
            addressline={Fetscherstraße 74}, 
            city={Dresden},
%          citysep={}, % Uncomment if no comma needed between city and postcode
            postcode={Postfach 151, 01307}, 
            country={Germany}}

% MTS_REG_2025
\affiliation[22]{organization={MTS Company R\&D Team},
            addressline={434 Samsung-ro, Gangnam-gu}, 
            city={Seoul},
%          citysep={}, % Uncomment if no comma needed between city and postcode
            postcode={06178}, 
            country={Republic of Korea}}

% NW-TIA
\affiliation[23]{organization={NW-TIA},
            addressline={University of Warwick}, 
            city={Coventry},
%          citysep={}, % Uncomment if no comma needed between city and postcode
            postcode={CV4 7AL}, 
            country={United Kingdom}}

% REG_Path 
\affiliation[24]{organization={The Hong Kong University of Science and Technology},
            addressline={Clear Water Bay, Kowloon}, 
            country={Hong Kong}}
%          citysep={}, % Uncomment if no comma needed between city and postcode
            % country={}}

\affiliation[25]{organization={Harbin Institute of Technology},
            addressline={HIT Campus of University Town of Shenzhen}, 
            city={Shenzhen},
%          citysep={}, % Uncomment if no comma needed between city and postcode
            postcode={518055}, 
            country={China}}

\cortext[1]{Corresponding author}

\fntext[1]{These authors contributed equally to this work.}


\begin{abstract}
% Does not exceed 250 words \nocite{*}%% Remove this line from your manuscript.
%The rapid advancement of vision-language models (VLMs) has accelerated progress in computational pathology; however, whole-slide image (WSI)-based pathology report generation remains limited by the scarcity of large-scale WSI--report datasets and the complexity of mapping spatially distributed visual patterns to structured clinical text. To address this, we introduce a clinically curated Pan-Asia WSI--report dataset of approximately 10,500 pairs from five institutions and establish the REG 2025 benchmark through a MICCAI challenge for systematic evaluation of multimodal models. We analyze submitted methods, including pretrained VLMs, multiple-instance learning frameworks, and cross-modal Transformers, showing that VLM-based models with effective multimodal fusion achieve superior performance and strong generalization. We identify key limitations, including instability in quantitative attribute estimation (e.g., numeric hallucination) and a tendency toward diagnostic overspecification, with error patterns resembling those of less experienced pathologists. These findings establish REG 2025 as a benchmark for evaluating report generation and vision-language understanding in computational pathology, providing insights for multimodal pathology models.
The rapid advancement of vision-language models (VLMs) has accelerated progress in computational pathology; however, whole-slide image (WSI)-based pathology report generation remains limited by the scarcity of large-scale WSI--report datasets and the complexity of mapping spatially distributed visual patterns to structured clinical text. To address this, we introduce a clinically curated Pan-Asia WSI--report dataset of approximately 10,500 pairs from five institutions and establish the REG 2025 benchmark through a MICCAI challenge for systematic evaluation of multimodal models. We analyze submitted methods spanning pretrained VLMs, multiple-instance learning frameworks, hierarchical expert models, retrieval-augmented generation, and cross-modal Transformers. Rather than indicating that VLM use alone was sufficient for superior performance, the results suggest that top-performing methods benefited from structured report representations, hierarchical diagnostic decomposition, and effective multimodal grounding. We identify key limitations, including instability in quantitative attribute estimation (e.g., numeric hallucination) and a tendency toward diagnostic overspecification, with some errors resembling known diagnostic pitfalls in routine pathology. These findings establish REG 2025 as a benchmark for evaluating WSI-based structured report generation and vision-language understanding in computational pathology, providing insights for the design of clinically grounded multimodal pathology models.
\end{abstract}


\begin{highlights}
\item Introduced the first large-scale, clinically curated Pan-Asia WSI–report dataset for vision–language learning in computational pathology.
\item Established the REG 2025 benchmark through a MICCAI challenge for systematic evaluation of pathology report generation models.
\item Demonstrated strong cross-regional generalization of models trained on Pan-Asia data to independent European cohorts.
\item Provided a comprehensive comparative analysis of preprocessing strategies and multimodal model architectures across 11 submissions.
\item Revealed key limitations and characteristics of pathology report generation, including structured reasoning requirements and quantitative instability.
\end{highlights}

%\nocite{*}

% Keywords

\maketitle

% Main text
%%%%%%%%%%%%%%%%%%%%%%%%%%%%%%%%%%%%%%%%%%%%%%%%%%%%%%%%%%%%%%%%%%%%%%%%%%%%%%%%%%%%1.
\section{Introduction}

Pathology report generation from whole-slide images (WSIs) remains a fundamentally challenging problem in computational pathology, despite recent advances in artificial intelligence~\cite{song2023artificial,hosseini2024computational,lu2021data}. Unlike conventional vision–language tasks, this problem requires translating giga\-pixel-scale images, characterized by spatially distributed and weakly localized visual patterns, into clinically meaningful textual reports that summarize global tissue characteristics and diagnostic conclusions~\cite{liu2026pathflip,wang2026qcagent,wahab2025mpath}. Such reports are inherently abstract, often reflecting integrated clinical reasoning rather than direct visual descriptions~\cite{mossanen2014surgical,steimetz2024forgotten}. This introduces a substantial spatial–semantic gap between image and text modalities, further compounded by weak and indirect alignment, where slide-level supervision provides no explicit correspondence between local regions and textual components~\cite{zhao2024aligning,ding2025multimodal,guo2024histgen}. In addition, the extreme scale and sparsity of WSIs, where diagnostically relevant regions occupy only a small fraction of the image, make reliable feature aggregation particularly difficult~\cite{bilal2023aggregation,wang2026qcagent}. The problem is further complicated by the one-to-many nature of report generation, as multiple clinically valid descriptions may exist for a single case, and by the variability in reporting styles across pathologists and institutions~\cite{kalra2019automatic,bracamonte2016communicating,ng2025understanding,robert2023high}. Together, these characteristics distinguish pathology report generation from conventional multimodal learning problems and highlight the need for specialized modeling strategies and standardized data resources.

To address these challenges, a wide range of artificial intelligence–based approaches have been developed in computational pathology following the digitization of WSIs~\cite{song2023artificial,hosseini2024computational,lu2021data}. Early methods primarily relied on convolutional neural network (CNN)–based patch-level analysis~\cite{cruz2014automatic,hou2016patch,zhou2014classification,litjens20181399,steiner2018impact,chang2015stacked}, which were later extended to multiple-instance learning (MIL) frameworks~\cite{lu2021data,li2021dual,ilse2018attention,shao2021transmil,chen2022scaling,mao2025camil,cai2025attrimil,jang2024molecular} to enable slide-level inference without requiring exhaustive annotations. More recently, foundation models trained on large-scale datasets have been introduced to learn general-purpose representations that can be transferred across downstream tasks~\cite{wang2022transformer,chen2024towards,lu2024visual,xu2024whole,vorontsov2024foundation,ding2025multimodal}. Building upon these advances, recent studies have further explored vision–language models (VLMs) for pathology report generation, aiming to bridge visual and textual modalities in a unified framework~\cite{liang2025wsi,chen2025slidechat,chen2024wsi,liu2025histopathology,tan2024clinical,yuan2026hipath}. Despite these efforts, existing approaches remain limited in effectively addressing the intrinsic challenges of pathology report generation, particularly in terms of weak alignment, abstract semantic reasoning, and the lack of standardized paired datasets.

This discrepancy becomes more evident when compared to the rapid progress observed in radiology report generation~\cite{wang2025survey,sloan2024automated}, where large-scale paired image–report datasets and more explicit global image–text alignment have enabled substantial advances. In contrast, advancements in pathology report generation remain comparatively limited~\cite{guo2024histgen,hu2025pathology}. These limitations are primarily driven by the intrinsic characteristics of WSIs. Unlike radiology images, WSIs are gigapixel in scale and require patch-based processing, which disrupts global spatial coherence~\cite{bilal2023aggregation}. Moreover, pathology reports summarize complex spatial patterns that are not directly localized, leading to weak image--text alignment~\cite{zhao2024aligning,ding2025multimodal,guo2024histgen}. 

In addition, pathology reports are predominantly semi-structured free text~\cite{simpson2015cancer,santos2022automatic}, and in practice, standardized reporting protocols such as CAP synoptic guidelines are not consistently adopted across institutions~\cite{christinat2024reporting,cap2018synoptic}. This issue is particularly pronounced in settings with limited digital pathology infrastructure or incomplete adoption of synoptic reporting systems~\cite{sluijter2016effects,valverde2022establishing}, where diagnostic findings are often documented in narrative form without consistently structured diagnostic fields or explicit final diagnoses. Even when formal guidelines exist, reporting practices frequently vary across pathologists, institutions, and countries in terms of terminology, report structure, and diagnostic granularity~\cite{tripathi2025using}. For example, breast pathology reporting practices vary considerably across institutions, with some centers adopting CAP synoptic templates~\cite{cho2021standardized}, whereas others continue to rely on institution-specific narrative reporting styles. Such heterogeneity introduces significant supervision noise between gigapixel WSIs and text, thereby limiting reliable multimodal learning, weakening image--text alignment, and complicating quantitative evaluation~\cite{li2024generalizable,tran2025generating}.

Publicly available WSI benchmarks, including CAMELYON~\cite{litjens20181399}, PANDA~\cite{bulten2022artificial}, and SurGen~\cite{myles2025surgen}, have substantially advanced computational pathology research in tasks such as classification, segmentation, and survival prediction. However, pathology image--text benchmarks for WSI-based report generation remain extremely limited, with representative examples including TCGA~\cite{weinstein2013cancer}, HISTAI~\cite{nechaev2025histai}, and HANCOCK~\cite{dorrich2025multimodal}. In TCGA, pathology reports are typically provided as manually written PDF documents, requiring extensive preprocessing and parsing to extract usable text~\cite{kefeli2024tcga,chen2024wsicaption,kemper2010pathtext}, while the reports themselves remain highly heterogeneous in structure, terminology, and linguistic style across institutions, weakening the supervision signal between WSIs and text. HISTAI, although designed for large-scale pretraining, primarily focuses on representation learning rather than standardized report generation, and its reports remain largely unstructured despite text cleaning procedures~\cite{liu2026using,lam2022accessible}. HANCOCK~\cite{dorrich2025multimodal} is further constrained by its focus on a specific cancer type, limiting both data diversity and cross-organ generalization. Collectively, these limitations hinder reliable quantitative evaluation and clinically grounded multimodal learning~\cite{li2025survey,Li_2025_CVPR}, highlighting the lack of large-scale, clinically standardized, and multi-institutional WSI--report benchmarks for pathology report generation.

To address these limitations, we introduce a high-quality, clinically curated WSI--report dataset constructed in accordance with the College of American Pathologists (CAP)\footnote{\url{https://www.cap.org/}} guidelines, which define the core elements of complete pathology reporting and provide a synoptic framework for reducing inter-institutional variability~\cite{sluijter2016effects}. The proposed dataset comprises approximately 10,500 paired WSI--report samples collected from five institutions across South Korea, India, Japan, Turkey, and Germany, representing the first large-scale Pan-Asia--centric multi-institutional dataset for pathology report generation, to the best of our knowledge.

\begin{figure*}[t]
	\centering
	\includegraphics[width=\textwidth]{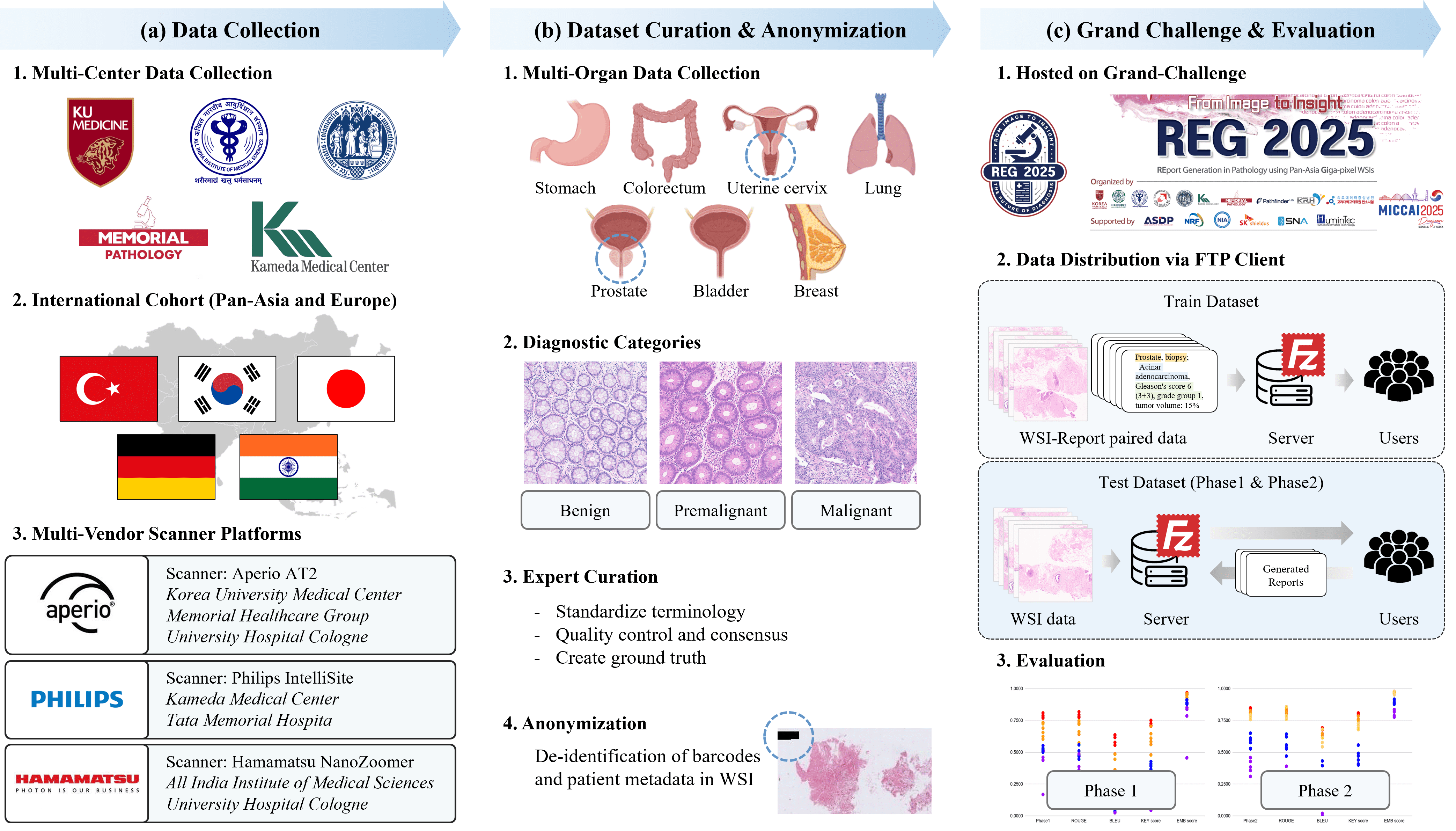}
	\caption{\textbf{REG 2025 challenge workflow.} (a) Multi-institutional WSI–report pairs were collected using three scanner platforms. (b) The data were curated through organ- and diagnosis-level categorization and anonymization. (c) The curated dataset was distributed to participants, and the generated reports were collected and evaluated as part of the challenge.}
	\label{fig:workflow}
\end{figure*}

Building upon this dataset, we further establish a large-scale benchmark through the REport Generation in Pathology using Pan-Asia Giga-pixel WSIs 2025 (REG 2025) Challenge at the Medical Image Computing and Computer Assisted Intervention\footnote{\url{https://conferences.miccai.org/2025/}}
 conference, enabling systematic evaluation of multimodal models under realistic clinical settings. A total of 389 participants from 40 countries took part in the challenge, among which 24 teams successfully submitted their final models. The submitted approaches span diverse paradigms, including pretrained VLM-based methods~\cite{liang2025wsi,chen2025slidechat,chen2024wsi,liu2025histopathology,tan2024clinical,yuan2026hipath}, pathology-specific MIL frameworks~\cite{lu2021data,ilse2018attention,shao2021transmil,chen2022scaling}, and cross-modal Transformer architectures~\cite{wang2022transformer,ding2025multimodal}, reflecting the current landscape of computational pathology.

 Beyond dataset construction and benchmarking, we provide a comprehensive empirical analysis of the REG 2025 Challenge, offering insights into the fundamental characteristics of pathology report generation. Specifically, our analysis reveals that:

\begin{enumerate}
    \item Models leveraging pretrained VLM backbones with effective multimodal fusion consistently achieve superior performance across report generation quality metrics.
    \item Models trained on Pan-Asia data exhibit strong generalization to independent European cohorts, suggesting the potential to mitigate Western-centric bias in existing pathology datasets.
    \item Clinical precision, linguistic coherence, and completeness of pathological reasoning are strongly correlated with overall model performance.
    \item Conventional approaches remain limited in capturing the structured nature of pathology reports, highlighting the importance of structured learning for pathology report generation.
\end{enumerate}

These findings establish REG 2025 as the first benchmark for jointly evaluating report generation and vision\-language understanding in computational pathology, while providing practical insights for the development of future multimodal pathology models.

\begin{table*}[t]
\caption{\textbf{Representative diagnostic categories across seven organs.} The pathology dataset included in the challenge were curated and organized according to the following diagnostic categories.}
\label{tbl:diagnosis}
\centering
\small
\begin{tabular*}{0.95\textwidth}{@{\extracolsep{\fill}} @{} m{0.12\textwidth}|m{0.80\textwidth}@{} }
\toprule
\textbf{Organ} & \textbf{Diagnostic Categories} \\
\midrule
Stomach &
Adenocarcinoma; Tubular adenoma (low/high-grade dysplasia); 
Extranodal marginal zone B cell lymphoma (MALT type); 
Chronic (active) gastritis; Others \\
\midrule
Prostate &
Acinar adenocarcinoma; Normal; Others \\
\midrule
Lung &
Adenocarcinoma; Squamous cell carcinoma; Small cell carcinoma; 
Chronic granulomatous inflammation; 
No evidence of malignancy or granuloma; Others \\
\midrule
Colorectum &
Adenocarcinoma; Tubular adenoma (low/high-grade dysplasia); 
Hyperplastic polyp; Serrated lesion; 
Chronic nonspecific inflammation; Others \\
\midrule
Cervix &
Low-grade squamous intraepithelial lesion (LSIL); 
High-grade squamous intraepithelial lesion (HSIL); 
Invasive squamous cell carcinoma; 
Endocervical adenocarcinoma in situ (AIS), HPV-associated; 
Chronic nonspecific cervicitis; Others \\
\midrule
Bladder &
Invasive urothelial carcinoma; 
Non-invasive papillary urothelial carcinoma; 
Urothelial carcinoma in situ; 
No tumor present \\
\midrule
Breast &
Invasive ductal carcinoma; Invasive lobular carcinoma; 
Ductal carcinoma in situ (DCIS); 
Fibroepithelial tumor; Papillary neoplasm; Others \\
\bottomrule
\end{tabular*}
\end{table*}

\begin{table}[t]
\caption{\textbf{Number of cases by organ across training, Phase 1, and Phase 2 datasets.} Each dataset was stratified to maintain a consistent distribution across organs and institutions.}
\label{tbl:data_distribution}
\centering
\begin{tabular}{l|ccc}
\toprule
\textbf{Organs} & \multicolumn{3}{c}{\textbf{Number of cases by organ}} \\ 
\cmidrule(lr){2-4}
 & \textbf{Train} & \textbf{Phase 1} & \textbf{Phase 2} \\
\midrule
Breast & 1916 & 112 & 112 \\
Cervix & 621 & 37 & 37 \\
% Rectum & 211 & 14 & 16 \\
% Colon & 733 & 314 & 282 \\
Colorectum & 944 & 328 & 298 \\
Lung & 898 & 50 & 50 \\
Prostate & 1770 & 331 & 361 \\
Stomach & 1477 & 87 & 87 \\
Bladder & 868 & 55 & 55 \\
\midrule
Total & 8494 & 1000 & 1000 \\
\bottomrule
\end{tabular}
\end{table}

%%%%%%%%%%%%%%%%%%%%%%%%%%%%%%%%%%%%%%%%%%%%%%%%%%%%%%%%%%%%%%%%%%%%%%%%%%%%%%%%%%%%2.
\section{Materials and Methods}
% \subsection{Challenge Organization}
% ...

\subsection{Dataset Description and Preprocessing Pipeline}
WSIs and their corresponding pathology reports were collected from five institutions: Korea University Anam Hospital (Korea), Kameda Medical Center (Japan), All India Institute of Medical Sciences (India), Memorial Healthcare Group Istanbul (T{\"{u}}rkiye) and University Hospital Cologne (Germany) (\textcolor{blue}{Fig.~\ref{fig:workflow}}-(a)). WSIs were digitized using institution-specific scanners: Aperio AT2 at Korea University Medical Center and Memorial Healthcare Group; Philips IntelliSite Pathology Solution at Kameda Medical Center; and Hamamatsu NanoZoomer at All India Institute of Medical Sciences and University Hospital Cologne. 

The dataset spans seven organs-breast, bladder, cervix, colorectum, lung, prostate, and stomach- and each WSI was paired with its corresponding pathology report (\textcolor{blue}{Fig.~\ref{fig:workflow}}-(b)). As the original reports differed substantially across institutions in structure, terminology,  level of descriptive detail, and language, we generated standardized ground-truth pathology reports using a unified template with consistent structure and terminology (\textcolor{blue}{Fig.~\ref{fig:appendix_data_workflow}}-(a)). Initial screening was performed by four pathology trainees and subsequently reviewed by three board-certified pathologists. Each curated ground-truth report included the organ, procedure, histologic type and, when applicable, histologic grade, while excluding information not directly observable from pathological images, such as anatomical location or diagnoses based on clinical context.

Histologic terminology was standardized according to the 5th edition of the World Health Organization (WHO) Classification of Tumours; for example, the term "invasive ductal carcinoma" was redesignated as "invasive carcinoma of no special type". For malignant tumor cases, selected organ-specific elements were additionally included in the pathology report with reference to the corresponding CAP Cancer Protocol. These included parameters such as the Nottingham histologic score for breast carcinoma and Gleason grading for prostate needle biopsy specimens. The CAP Cancer Protocol versions used in this study were invasive carcinoma of breast v1.2.0.0, ductal carcinoma in situ of breast v1.0.1.0, urinary bladder v4.2.0.0 and prostate gland needle biopsy v1.1.0.0 (\textcolor{blue}{Fig.~\ref{fig:appendix_data_workflow}}-(c)). For stomach biopsy specimens, the terminology for low-grade dysplasia differed across institutions and was unified as “tubular adenoma with low-grade dysplasia” through consensus meetings.

Inclusion and exclusion criteria were applied in two stages: technical WSI quality screening followed by expert review of diagnostic certainty. First, cases were excluded on technical grounds if the slide contained insufficient tissue for pathologic assessment, if the lesional focus was too limited for reliable interpretation, or if there was out-of-focus scanning, poor staining, excessive cautery artifact, or other technical deficiencies. Second, cases that were technically acceptable but diagnostically equivocal were excluded after expert review. This latter step was intended to reduce label noise arising from unavoidable interobserver variability in borderline lesions (e.g. stomach biopsy between well-differentiated adenocarcinoma vs. high-grade dysplasia). Together, these steps ensured that the final dataset comprised cases that were both technically adequate and diagnostically reliable, resulting in a curated cohort 15.4\% smaller than the initial collection.

All curated datasets were anonymized and the pyramid level corresponding to  20× magnification was extracted and  exported as TIFF files. During anonymization, associated images (“label”, “macro”, and “thumbnail”) were removed from Aperio-scanned files (\textcolor{blue}{Fig.~\ref{fig:appendix_data_workflow}}-(b)).
For WSIs containing multiple specimens on a single image, each image was manually cropped into separate image regions, ensuring that each cropped slide image corresponded to its respective pathology diagnosis. Anonymized filenames followed the format \texttt{PIT\_\{institution ID\}\_\{slide number\}\_\{split number\}.tiff}, where the split number was used for manually cropped images containing multiple specimens.

In total, the final dataset consists of 10,494 WSI--report pairs, representing a broad spectrum of diagnostic categories across seven organs, including malignant entities (e.g., adenocarcinoma, squamous cell carcinoma), premalignant lesions (e.g., tubular adenoma with low-grade dysplasia), benign conditions, and non-neoplastic tissues (\textcolor{blue}{Table~\ref{tbl:diagnosis}}). This composition reflects the diagnostic spectrum encountered in routine pathology practice and provides a clinically relevant setting for model development.

This study was approved by the Institutional Review Boards (IRB) of Korea University Medical Center (No. K2024-2640-001) and Kameda Medical Center (No. 22-094), the Ethics Committee of Memorial Healthcare Group Istanbul (decision no. 28.03.2025/001), and the Ethics Committee of the University of Cologne (ethics votes 25-1057 and 20-1583). Data transfer and external use were additionally reviewed and approved by the KUMC Data Review Board (No. D2024-074-CR). For datasets provided by the All India Institute of Medical Sciences, separate IRB approval at the contributing institution was not required because the data were fully anonymized and contained no patient-identifiable information. The REG2025 dataset was released under the CC BY-NC-SA 4.0 license for non-commercial research and benchmarking purposes.

\begin{table*}[t]
\caption{\textbf{Overall performance comparison of participating teams in the REG 2025 Challenge.} Phase-specific scores represent the weighted sum of ROUGE, BLEU, KEY, and EMB metrics. Teams are ranked according to the final score (--- denotes no submission for the corresponding phase.).}
\label{tbl:final_results}
\centering
\small
\begin{tabular*}{0.95\textwidth}{@{\extracolsep{\fill}} @{} L|L|LL|L||L|L|LL|L@{} }
\toprule
\textbf{Rank} & \textbf{Team} & \textbf{Phase 1} & \textbf{Phase 2} & \textbf{Final} &
\textbf{Rank} & \textbf{Team} & \textbf{Phase 1} & \textbf{Phase 2} & \textbf{Final}\\
\midrule
1 & ICGI                & \textbf{0.8098} & 0.8472 & \textbf{0.8397} & 13 & TokenStreamers & 0.7248 & 0.5773 & 0.6068 \\
2 & ICL\_PathReport     & 0.7352 & 0.8415 & 0.8202 & 14 & NUST-SEECS     & 0.5144 & 0.6121 & 0.5926 \\
3 & IMAGINE Lab         & 0.6258 & \textbf{0.8494} & 0.8047 & 15 & Clinical\_Omics\_Institute  & 0.5531 & 0.5984 & 0.5894 \\
4 & ADCT                & 0.7704 & 0.8040 & 0.7973 & 16 & JX\_REG        & 0.5073 & 0.5309 & 0.5262 \\
5 & IUCompPath          & 0.7906 & 0.7899 & 0.7900 & 17 & DSG            & 0.4385 & 0.4286 & 0.4306 \\
6 & TrustPath           & 0.6565 & 0.8127 & 0.7815 & 18 & sk             & ---    & 0.5277 & 0.4221 \\
7 & TeamTiger@REG2025   & 0.7777 & 0.7763 & 0.7766 & 19 & virasoft       & 0.1691 & 0.4590 & 0.4010 \\
8 & MedInsight-ViseurAI & 0.5326 & 0.8093 & 0.7539 & 20 & Shubham47      & 0.4402 & 0.3849 & 0.3960 \\
9 & PathoMozhi           & 0.4611 & 0.7960 & 0.7290 & 21 & DiceMed\_labs  & ---    & 0.3577 & 0.2861 \\
10 & MTS\_REG\_2025     & 0.6046 & 0.7584 & 0.7276 & 22 & UTWL           & ---    & 0.3110 & 0.2488 \\
11 & NW-TIA             & ---    & 0.8282 & 0.6626 & 23 & May\_REG       & 0.5245 & ---    & 0.1049 \\
12 & REG\_Path          & 0.6907 & 0.6510 & 0.6589 & 24 & IMI\_REG       & 0.5027 & ---    & 0.1005 \\
\bottomrule
\end{tabular*}
\end{table*}

\subsection{Challenge Design and Workflow}
The challenge was hosted on the Grand Challenge\footnote{\url{https://grand-challenge.org/}} platform (\textcolor{blue}{Fig.~\ref{fig:workflow}}-(c)). Participants were provided with approximately 8,500 preprocessed WSI–report paired samples as the training dataset, and the competition was conducted in two stages: Test Phase 1 and Test Phase 2 (\textcolor{blue}{Table~\ref{tbl:data_distribution}}). The dataset was partitioned at the patient level to prevent data leakage. The training data were distributed through the remote server, and participants were given two months after the dataset release to develop and train their models. In each test phase, participants received 1,000 WSI-only samples, for which they were required to generate corresponding pathology reports within a three-week submission period. The evaluation procedure for the generated reports is described in \textcolor{blue}{Section~\ref{2.3}}.

Each team was allowed two submissions per phase, and all submitted reports were evaluated within a Docker-based environment to ensure consistency and reproducibility. The leaderboard displayed only the weighted Ranking Score--aggregated from the four evaluation metrics--to prevent participants from fine-tuning their models toward specific metrics. The leaderboard was updated in real time via GitHub\footnote{\url{https://github.com/hrb0/reg/tree/main/leaderboard}}, allowing participants to monitor their relative performance throughout the competition.

The primary distinction between Test Phase 1 and Test Phase 2 was the inclusion of European data. Phase 1 consisted of 1,000 Pan-Asia samples collected from the same four hospitals as the training set, whereas Phase 2 included 500 Pan-Asia samples and an additional 500 European samples obtained from a German medical center. This design aimed to evaluate whether the models developed by participants could maintain robust performance across different ethnic and institutional domains. The final ranking score was computed as a weighted sum of the two phases:
\[
\text{Final Score} = (0.2 \times \text{Phase 1 Score}) + (0.8 \times \text{Phase 2 Score})
\]

\subsection{Evaluation Metrics}\label{2.3}
Traditional natural language processing (NLP) metrics were not designed to evaluate clinically structured pathology reports, where subtle differences in wording can alter diagnostic meaning.
To address this limitation, we developed a clinically grounded composite evaluation framework in the REG 2024 Challenge\footnote{\url{https://github.com/PathfinderLab/REG}}, which was conducted as a domestic pilot study with limited scale and served as a preliminary validation of the proposed framework. Rather than relying on a single metric, the evaluation employed a composite score that integrates multiple complementary evaluation metrics designed to reflect the structured and diagnostically relevant properties of pathology reports. 

The ranking score was defined as follows:
\begin{align*}
\text{Ranking score} 
&= 0.15 \times (\text{ROUGE} + \text{BLEU}) + \\
& 0.4 \times \text{KEY score} + 0.3 \times \text{EMB score}
\end{align*}
where the KEY score represents the Jaccard similarity between sets of clinically relevant keywords extracted from reports, and the EMB score measures the cosine similarity between sentence embeddings derived from a biomedical domain-specialized language model ~\cite{OpenBioLLMs}. The detailed implementation is available at \href{https://github.com/hrb0/reg/}{https://github.com/hrb0/reg/}.

As the participating models were trained on curated pathology reports and aimed to reproduce established diagnostic expressions, we evaluated surface-level textual overlap between generated and ground truth reports using ROUGE-L~\cite{lin2004rouge} and BLEU-4~\cite{papineni2002bleu}, two conventional word-level similarity metrics. However, simple lexical overlap alone cannot adequately capture clinical equivalence in pathology reporting. For example, semantically equivalent statements such as \textit{“core-needle biopsy”} and \textit{“sono-guided core biopsy”} convey identical diagnostic conclusions despite differences in wording. 

Moreover, keyword-based matching alone may fail to distinguish clinically critical negations, such as \textit{“the specimen includes muscle proper”} versus \textit{“the specimen does \textbf{not} include muscle proper”}. To address these limitations, the evaluation framework integrates both keyword-level assessment of essential diagnostic elements and a semantic similarity component that captures the overall meaning of the report. In the REG 2024 Challenge, this ranking score demonstrated a high level of concordance with expert-based mean opinion scores. Rankings derived from the composite score aligned most closely with pathologists’ subjective assessments of overall report quality and diagnostic adequacy, supporting its use as a practical and clinically meaningful surrogate metric for large-scale benchmarking of pathology report generation models.

%%%%%%%%%%%%%%%%%%%%%%%%%%%%%%%%%%%%%%%%%%%%%%%%%%%%%%%%%%%%%%%%%%%%%%%%%%%%%%%%%%%%3.

\section{Results}
The challenge attracted a total of 389 participants from 40 countries. In Phase 1, submissions were received from 20 teams, comprising 8 individual and 12 group teams, while Phase 2 included 22 teams, consisting of 9 individual and 13 group teams. \textcolor{red}{The algorithms adopted by the participating teams are discussed in Section~\ref{sec:algorithms}.}

\subsection{Quantitative Evaluation Results}
The ranking scores of all teams for each phase are summarized in \textcolor{blue}{Table~\ref{tbl:final_results}}. In Phase 1, the highest score was achieved by the ICGI team with a score of 0.8098, while the overall average reached 0.5915. In Phase 2, the IMAGINE Lab team obtained the highest score of 0.8494, with an average score of 0.6523, indicating a clear performance improvement compared to Phase 1.
These results suggest that models trained on the Pan-Asia dataset maintained strong performance when evaluated on European cohorts, indicating the potential for cross-regional generalization.

Based on Final Score, the top three teams were ICGI, ICL\_PathReport, and IMAGINE Lab, all achieving final scores above 0.8000, underscoring their outstanding performance.

\begin{figure}
    \centering
    \includegraphics[width=0.9\linewidth]{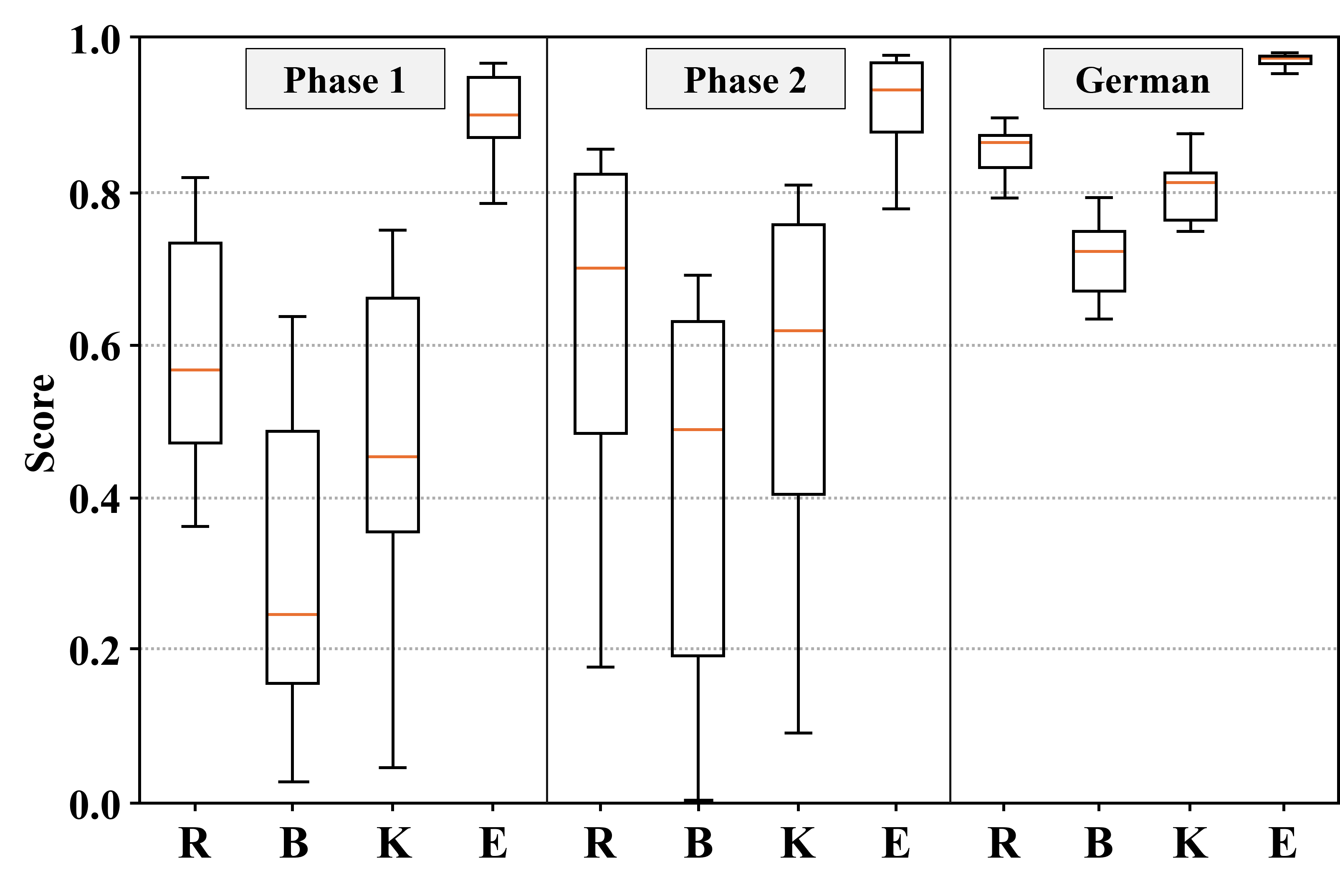}
    \caption{\textbf{Score distribution of participating teams across Phase 1, Phase 2, and the German subset of Phase 2.} Similar score distribution patterns are observed for ROUGE, BLEU, KEY, and EMB metrics in both phases. (R: ROUGE, B: BLEU, K: KEY score, E: EMB score)}
    \label{fig:Score}
\end{figure}

Furthermore, we compared the results according to four evaluation metrics -- ROUGE, BLEU, KEY, and EMB scores -- as presented in \textcolor{blue}{Table~\ref{tbl:phase1_results},~\ref{tbl:phase2_results}}. When the teams were ranked in descending order of their overall Ranking Score, the KEY and EMB metrics exhibited similar trends in both Phase 1, Phase 2, and the German subset of Phase 2 (\textcolor{blue}{Fig.~\ref{fig:Score}}).

Interestingly, while distribution patterns were largely consistent across Phase 1, Phase 2, and the German subset, the German subset showed higher mean performance with lower variance than the Pan-Asia subset. In contrast, the Pan-Asia dataset exhibited lower overall performance and higher inter-team variability, suggesting broader institutional and demographic heterogeneity. 
% These results suggest that, although model rankings are preserved, models trained on relatively homogeneous data may face challenges when applied to more diverse populations, highlighting the role of the Pan-Asia dataset as a more challenging benchmark.
Although relative team rankings were largely preserved across evaluation subsets, these findings indicate that the Pan-Asia dataset may provide a more demanding setting for evaluating model robustness under diverse clinical conditions. Importantly, this interpretation reflects differences in dataset composition and variability rather than inherent differences in population-level difficulty.

\begin{figure*}
    \centering
    \includegraphics[width=1.0\linewidth]{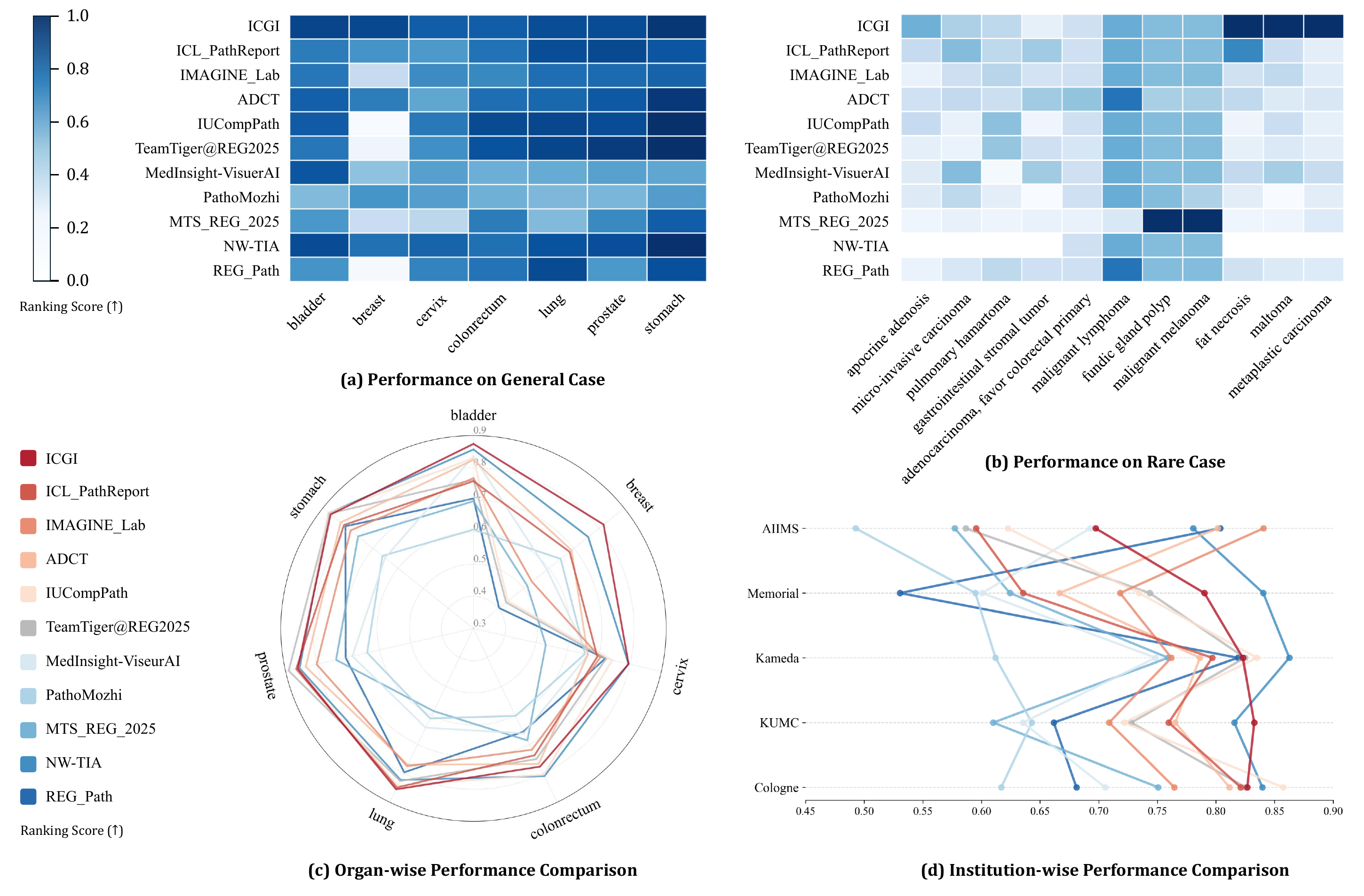}
    \caption{\textbf{Category-wise performance analysis of participating models.} (a) Organ-wise performance comparison for general cases. (b) Performance comparison on 11 rare diagnostic cases. (c) Radar plot illustrating organ-wise performance across all cases. (d) Score distribution across contributing institutions. \textcolor{red}{All performance comparisons are based on the REG Ranking Score (0--1), with higher scores indicating better performance.}}
    \label{fig:quantitative_anlaysis}
\end{figure*}

Moreover, in Phase 2, we observed a sharp decline in BLEU and KEY scores among teams whose Ranking Scores fell below 0.7000. Notable cases include:
\begin{enumerate}[\textbullet]
    \item In Phase 1, the \textit{IMAGINE Lab} team showed an unusually low BLEU score compared to its ROUGE performance.
    \item In Phase 1, the \textit{virasoft} team presented abnormally low EMB scores despite typically high values for most participants.
    \item In Phase 2, the \textit{ICL\_PathReport} and \textit{PathoMozhi} teams achieved similar ROUGE scores but displayed substantial differences in KEY scores.
    \item In Phase 2, the \textit{MedInsight-ViseurAI}, \textit{ADCT}, and \textit{IUCompPath} teams showed comparable BLEU scores but varied significantly in ROUGE values.
\end{enumerate}
These case studies are further discussed in \textcolor{blue}{Section~\ref{3.2}}, where we provide a detailed comparative analysis.
Based on these observations, we confirm that the Ranking Score employed in this challenge appropriately balanced penalization and reward, effectively reflecting the overall quality of the generated pathology reports.

\subsection{Qualitative Analysis of Generated Reports}\label{3.2}
We performed a qualitative assessment to examine how well the generated reports captured the diagnostic content and contextual features of the corresponding WSIs and ground-truth pathology reports. The representative WSIs with generated pathology reports are shown in Supplementary Tables (\textcolor{blue}{Table~\ref{tab:PIT_01_01467_01} -- \ref{tab:PIT_01_00111_01}}).

%%%%%%%%%%%%%%%%%%%%%% figs/PIT_02_00166_01 %%%%%%%%%%%%%%%%%%%%%
\begin{table*}[htbp]
\centering

\caption{\textbf{Quantitative attribute instability in generated pathology reports.} The \textcolor{red}{red} annotation highlights small tumor area in a prostate biopsy whole-slide image, corresponding to a ground-truth tumor volume of 5\%. Although several teams correctly identified the prostate origin, Gleason's score and grade group, estimation of tumor burden was highly unstable and frequently inflated, with generated tumor-volume values ranging from 10\% to 90\%. (\hl{organColor}{\phantom{--}}=organ, \hl{procedureColor}{\phantom{--}}=procedure, \hl{correctColor}{\phantom{--}}=correct, \hl{incorrectColor}{\phantom{--}}=incorrect)
}
\label{tab:PIT_02_00166_01}

\scriptsize % \footnotesize
\renewcommand{\arraystretch}{1.6}
\setlength{\tabcolsep}{5pt}

\begin{tabularx}{\textwidth}{|Y|Y|Y|Y|}
\hline

% WSI image
\multicolumn{4}{|l|}{\textbf{Whole Slide Image}} \\ \hline
\multicolumn{4}{|c|}{
    \rule{0pt}{15pt} 
    \includegraphics[width=0.6 \textwidth]{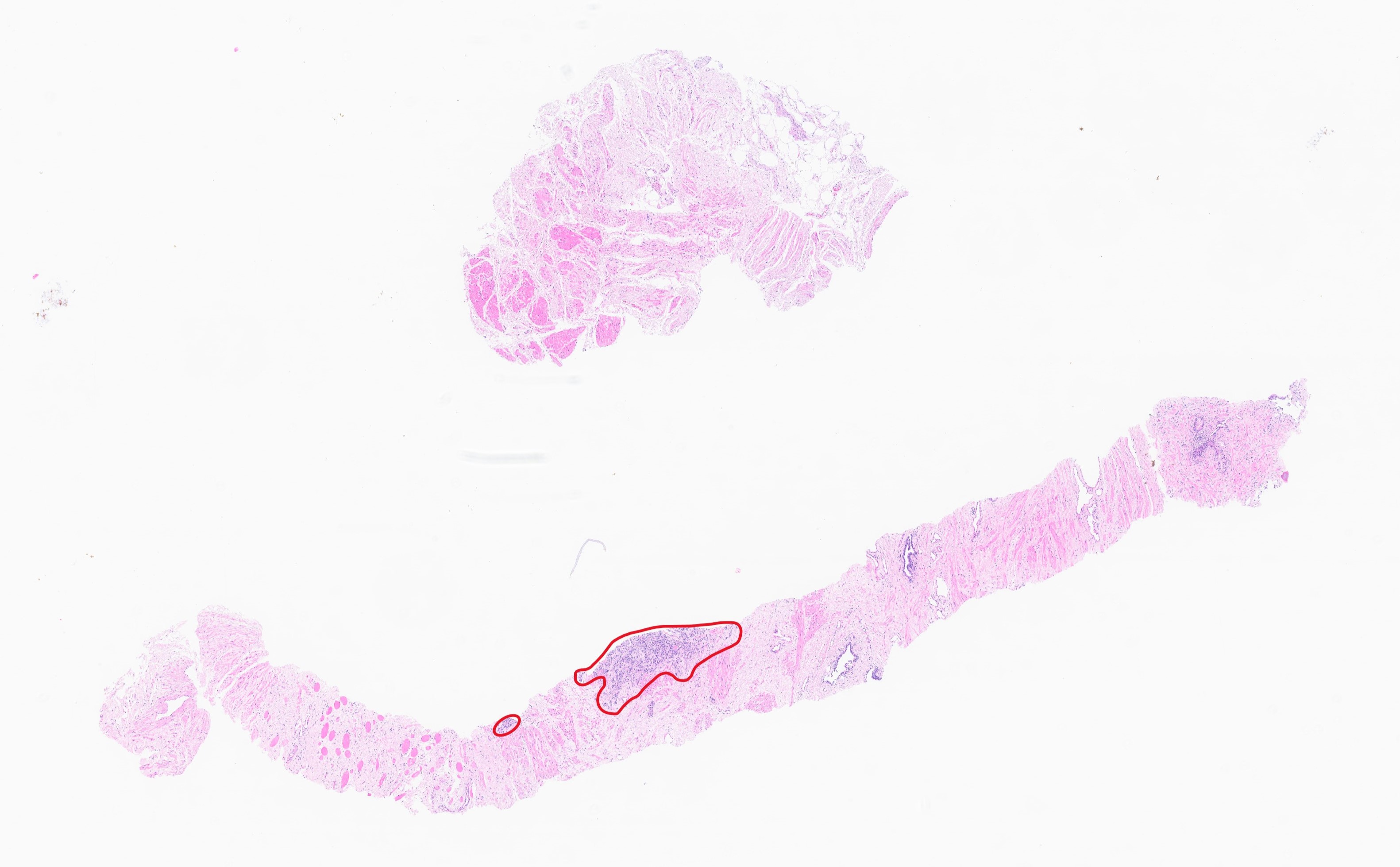} 
    \rule{0pt}{15pt} 
} \\ \hline

\multicolumn{2}{|l|}{\textbf{Ground Truth}} & \textbf{ADCT (0.4056)} & \textbf{ICGI (0.7316)} \\ \hline

% Ground Truth
\multicolumn{2}{|p{0.45\textwidth}|}{
    \hl{organColor}{Prostate,} \hl{procedureColor}{biopsy;} \newline
    \tab Acinar adenocarcinoma, \newline
    \tab Gleason's score 8 (4+4), \newline
    \tab grade group 4, \newline
    \tab tumor volume: 5\% \newline
} & 

    % ADCT
    \hl{correctColor}{Prostate, biopsy;} \newline
    \tab \hl{incorrectColor}{No tumor present} & 

    % ICGI
    \hl{correctColor}{Prostate, biopsy;} \newline
    \tab \hl{correctColor}{Acinar adenocarcinoma,} \newline
    \tab \hl{correctColor}{Gleason's score} \hl{incorrectColor}{6 (3+3),} \newline
    \tab \hl{correctColor}{grade group} \hl{incorrectColor}{1,} \newline
    \tab \hl{correctColor}{tumor volume: } \hl{incorrectColor}{80\%}\\ \hline

\textbf{ICL\_PathReport (0.9751)} & \textbf{IMAGINE\_Lab (0.7317)} & \textbf{IUCompPath (0.4056)} & \textbf{MedInsight-ViseurAI (0.9749)} \\ \hline

% ICL_PathReport
\hl{correctColor}{Prostate, biopsy;} \newline
\tab \hl{correctColor}{Acinar adenocarcinoma,} \newline
\tab \hl{correctColor}{Gleason's score 8 (4+4),} \newline 
\tab \hl{correctColor}{grade group 4,} \newline
\tab \hl{correctColor}{tumor volume: } \hl{incorrectColor}{15\%} & 

% IMAGINE_Lab
\hl{correctColor}{Prostate, biopsy;} \newline
\tab \hl{correctColor}{Acinar adenocarcinoma,} \newline
\tab \hl{correctColor}{Gleason's score } \hl{incorrectColor}{10 (5+5),} \newline 
\tab \hl{correctColor}{grade group} \hl{incorrectColor}{5,} \newline
\tab \hl{correctColor}{tumor volume:} \hl{incorrectColor}{70\%} & 

% IUCompPath
\hl{correctColor}{Prostate, biopsy;} \newline
\tab \hl{incorrectColor}{No tumor present} & 

% MedInsight-ViseurAI
\hl{correctColor}{Prostate, biopsy;} \newline
\tab \hl{correctColor}{Acinar adenocarcinoma,} \newline
\tab \hl{correctColor}{Gleason's score 8 (4+4),} \newline 
\tab \hl{correctColor}{grade group 4,} \newline
\tab \hl{correctColor}{tumor volume:} \hl{incorrectColor}{50\%} \\ \hline

\textbf{REG\_Path (0.4056)} & \textbf{TeamTiger@REG2025 (0.9750)} & \textbf{PathoMozhi (0.3866)} & \textbf{NW-TIA (0.7316)} \\ \hline

% REG_Path
\hl{correctColor}{Prostate, biopsy;} \newline
\tab \hl{incorrectColor}{No tumor present} &

% TeamTiger@REG2025
\hl{correctColor}{Prostate, biopsy;} \newline
\tab \hl{correctColor}{Acinar adenocarcinoma, } \newline
\tab \hl{correctColor}{Gleason's score 8 (4+4), } \newline 
\tab \hl{correctColor}{grade group 4,} \newline
\tab \hl{correctColor}{tumor volume:} \hl{incorrectColor}{90\%} &

% PathoMozhi
\hl{correctColor}{Prostate, biopsy;} \newline
\hl{incorrectColor}{No tumor present} &

% NW-TIA
\hl{correctColor}{Prostate, biopsy;} \newline
\tab \hl{correctColor}{Acinar adenocarcinoma,} \newline
\tab \hl{correctColor}{Gleason's score} \hl{incorrectColor}{10 (5+5),} \newline 
\tab \hl{correctColor}{grade group} \hl{incorrectColor}{5,}\newline
\tab \hl{correctColor}{tumor volume:} \hl{incorrectColor}{10\%}  

\\ \hline
\end{tabularx}

\vspace{5mm}

\end{table*}

\subsubsection{ \textbf{Diagnostic Concordance: Attribute Fidelity and Robustness}}\label{3.2.1}

1) \textbf{Comprehensive Diagnostic Accuracy with Attribute-level Fidelity} 
The models generated pathology reports that showed a high level of diagnostic concordance with the ground-truth (GT) while preserving clinically relevant attribute-level pathologic parameters (\textcolor{blue}{Fig.\ref{fig:quantitative_anlaysis}}-(a)). In breast core biopsy case, the ICGI team correctly generated the main diagnosis as “Ductal carcinoma in situ (DCIS)” and appropriately specified the DCIS type, nuclear grade, and necrosis status in alignment with the GT report (\textcolor{blue}{Table~\ref{tab:PIT_01_01467_01}}).

Similarly, in prostate biopsies, the ICGI team's generated report--``Acinar adenocarcinoma, Gleason score 7 (3+4), grade group 2 (Gleason pattern 4: \textcolor{red}{30\%}), tumor volume: 90\%''--accurately captured the histologic pattern and Gleason grading in a structured format consistent with routine pathology reporting. Although tumor volume was overestimated (90\% vs. 50\% in the reference), the hierarchy and organization of the diagnostic elements closely mirrored real-world pathology reporting practice (\textcolor{blue}{Table~\ref{tab:PIT_01_05697_01}}). Notably, similar limitations in quantitative attribute estimation were also observed across other participating teams, which will be further discussed in \textcolor{blue}{Section~\ref{3.2.2}}.

2)  \textbf{Diagnostic Robustness Across Rare and Minor Pathologic Categories}
The models also exhibited notable diagnostic robustness in rare or minor diagnostic categories represented by fewer than 10 cases in the full dataset (\textcolor{blue}{Fig.\ref{fig:quantitative_anlaysis}}-(b)). Accurate generation was observed for entities such as malignant lymphoma, metaplastic carcinoma of the breast (ICGI team), and fundic gland polyp of the stomach (MTS\_REG team) (\textcolor{blue}{Table~\ref{tab:PIT_01_04276_01}, ~\ref{tab:PIT_01_00193_01}}, and  ~\ref{tab:PIT_01_09080_01}).

Although gastric malignant melanoma was misclassified in most models as chronic gastritis, the MTS\_REG team correctly generated “malignant melanoma,” suggesting preserved discriminatory capacity even in diagnostically uncommon settings. 

3) \textbf{High-fidelity Histomorphology-driven Phenotype Recognition} 
In selected challenging cases, the generated reports appeared to be primarily anchored to histomorphologic pattern recognition (i.e., what the tumor looks like on H\&E). 
Gastric malignant lymphoma was correctly generated by most models; For PIT\_01\_08124\_01, four models misclassified the case as “adenocarcinoma, poorly differentiated”. For PIT\_01\_09100\_01, two models misclassified the case as “small cell carcinoma” (\textcolor{blue}{Table~\ref{tab:PIT_01_09100_01}}). This error is consistent with the well‑recognized histomorphologic overlap encountered in limited gastric biopsies and represents a common diagnostic pitfall in routine practice, particularly among less‑experienced pathologists. 

A particularly illustrative example was the uterine cervix case with “Adenocarcinoma, favor colorectal primary”. Most models generated “Adenocarcinoma, moderately differentiated” of colon/rectum (\textcolor{blue}{Table~\ref{tab:PIT_01_03514_01}}). Although the anatomic site in the generated text was incorrect, the output captured a colorectal‑type gland‑forming phenotype consistent with the suspected primary, underscoring the model’s ability to encode and reproduce diagnostically relevant morphologic features.

\begin{table*}[t]
\caption{\textbf{Quantitative comparison of top-performing teams and benchmark models across evaluation metrics.} Comparison of the first-place teams from Phase 1 and Phase 2 of the REG 2025 Challenge with existing publicly available benchmark models.}
\label{tbl:benchmark_test}
\centering
\resizebox{\textwidth}{!}{
\begin{tabular*}{\textwidth}{@{\extracolsep{\fill}} L||c|cccc||c|cccc}
\toprule
\textbf{Model} & \textbf{Phase 1} & \textbf{ROUGE} & \textbf{BLEU} & \textbf{KEY} & \textbf{EMB}  & \textbf{Phase 2} & \textbf{ROUGE} & \textbf{BLEU} & \textbf{KEY} & \textbf{EMB} \\
\midrule
ICGI            & \textbf{0.8098} & \textbf{0.8188} & \textbf{0.6372} & \textbf{0.7514} & \textbf{0.9696} & 0.8472 & 0.8564 & 0.6852 & 0.8080 & 0.9759 \\
IMAGINE Lab     & 0.6258 & 0.7485 & 0.0895 & 0.5620 & 0.9177 & \textbf{0.8494} & \textbf{0.8572} & \textbf{0.6928} & \textbf{0.8094} & \textbf{0.9770} \\
MI-Gen          & 0.6928 & 0.6870 & 0.4480 & 0.6003 & 0.9414 & 0.7055 & 0.6981 & 0.4562 & 0.6223 & 0.9447 \\
HistGen         & 0.4576 & 0.5429 & 0.4182 & 0.3555 & 0.5708 & 0.4647 & 0.5500 & 0.4233 & 0.3664 & 0.5739 \\
\bottomrule
\end{tabular*}
}
\end{table*}

\subsubsection{\textbf{Diagnostic Discordance: Quantitative \textcolor{red}{Estimation Errors} and Overcalling}}\label{3.2.2}

1) \textbf{\textcolor{red}{Errors in Quantitative Attribute Estimation}} Despite strong pattern recognition, the models demonstrated instability in quantitative attribute estimation, with frequent inflation of numeric values consistent with  numeric hallucination. Tumor proportion was particularly prone to overestimation. In the representative prostate biopsy case in \textcolor{blue}{Table~\ref{tab:PIT_02_00166_01}}, \textit{ICL\_PathReport}, \textit{TeamTiger@REG2025}, and \textit{MedInsight-ViseurAI}  correctly identified acinar adenocarcinoma, Gleason score, and grade group and achieved high REG scores (0.9749-0.9751),
% but all overestimated tumor volume (80\%  vs. 5\%  in GT report).
\textcolor{red}{but all three overestimated tumor volume, predicting 15\%, 90\%, and 50\%, respectively, compared with 5\% in the GT report.}
Models that preserved the correct diagnosis of acinar adenocarcinoma but misclassified the Gleason score and grade group—\textit{IMAGINE\_Lab, ICGI, and NW-TIA}—received intermediate scores (0.73x). By contrast, models that failed to detect carcinoma and instead reported “No tumor present”—\textit{ADCT, IUCompPath, REG\_Path}, and \textit{PathoMozhi}—received substantially lower scores (0.3866–0.4056). Together, these findings show that the REG score tracked the hierarchical severity of diagnostic errors.

In breast biopsies, grading-related discordance was observed in mitotic activity and tubule formation scores. For tubule formation, scores were assigned as follows: <10\% for score 3, 10–75\% for score 2, >75\% for score 1 (\textcolor{blue}{Table~\ref{tab:PIT_01_00111_01}}). These findings suggest limited capacity to translate visual estimation onto rule-based grading frameworks required for routine reporting.

2) \textbf{Over-specification and Failure to Preserve Uncertainty}
A recurring error was diagnostic commitment in settings where diagnostic retention of a broader category would have been more appropriate. Rather than preserving diagnostically meaningful uncertainty, the model frequently collapsed diagnoses into definite labels. Fibroepithelial lesions were frequently over-specified as \textit{fibroadenoma} or \textit{phyllodes tumor}. Similarly, diagnoses such as “\textit{non-small cell carcinoma, favor adenosquamous carcinoma”} was collapsed into  definitive adenocarcinoma or squamous cell carcinoma.

In breast biopsies, atypical ductal hyperplasia (ADH) was commonly overcalled as DCIS, indicating difficulty preserving the conservative diagnostic threshold that pathologists apply at the ADH–DCIS boundary.

% \begin{figure}[htbp]
%     \centering
%     \includegraphics[width=1\linewidth]{figs/Figure_4.png}
%     \caption{Representative image and generated reports demonstrating quantitative attribute instability in a prostate biopsy specimen}
%     \label{fig:figure4}
% \end{figure}

%%%%%%%%%%%%%%%%%%%%%%%%%%%%%%%%%%%%%%%%%%%%%%%%%%%%%%%%%%%%%%%%%%%%%%%%%%%%%%%%%%%%4.

\begin{table*}[t]
\caption{\textbf{Category-wise summary of approaches and components.} Overview of methods used by participating teams in the REG 2025 Challenge.}
\label{tbl:Top11_summary}
\centering
\renewcommand{\arraystretch}{1.3}
\scriptsize
\begin{tabular}{p{1.5cm}|p{3cm}|p{3cm}|p{5cm}|p{2.5cm}}
\toprule
\textbf{Team} & \textbf{Report Processing} & \textbf{WSI Preprocessing} & \textbf{Model Architecture / Methods} & \textbf{Foundation Models} \\
\toprule
ICGI & Category-wise separation and typo correction & Trident & ABMIL-based tile integration + n-gram-constrained Transformer decoder & H-Optimus-1 \\
\midrule
ICL\_ PathReport & Redefined as a hierarchical annotation tree & Trident & PRISM slide encoder aggregation + Tree-of-Experts (ToE)-based hierarchical report generation & Virchow \\
\midrule
IMAGINE Lab & GPT-generated concept prompts for organ and diagnosis categories & Trident & MLP-based multimodal fusion + GPT-guided cross-attention Transformer decoder & CONCH, TITAN, GPT \\
\midrule
ADCT & Codebook generation for each cancer type & Organ-specific patch sizing & ABMIL-based slide embedding + Flan-T5-based structured report generation & CONCH, GigaPath, UNI, Virchow, HistoEncoder \\
\midrule
IUCompPath & HistGen-style tokenization & CLAM & LGH-based hierarchical encoding + Cross-modal context Transformer decoder & UNI, H-Optimus-1 \\
\midrule
TeamTiger @REG2025 & WsiCaption-style tokenization & Divided into non-overlapping patches & MI-Gen Transformer architecture & UNI \\
\midrule
MedInsight-ViseurAI & Sentence-BERT based semantic similarity post-processing & HSV-based patch selection & ViT-based visual encoder + BioGPT-based report generation decoder & UNI, BioGPT \\
\midrule
PathoMozhi & None & Canny-filtered patch selection + STAMPv2 preprocessing & Gated cross-attention Flamingo framework + BioGPT for pathology report generation & CONCH, BioGPT \\
\midrule
MTS\_REG \_2025 & Report normalization and structured template reconstruction & Prov-Gigapath preprocessing pipeline & RAG-based pipeline + Qwen-3 8B report generator & Prov-Gigapath, Qwen-3 8B \\
\midrule
NW-TIA & Auxiliary label generation & Trident & Embedding-driven BioBART framework with Multi-task auxiliary heads & CONCH, TITAN, BioBART \\
\midrule
REG\_Path & Q-A pair generation & Threshold-adaptive preprocessing & Two-stage fine-tuned SlideChat (VQA + report generation) & CONCH, SlideChat \\
\bottomrule
\end{tabular}
\end{table*}

\begin{figure*}[t]
	\centering
	\includegraphics[width=\textwidth]{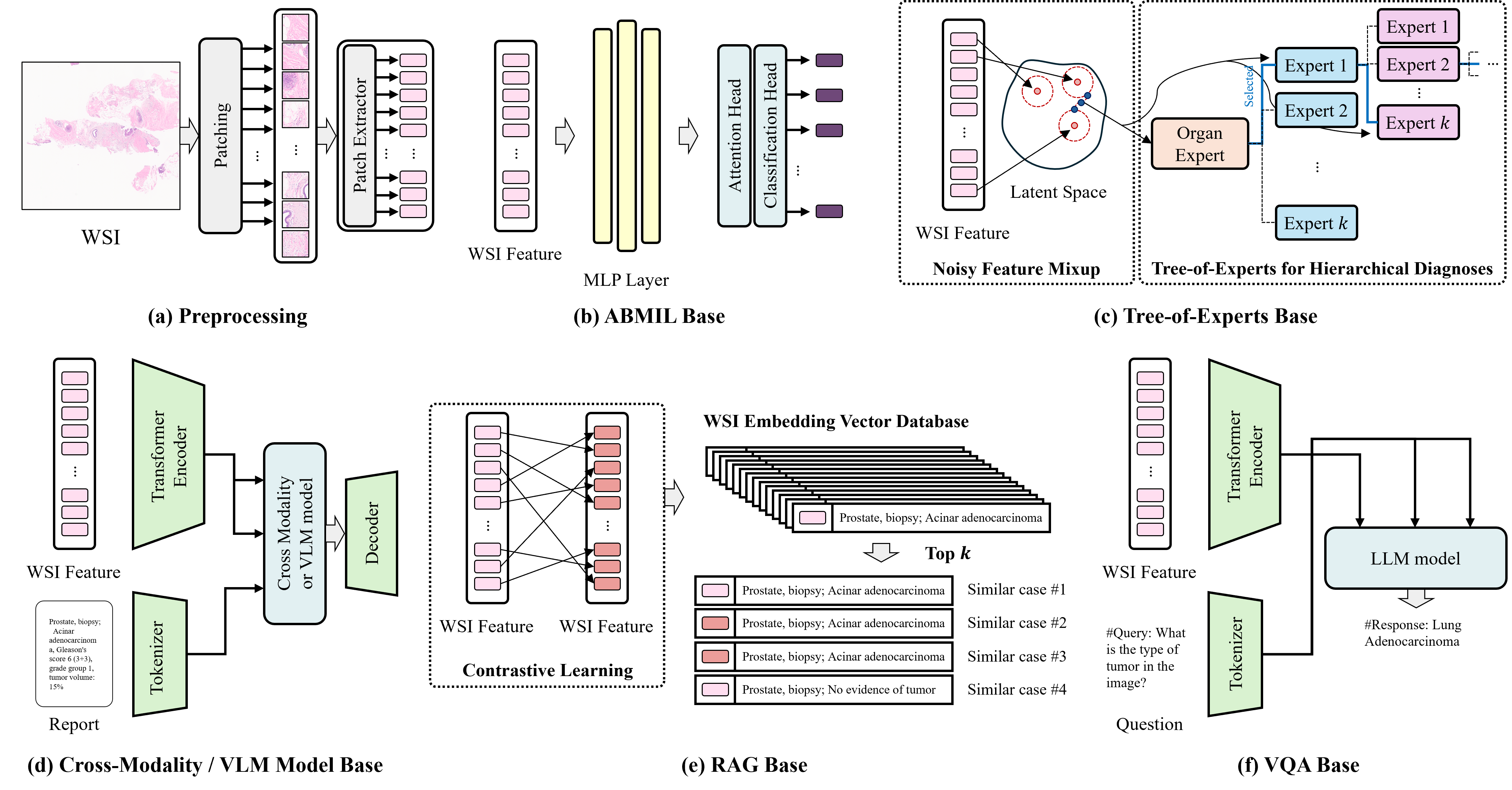}
	\caption{\textbf{Overview of major computational paradigms for pathology report generation from WSIs.} (a) Patch-level feature extraction using CNN or transformer encoders (b) Slide-level feature aggregation for whole-slide representation (c) Organ- or task-specific expert modules (d) Multi-modal fusion frameworks with vision–language models (VLMs) (e) Retrieval-augmented generation leveraging similar historical cases (f) VQA-based diagnostic reasoning for structured report generation.}
	\label{fig:models}
\end{figure*}

\subsection{Benchmark Comparison with Existing Approaches}
We further compared the top-performing challenge submissions with previously reported pathology report generation methods and general-purpose vision--language models~\cite{chen2024wsicaption,guo2024histgen}. As shown in \textcolor{blue}{Table~\ref{tbl:benchmark_test}}, publicly available benchmark models exhibited average performance drops of approximately 16.9\% in Phase 1 and 30.4\% in Phase 2 relative to the leading challenge submissions. Notably, many baseline approaches primarily formulate report generation as a direct sequence generation task from slide-level visual features, without explicitly modeling the structured and hierarchical nature of pathology reports.

In contrast, the highest-performing challenge methods consistently incorporated mechanisms for structured clinical representation, such as pathology-aware report organization, multimodal grounding, or intermediate diagnostic reasoning, enabling more effective preservation of clinically relevant information (\textcolor{blue}{Fig.~\ref{fig:quantitative_anlaysis}}-(c), (d)). These findings suggest that pathology report generation extends beyond conventional image captioning and instead requires structured multimodal reasoning capable of maintaining both hierarchical diagnostic relationships and clinically grounded concept consistency.

\section{Overview of Submitted Algorithms} \label{sec:algorithms}
For the final evaluation, we requested code and algorithm descriptions from the top 13 teams that achieved a final score of 0.6000 or higher, and conducted a detailed analysis of the 11 teams that responded. A summary of their algorithmic frameworks is presented in \textcolor{blue}{Table~\ref{tbl:Top11_summary}}. The submitted methods commonly followed three main stages -- Report Processing, WSI Preprocessing, and Model Architecture -- with most teams demonstrating active utilization of foundation models.

In terms of text preprocessing, two key strategies were consistently observed: normalization and hierarchical organization of the pathology reports. For WSI preprocessing (\textcolor{blue}{Fig.~\ref{fig:models}}-(a), many teams reused verified open frameworks such as Trident~\cite{zhang2025accelerating}, CLAM~\cite{lu2021data}, and Prov-Gigapath~\cite{xu2024whole}, reflecting a trend toward the reuse of standardized pipelines. Regarding model architecture, the majority of methods adopted ABMIL- or Transformer-based structures~\cite{ilse2018attention,shao2021transmil,chen2022scaling,liang2025wsi}, where approaches emphasizing contextual grounding rather than simple feature aggregation tended to achieve higher performance.

Accordingly, this paper provides a detailed comparative analysis of the submitted methods, categorized into two main perspectives: Preprocessing Strategies and Model Architectures. \textcolor{red}{Detailed descriptions of individual teams' methods are provided in Supplementary Section~\ref{sup:algorithm}, with their architectures illustrated in \textcolor{blue}{Fig.~\ref{fig:appendix_team_figures1}, \ref{fig:appendix_team_figures2}, and \ref{fig:appendix_team_figures3}}.}

\subsection{Preprocessing Strategies}
Most teams employed standardized MIL-based frameworks such as Trident~\cite{zhang2025accelerating} or CLAM~\cite{lu2021data}, or adopted similar mechanisms for WSI feature extraction. Several teams additionally incorporated strategies to account for organ-specific variations or to select visually salient tiles for more informative representations. However, the overall WSI preprocessing pipelines were largely consistent across participants, showing no significant performance differences attributable to preprocessing strategies alone.

In contrast, the report processing stage exhibited notable differences across participants. In particular, the top three teams--\textit{ICGI}, \textit{ICL\_PathReport}, and \textit{IMAGINE Lab}--which consistently achieved high performance in both Phase 1 and Phase 2, employed strategies that partitioned report information into distinct categories:

\begin{enumerate}[\textbullet]
    \item \textit{ICGI}: This team explicitly separated text sections using tags such as \texttt{<organ>} and \texttt{<diagnosis>}, aligning each section with the corresponding tile features extracted from the Trident slide encoder. They further reduced redundant expressions by utilizing section-wise n-gram frequency analysis.
    \item \textit{ICL\_PathReport}: Extending beyond simple section division, this team established hierarchical relationships among report categories using a hierarchical annotation tree. Each sentence was labeled by category, and the relationships between labels were defined as tree nodes, enabling the Tree-of-Experts (ToE) model to generate text sequentially from the root to preserve category-wise coherence.
    \item \textit{IMAGINE Lab}: This team automatically generated concept prompts for each report category using a GPT-based model. The resulting prompts were used as guidance signals to enhance fluency and contextual coherence during text generation.
\end{enumerate}

% These three teams, all of which leveraged category-specific report structuring, achieved substantially higher text quality than other participants, reaching average scores of ROUGE 0.8572 and BLEU 0.6853 in Phase 2.

These three teams \textcolor{red}{leveraged category-specific report generation strategies, differing in how structured information was incorporated into the generation process. Specifically, \textit{ICGI} employed n-gram-based output constraints and \textit{ICL\_PathReport} adopted hierarchical classification-based reconstruction, both relying on explicit report structuring, whereas \textit{IMAGINE Lab} utilized softer concept guidance for more flexible text generation. These category-specific approaches} achieved substantially higher text quality than other participants, reaching average ROUGE and BLEU scores of 0.8572 and 0.6853 in Phase 2.

In addition, the \textit{NW-TIA} team, which achieved the next highest performance following the top three teams in Phase 2, generated auxiliary clinical labels using a GPT-based text parser. These labels represented clinical categories not explicitly mentioned in the original reports and were used to guide the model in learning pathology-grounded reasoning regions. Similar to the top three teams, these four teams collectively treated pathology reports not as plain text sequences but as structured semantic maps, resulting in a markedly superior average KEY score of 0.7990 compared with other participants. Furthermore, the \textit{ADCT} team constructed cancer-type-specific codebooks to enable section-specific report generation, while the \textit{MTS\_REG\_2025 team} employed a template-based reconstruction approach for section-wise report synthesis.

In contrast, the remaining teams that tokenized the report data without category-wise structuring exhibited a noticeable decline in both BLEU and KEY scores. When comparing performance across different report processing strategies, teams employing category-aware methods achieved an average improvement of +0.09 in BLEU and +0.13 in KEY between Phase 1 and Phase 2, whereas tokenization-based approaches showed only marginal gains ($\Delta \text{BLEU} \approx +0.01,\ \Delta \text{KEY} \approx +0.02$). This performance gap further widened in Phase 2, with differences of approximately 0.14 in BLEU and 0.18 in KEY between the two groups.

These findings suggest that explicit report structuring leads to better cross-domain generalization and more stable preservation of key clinical terms under distributional shifts, underscoring the importance of developing structured report representations for robust pathology report generation.

\subsection{Model Architectures}

\subsubsection{ABMIL-Based Architectures}
The \textit{ICGI} and \textit{ADCT} teams employed Attention-based Multiple Instance Learning (ABMIL)~\cite{ilse2018attention} to generate slide-level embeddings, which were then passed to Transformer- or LLM-based text decoders in a conventional image-captioning pipeline for pathology report generation (\textcolor{blue}{Fig.~\ref{fig:models}}-(b)). ABMIL effectively summarizes a WSI into a global representation by aggregating tile-level features, enabling downstream text decoders to infer pathological context in a more stable and coherent manner.

The \textit{ICGI} team incorporated an n-gram constrained Transformer decoder to suppress hallucinations commonly observed in LLM-based report generation. This architectural design demonstrated correct predictions in several rare (long-tail) cases (\textcolor{blue}{Fig.~\ref{fig:quantitative_anlaysis}}-(b)). In addition, the generated reports occasionally reflected subtle histomorphologic features; for instance, the model explicitly mentioned microcalcifications measuring approximately 30 µm, indicating its ability to generate reports with fine-grained pathological detail.

The \textit{ADCT} team introduced a cancer-type-specific codebook to reinforce the prior knowledge required for Flan-T5–based structured generation~\cite{chung2024scaling}, and further integrated cancer-specific knowledge distillation, resulting in more efficient and domain-aware text synthesis. These models also exhibited strong organ anchoring characteristics. In typical cases, this property contributed to producing stable and internally consistent diagnostic outputs. However, in the breast lymphoma case, ADCT generated “breast metaplastic carcinoma” instead of lymphoma. In this instance, organ anchoring itself was preserved, but the model struggled to accurately capture the rare morphologic pattern.

\subsubsection{Tree-of-Experts-Based Architectures}
The \textit{ICL\_PathReport} team proposed a Tree-of-Experts (ToE) framework that departs from the conventional reliance on a single slide-level embedding (\textcolor{blue}{Fig.~\ref{fig:models}}-(c)). Instead, their method explicitly models the inherent hierarchical structure of pathology reports -- organ → diagnosis → subtype -- through hierarchical feature aggregation and a category-specialized expert hierarchy.
In this architecture, each node in the tree corresponds to a category-specific sub-expert, enabling the model to structurally disentangle diverse pathological knowledge encoded in the report. This design allows the model to separately learn organ-level semantics, diagnosis-level decision boundaries, and subtype-level fine-grained attributes, providing a clearer hierarchical partitioning of the feature space compared with standard MIL-based global embeddings. As a result, the approach offers inherent advantages in ensuring slot-level information disentanglement, making it better aligned with the structured nature of pathology reporting. A detailed illustration of the tree architecture is provided in \textcolor{blue}{Appendix~\ref{appendix:ICL_PathReport}}.

This architecture aligns well with the inherent hierarchical structure of pathology reporting, particularly for breast and prostate specimens. It is especially effective at capturing fine-grained diagnostic attributes (e.g., grade/score subcomponents, and tumor extent) In the breast lymphoma case, the model generated “prostate lymphoma,” even though prostate lymphoma cases were relatively few in the dataset. This pattern suggests that the model first committed to an incorrect organ-level routing decision (prostate) and subsequently searched for the most morphologically compatible diagnosis within that constrained branch. This behavior underscores a trade-off: hierarchical decomposition enhances logical clarity but reduces flexibility when upstream routing errors occur.

\subsubsection{Cross-Modality Transformer-Based Architectures}
The \textit{IMAGINE Lab}, \textit{IUCompPath}, and \textit{TeamTiger@REG2025} teams adopted multi-modal Transformer architectures that introduce cross-attention modules between image tokens extracted from the WSI encoder and textual tokens from the report(\textcolor{blue}{Fig.~\ref{fig:models}}-(d)). This design explicitly models the interactions between visual and textual modalities, allowing the model to learn fine-grained alignment between specific regions in the WSI and corresponding categories or expressions in the generated report. Compared with approaches that simply feed a global WSI embedding into a text decoder, these cross-modality frameworks offer substantially greater flexibility in capturing region-text correspondences.

The \textit{IMAGINE Lab} team generated organ- and diagnosis-level concept prompts using a GPT-based model (as detailed in \textcolor{blue}{Table~\ref{tbl:concepts}}.) and used them as additional guidance signals during multi-modal fusion, thereby strengthening semantic alignment between visual and textual representations.

The \textit{IUCompPath} team introduced a Cross-modal Context Module (CMC)~\cite{guo2024histgen} that contextually integrates image and text tokens, jointly supporting hierarchical encoding and cross-modal reasoning.

Meanwhile, the \textit{TeamTiger@REG2025} team preserved patch-level spatial information through a Position-Aware module, combining these spatially enriched image tokens with text tokens via cross-attention to enhance spatially grounded reasoning throughout the report generation process.

The outputs of these models were largely driven by histomorphologic pattern. In some cases, the generated pathology reports remained pathologically plausible despite incorrect organ-of-origin assignment. Language quality varied widely across these architectures, reflecting differences in the underlying language model and in how strictly report generation was controlled.

\subsubsection{Vision-Language Model-Based Architectures}
The \textit{MedInsight-ViseurAI}, \textit{PathoMozhi}, and \textit{NW-TIA} teams adopted Vision-Language Model (VLM)-based architectures, in which WSI features extracted from powerful vision encoders were provided as input to biomedical foundation models such as BioGPT~\cite{luo2022biogpt} or BioBART~\cite{yuan2022biobart} for pathology report generation (\textcolor{blue}{Fig.~\ref{fig:models}}-(d)). Because BioGPT and BioBART models are pretrained extensively on biomedical terminology and clinical narrative structures, they are particularly well suited for medical domain text generation tasks, including pathology reporting.

The \textit{MedInsight-ViseurAI} and \textit{PathoMozhi} teams employed BioGPT decoders to enhance the generation of domain-specific terminology in WSI-based medical captioning.

The \textit{PathoMozhi} team further incorporated a Flamingo-style gated cross-attention mechanism~\cite{alayrac2022flamingo} to refine vision-language alignment.

Meanwhile, the \textit{NW-TIA} team used a BioBART encoder-decoder architecture augmented with multi-task auxiliary heads to improve training stability and preserve the semantic richness of WSI features throughout the generation process.

\subsubsection{Retrieval-Augmented Generation (RAG)-Based Architectures}
The \textit{MTS\_REG\_2025} team proposed a Retrieval Augmented Generation (RAG) framework in which WSI-derived features were used to retrieve semantically similar pathology reports, and both the retrieved documents and the input context were provided to the Qwen3-8B~\cite{yang2025qwen3} model to generate the final report (\textcolor{blue}{Fig.~\ref{fig:models}}-(e)). RAG-based approaches mitigate hallucinations by incorporating explicit external knowledge, offering improved semantic grounding--particularly for rare subtypes or cases involving complex morphological patterns.

To further enhance reliability, the team employed an Adaptive Retrieval strategy that dynamically adjusts the number of retrieved documents based on the similarity gap. This mechanism prevents both redundant retrieval (which introduces noise) and insufficient retrieval (which leads to information loss), thereby contributing directly to improved report consistency and semantic alignment.

These approaches achieved noticeably higher accuracy on rare (long-tail) cases, but showed reduced robustness on majority (high-prevalence) cases.

\subsubsection{Visual Question Answering (VQA)-Based Architectures}
The \textit{REG\_Path} team proposed a two-stage framework that departs from conventional end-to-end captioning pipelines (\textcolor{blue}{Fig.~\ref{fig:models}}-(f)). Instead of directly generating the entire report, their method first performs slot-level querying of the WSI using a SlideChat~\cite{chen2025slidechat}-based VQA model, and then composes the final report by assembling the predicted answers into a predefined report template.
% By prompting the VQA model to explicitly answer key pathological slots--such as organ, histologic type, and invasiveness--this approach ensures clearer concept grounding, reduces hallucinations, and facilitates the generation of clinically meaningful expressions.
Furthermore, the \textit{REG\_Path} team adopted a two-stage training strategy, consisting of VQA pretraining followed by report fine-tuning.
% This progressive incorporation of slot-level supervision signals enhanced both the accuracy and consistency of the final reports.

%%%%%%%%%%%%%%%%%%%%%%%%%%%%%%%%%%%%%%%%%%%%%%%%%%%%%%%%%%%%%%%%%%%%%%%%%%%%%%%%%%%%5.
\section{Discussion}
The quantitative evaluation results also clearly revealed the limitations of conventional natural language generation metrics for pathology report generation. In this challenge, the overall ranking score was constructed by combining ROUGE, BLEU, KEY, and EMB metrics, with the KEY and EMB metrics demonstrating particularly strong consistency with the final ranking. In contrast, ROUGE and BLEU exhibited anomalously high or low values for certain teams, indicating that these metrics do not always align with clinical correctness. For example, some models achieved high ROUGE scores despite failing to generate critical diagnostic terms accurately, while others obtained relatively low BLEU scores but correctly preserved clinically important diagnostic attributes, such as Gleason score, nuclear grade, and necrosis status, as demonstrated in the qualitative cases described in \textcolor{blue}{Section~\ref{3.2.1}}. These findings suggest that lexical overlap-based metrics alone may not adequately capture the clinical accuracy required for pathology report generation.

This discrepancy between metric performance and clinical correctness arises because pathology reports, unlike general captioning tasks, require structured diagnostic reasoning and precise representation of critical concepts rather than superficial sentence similarity. Key diagnostic attributes--such as Gleason score, tumor proportion, and histologic subtype--play a decisive role in determining the clinical interpretation of the report. However, conventional metrics do not sufficiently evaluate concept-level correctness. While the KEY score used in this study partially mitigates this limitation, more precise evaluation will require the development of pathology-specific metrics, such as ontology-aware evaluation, slot-level accuracy metrics, or clinically grounded concept consistency measures.

\textcolor{red}{
Beyond evaluation metrics, the reporting format itself is an important consideration when interpreting model performance. The standardized format adopted in this challenge, based on the College of American Pathologists (CAP) guidelines, provides a consistent basis for assessing essential diagnostic attributes beyond linguistic similarity. At the same time, its predefined structure may be particularly compatible with approaches that explicitly predict diagnostic attributes or reconstruct reports using templates. These results establish comparative performance in generating clinically structured reports, while extending evaluation to more flexible reporting formats would provide further insights into broader report generation capabilities.
}

Furthermore, both qualitative and quantitative analyses revealed recurring issues of numeric hallucination and diagnostic misclassification, reflecting fundamental limitations of current VLM-based report generation frameworks.
In particular,
% for structured diagnostic attributes such as tumor proportion, grading score, and subtype classification, models tend to generate text based on semantic plausibility rather than grounded evidence.
\textcolor{red}{
errors in structured diagnostic attributes such as tumor proportion, grading score, and subtype classification demonstrate that semantically plausible reports do not always preserve clinically accurate information.
}
% This limitation stems from the probabilistic next-token prediction nature of language models.
\textcolor{red}{This limitation can arise from multiple factors, including the probabilistic nature of next-token prediction in language models, limited region-level or measurement-specific supervision, insufficient visual resolution for capturing fine-grained histopathological features, and imperfect image-text alignment.}
% These findings indicate that multimodal alignment alone is insufficient to address this issue and highlight the need for architectures that incorporate explicit diagnostic reasoning processes \textcolor{red}{and fine-grained visual evidence}.
\textcolor{red}{
These findings highlight the importance of complementing multimodal alignment with explicit diagnostic reasoning processes and fine-grained visual evidence.
}

In this context, reasoning-augmented generation frameworks represent a promising direction for pathology report generation. Approaches such as slot-based intermediate prediction~\cite{wu2023slotdiffusion,xu2024slot,chi2025slot}, chain-of-thought style structured generation~\cite{wei2022chain,wang2022self,zhou2022least,zhang2023multimodal}, and VQA-based diagnostic decomposition~\cite{antol2015vqa,khan2023exploring} enable models to perform complex diagnostic reasoning in a stepwise and interpretable manner.
Notably,
% in this challenge, approaches incorporating VQA-based frameworks demonstrated reduced hallucination and improved slot-level concept grounding. This suggests that structured, reasoning-based generation frameworks are more suitable for pathology report generation than conventional end-to-end captioning approaches.
\textcolor{red}{approaches incorporating VQA-based frameworks decomposed report generation into explicit diagnostic queries, providing a structured mechanism for predicting clinical attributes. This highlights the potential of reasoning-based generation as an alternative to conventional end-to-end captioning approaches.}
\textcolor{red}{While this challenge focuses on report-level diagnostic content and generation quality, region-specific evidence support for individual statements and the ability to answer follow-up questions or revise uncertain findings remain complementary capabilities beyond its evaluation scope. These aspects can be further investigated through interactive VQA, agent-based workflows, and complementary benchmarks such as EviPathBench~\cite{liao2026evipathbench}.}
% In addition, Retrieval-Augmented Generation (RAG) and hierarchical expert-based architectures were found to contribute to reducing hallucination and improving semantic grounding. By leveraging external evidence or structured expert knowledge rather than relying solely on internal model representations, these approaches enhance diagnostic reliability. 
\textcolor{red}{Whether retrieval or hierarchical modeling reduces hallucinations or improves grounding remains to be tested.}
This highlights that reasoning-aware multimodal architectures and external knowledge integration should be considered key design components in future pathology foundation models.

Overall, the results of this challenge demonstrate that pathology report generation is not merely an image-to-text translation task, but rather a high-dimensional multimodal reasoning problem involving structured diagnostic reasoning, quantitative attribute estimation, and uncertainty-aware decision-making. Accordingly, future research should focus on (1) developing clinically grounded evaluation metrics, (2) designing reasoning-integrated generation frameworks, and (3) incorporating structured pathology knowledge to improve reliability and clinical applicability.

%%%%%%%%%%%%%%%%%%%%%%%%%%%%%%%%%%%%%%%%%%%%%%%%%%%%%%%%%%%%%%%%%%%%%%%%%%%%%%%%%%%%6.
\section{Conclusion}
In this study, through the REG 2025 Challenge, we constructed and publicly released the first large-scale Pan-Asia multi-institutional pathology dataset comprising approximately 10,500 WSI–report pairs. Based on this dataset, we systematically analyzed and compared the models submitted by participating teams for pathology report generation.

Analysis across diverse participant approaches demonstrated that Vision–Language Model (VLM) based frameworks can generate clinically coherent pathology reports across a wide range of organs and pathological categories. Furthermore, the models maintained stable performance on an independent European cohort, highlighting their cross-domain generalization capability.
In particular, approaches that leveraged structured representations of pathology reports and multimodal grounding strategies achieved superior performance, underscoring the importance of structured multimodal modeling in pathology report generation.

Importantly, the REG Challenge frames pathology report generation not merely as an image captioning task, but as a complex multimodal diagnostic understanding problem, and provides a comprehensive benchmark for its systematic evaluation. Through comparative analysis of diverse modeling strategies, we identify multimodal alignment, structured report representation, and clinically meaningful feature integration as key factors influencing report generation performance.
As the first large-scale benchmark enabling systematic evaluation of pathology report generation and multimodal diagnostic reasoning, the REG Challenge provides an essential foundation for future foundation-level computational pathology research.

%%%%%%%%%%%%%%%%%%%%%%%%%%%%%%%%%%%%%%%%%%%%%%%%%%%%%%%%%%%%%%%%%%%%%%%%%%%%%%%%%%%%

\section{Acknowledgement}
This work was partly supported by the National Research Foundation of Korea (NRF) grants funded by the Korea government (MSIT) (RS-2026-25475860, RS-2025-00520578, RS-2025-02215813, and RS-2025-02634603); by the NRF grant funded by the Korea government (MSIT and MOE) (RS-2025-16063688); by the Digital-Bio AI+X Global Innovative Talent Nurturing Project through the NRF funded by the Korea government (MSIT) (RS-2024-00441029); by a Korea University Grant (K2612911); \textcolor{red}{by the Commercialization Promotion Agency for R\&D Outcomes (COMPA) through a grant funded by the Ministry of Science and ICT (MSIT) (RS-2026-25539491);} by funding from Interreg Meuse-Rhine (NL-BE-DE) EU / EFRE North Rhine-Westphalia (Project DigiPathConnect) and the Wilhelm Sander Foundation (Munich, Germany) (Project 2022.040.1); and by the AI Computing Infrastructure Enhancement (GPU Rental Support) User Support Program of the Ministry of Science and ICT (MSIT), Republic of Korea (No. RQT-25-090048). \textcolor{red}{We also acknowledge the support of the Scientific Committee of the Asian Society of Digital Pathology (ASDP).}

% Use the style of numbering in square brackets.
% If nothing is used, default style will be taken.
%\begin{enumerate}[a)]
%\item 
%\item 
%\item 
%\end{enumerate}  

%\begin{itemize}
%\item 
%\item 
%\item 
%\end{itemize}  

%\begin{description}
%\item[]
%\item[] 
%\item[] 
%\end{description}  

% \clearpage %%Remove this from your manuscript

% % Figure
% \begin{figure}%[]
%   \centering
% %    \includegraphics{}
%     \caption{}\label{fig1}
% \end{figure}

% \begin{table}%[]
% \begin{tabular*}{\tblwidth}{@{}LL@{}}
% \toprule
% \midrule
%  & \\
%  & \\
%  & \\
%  & \\
% \bottomrule
% \end{tabular*}
% \end{table}

% Uncomment and use as the case may be
%\begin{theorem} 
%\end{theorem}

% Uncomment and use as the case may be
%\begin{lemma} 
%\end{lemma}


\printcredits

\bibliographystyle{cas-model2-names}

\bibliography{REG2025-refs}

% Appendix
\appendix
\clearpage
% numbering reset
\setcounter{table}{0}
\renewcommand{\thetable}{S\arabic{table}}
\renewcommand{\theHtable}{S\arabic{table}}   % ← 추가

\setcounter{figure}{0}
\renewcommand{\thefigure}{S\arabic{figure}}
\renewcommand{\theHfigure}{S\arabic{figure}} % ← 추가

\begin{table*}[t]
\color{red}
\caption{\textbf{REG 2025 Challenge (Phase 1): Quantitative performance comparison.} Evaluation of participating teams across report generation quality metrics.}
\label{tbl:phase1_results}
\centering
\resizebox{\textwidth}{!}{
\begin{tabular*}{\textwidth}{@{\extracolsep{\fill}} L|L||c|cccc}
\toprule
\textbf{Rank} & \textbf{Team} & \textbf{Phase 1} & \textbf{ROUGE} & \textbf{BLEU} & \textbf{KEY score} & \textbf{EMB score} \\
\midrule
1  & ICGI                       & \textbf{0.8098} & \textbf{0.8188} & \textbf{0.6372} & \textbf{0.7514} & \textbf{0.9696} \\
2  & ICL\_PathReport            & 0.7352 & 0.7291 & 0.3644 & 0.7049 & 0.9640 \\
3  & IMAGINE Lab                & 0.6258 & 0.7485 & 0.0895 & 0.5620 & 0.9177 \\
4  & ADCT                       & 0.7704 & 0.7753 & 0.5570 & 0.7097 & 0.9554 \\
5  & IUCompPath                 & 0.7906 & 0.7954 & 0.6173 & 0.7262 & 0.9607 \\
6  & TrustPath                  & 0.6565 & 0.6415 & 0.4884 & 0.5842 & 0.8445 \\
7  & TeamTiger@REG2025          & 0.7777 & 0.7805 & 0.5784 & 0.7116 & 0.9642 \\
8  & MedInsight-ViseurAI        & 0.5326 & 0.4754 & 0.2254 & 0.4092 & 0.8792 \\
9  & PathoMozhi                 & 0.4611 & 0.4877 & 0.1155 & 0.2868 & 0.8530 \\
10 & MTS\_REG\_2025             & 0.6046 & 0.5749 & 0.3497 & 0.4802 & 0.9128 \\
12 & REG\_Path                  & 0.6907 & 0.6626 & 0.4458 & 0.6114 & 0.9331 \\
13 & TokenStreamers             & 0.7248 & 0.7211 & 0.4860 & 0.6468 & 0.9501 \\
14 & NUST-SEECS                 & 0.5144 & 0.4810 & 0.2394 & 0.3555 & 0.8806 \\
15 & Clinical\_Omics\_Institute & 0.5531 & 0.5109 & 0.2535 & 0.4275 & 0.8916 \\
16 & JX\_REG                    & 0.5073 & 0.4470 & 0.1929 & 0.3676 & 0.8810 \\
17 & DSG                        & 0.4385 & 0.3617 & 0.1600 & 0.3108 & 0.7862 \\
19 & virasoft                   & 0.1691 & 0.0660 & 0.0257 & 0.0453 & 0.4573 \\
20 & Shubham47                  & 0.4402 & 0.3881 & 0.1453 & 0.2715 & 0.8387 \\
23 & May\_REG                   & 0.5245 & 0.4679 & 0.2126 & 0.3924 & 0.8851 \\
24 & IMI\_REG                   & 0.5027 & 0.5591 & 0.0377 & 0.3489 & 0.9122 \\
\bottomrule
\end{tabular*}
}
\end{table*}

\begin{table*}[t]
\color{red}
\caption{\textbf{REG 2025 Challenge (Phase 2): Quantitative performance comparison.} Evaluation of participating teams across report generation quality metrics.}
\label{tbl:phase2_results}
\centering
\resizebox{\textwidth}{!}{
\begin{tabular*}{\textwidth}{@{\extracolsep{\fill}} L|L||c|cccc}
\toprule
\textbf{Rank} & \textbf{Team} & \textbf{Phase 2} & \textbf{ROUGE} & \textbf{BLEU} & \textbf{KEY score} & \textbf{EMB score} \\
\midrule
1  & ICGI                       & 0.8472 & 0.8564 & 0.6852 & 0.8080 & 0.9759 \\
2  & ICL\_PathReport            & 0.8415 & \textbf{0.8578} & 0.6780 & 0.7937 & \textbf{0.9788} \\
3  & IMAGINE Lab                & \textbf{0.8494} & 0.8572 & \textbf{0.6928} & \textbf{0.8094} & 0.9770 \\
4  & ADCT                       & 0.8040 & 0.8037 & 0.6154 & 0.7507 & 0.9697 \\
5  & IUCompPath                 & 0.7899 & 0.7887 & 0.6141 & 0.7297 & 0.9585 \\
6  & TrustPath                  & 0.8127 & 0.8258 & 0.6340 & 0.7597 & 0.9661 \\
7  & TeamTiger@REG2025          & 0.7763 & 0.7790 & 0.5773 & 0.7119 & 0.9604 \\
8  & MedInsight-ViseurAI        & 0.8093 & 0.8159 & 0.6148 & 0.7604 & 0.9683 \\
9  & PathoMozhi                 & 0.7960 & 0.8574 & 0.6821 & 0.6792 & 0.9781 \\
10 & MTS\_REG\_2025             & 0.7584 & 0.7607 & 0.5436 & 0.6923 & 0.9528 \\
11 & NW-TIA                     & 0.8282 & 0.8362 & 0.6467 & 0.7850 & 0.9726 \\
12 & REG\_Path                  & 0.6510 & 0.6223 & 0.3959 & 0.5572 & 0.9179 \\
13 & TokenStreamers             & 0.5773 & 0.5277 & 0.2863 & 0.4718 & 0.8882 \\
14 & NUST-SEECS                 & 0.6121 & 0.6439 & 0.4328 & 0.4403 & 0.9151 \\
15 & Clinical\_Omics\_Institute & 0.5984 & 0.5583 & 0.3300 & 0.4921 & 0.8944 \\
16 & JX\_REG                    & 0.5309 & 0.4709 & 0.2133 & 0.4089 & 0.8825 \\
17 & DSG                        & 0.4286 & 0.3486 & 0.1412 & 0.3045 & 0.7780 \\
18 & sk                         & 0.5277 & 0.5371 & 0.1519 & 0.4015 & 0.8790 \\
19 & virasoft                   & 0.4590 & 0.3906 & 0.1887 & 0.3039 & 0.8350 \\
20 & Shubham47                  & 0.3849 & 0.3283 & 0.1137 & 0.1692 & 0.8364 \\
21 & DiceMed\_labs              & 0.3577 & 0.2731 & 0.0000 & 0.2003 & 0.7887 \\
22 & UTWL                       & 0.3110 & 0.1772 & 0.0204 & 0.0894 & 0.8185 \\
\bottomrule
\end{tabular*}
}
\end{table*}

%%%%%%%%%%%%%%%%%%%%%%%%%%%%%%%%%%%%%%%%%%%%
\section{Contributions and Methods of Each Team} \label{sup:algorithm}
\subsection{ICGI (\textcolor{blue}{Fig.~\ref{fig:appendix_team_figures1}}-(a))}
\textbf{Contributions} The ICGI team’s primary contributions include the utilization of the H-optimus-1 foundation model to extract tile-level embeddings and the multiple-instance slide level embedding to predict the anatomical site and the procedure that has been taken. Also, they implemented a decoder for predicting slide descriptions, coupled with a post-processing mechanism to refine the output text. 

\textbf{Method} The pipeline begins by extracting tile-level features using H-optimus-1, following color correction and tissue detection. They utilize an eight parallel attention heads within the ABMIL backbone, which concatenate weighted sums into a slide-level representation. This embedding is fed into three task-specific heads: two linear classifiers for site and procedure, and a Transformer decoder for text generation. During inference, an n-gram mask is derived from the training corpus to filter out unobserved token sequences.

%%%%%%%%%%%%%%%%%%%%%%%%%%%%%%%%%%%%%%%%%%%%
\subsection{ICL\_PathReport (\textcolor{blue}{Fig.~\ref{fig:appendix_team_figures1}}-(b))}\label{appendix:ICL_PathReport}
\textbf{Contributions} The ICL\_PathReport team proposes a Tree-of-Experts (ToE) framework that hierarchically predicts report components, including organs, procedures, and histological types, through multiple lightweight MLP classifiers. To address the data scarcity challenge in deep diagnostic levels, they introduce a Noisy Feature Mixup strategy, which enhances model generalization and prediction calibration.

\textbf{Method} The system employs the Trident framework for patch extraction with Virchow for patch-level features, which are then aggregated into slide-level embedding using the PRISM Perceiver encoder. The ToE architecture employs a hierarchy of specialized MLP experts designed to predict detailed diagnostic attributes. During training, the team applies noise perturbation and weighted Mixup to the slide features to augment the limited dataset available for rare or complex diagnoses. The final structured report is synthesized by navigating the ToE's forward path, ensuring that the generated output strictly adheres to the inherent hierarchy of clinical pathology reports.

An example illustrating the hierarchical tree structure of the pathology annotation format is shown below.
\begin{verbatim}
{
  "id": "PIT_01_00002_01.tiff",
  "organ": "Breast",
  "procedure": "core-needle biopsy",
  "sub_feature": [
    {
      "principal": "Invasive carcinoma of no special type",
      "scores": "",
      "sub_feature": [
        {
          "principal": "grade",
          "scores": "I",
          "sub_feature": []
        },
        {
          "principal": "Tubule formation",
          "scores": "2",
          "sub_feature": []
        },
        {
          "principal": "Nuclear grade",
          "scores": "2",
          "sub_feature": []
        },
        {
          "principal": "Mitoses",
          "scores": "1",
          "sub_feature": []
        }
      ]
    }
  ]
}
\end{verbatim}

\begin{figure*}[t]
	\centering
    \caption{\textbf{Overview of model architectures from teams ICGI, ICL\_PathReport, and IMAGINELab.} ICGI utilizes an ABMIL backbone with H-Optimus-1 for multi-task classification and report generation, while ICL\_PathReport features a "Tree-of-Experts" decoder for hierarchical diagnosis. IMAGINELab integrates CONCH and Titan encoders with MLP adapters and a concept-guided Transformer decoder.}
	\label{fig:appendix_team_figures1}
	\includegraphics[width=.7\textwidth]{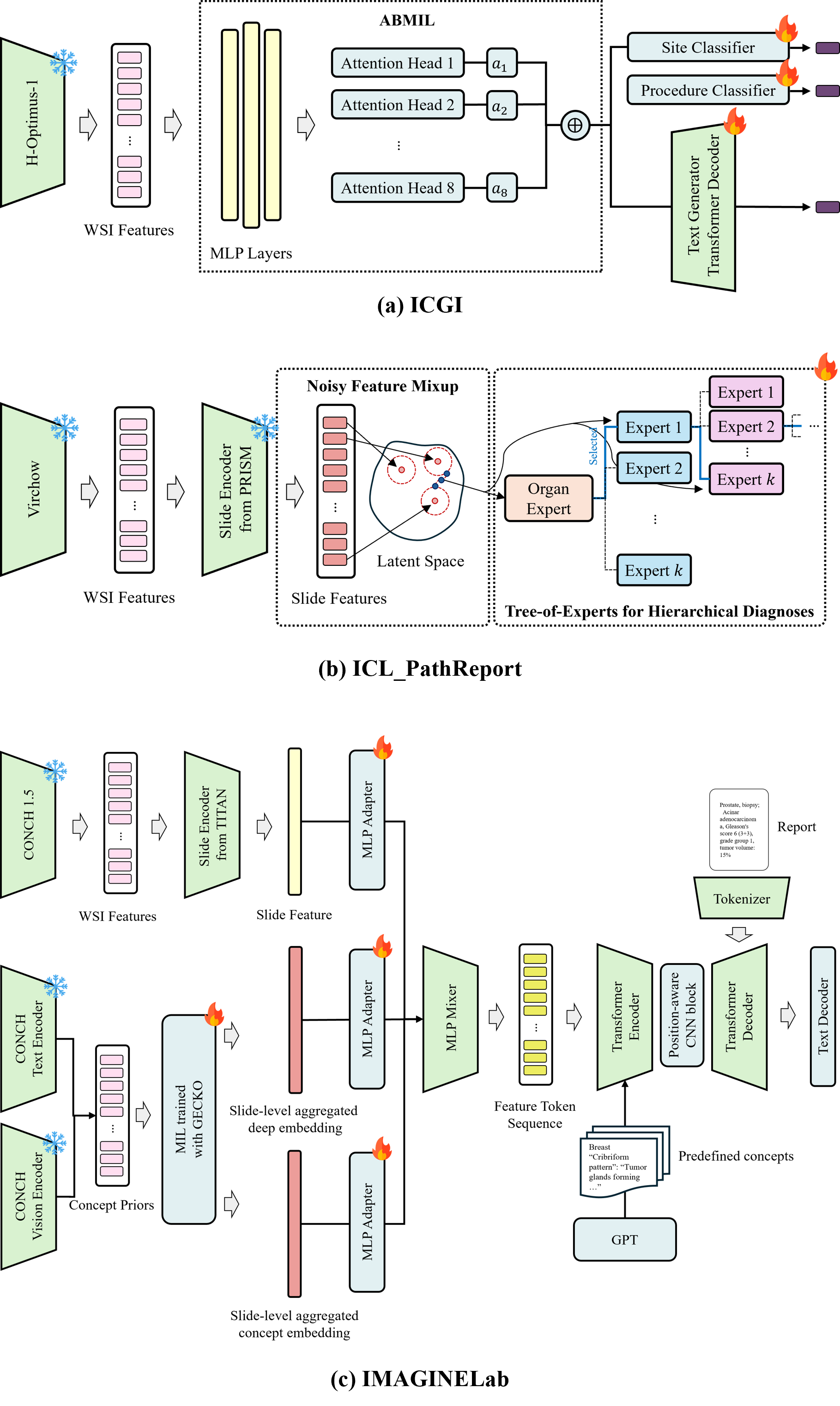}
\end{figure*}

%%%%%%%%%%%%%%%%%%%%%%%%%%%%%%%%%%%%%%%%%%%%
\subsection{IMAGINE Lab (\textcolor{blue}{Fig.~\ref{fig:appendix_team_figures1}}-(c))}
\textbf{Contributions}
The IMAGINE Lab team proposes a framework that emphasizes concept-level interpretability. Their primary contribution is the integration of multiple pathology foundation models (CONCH, TITAN, and GECKO) to capture multi-scale domain information, ranging from patch-level morphology to slide-level concepts. Additionally, they implement a vision-concept contrastive learning strategy to align visual features with predefined clinical concepts, ensuring the model's reasoning is grounded in expert-defined pathology.

\textbf{Method} The architecture is inspired by WsiCaption, employing Transformer-based encoder-decoder model with a CNN-based position-aware module to extract information. The pipeline begins by extracting embeddings from frozen CONCH, TITAN, and GECKO encoders, which are then projected into a common multimodal space via MLP adapters and mixers. They utilize GECKO-derived concept priors, where cosine similarity between visual and textual embeddings is used to pretrain a Multiple Instance Learning (MIL) aggregator. During inference, the team utilizes multiple iterations under perturbations and collates the results using maximum agreement logic to ensure the stability and clinical accuracy of the final report.

%%%%%%%%%%%%%%%%%%%%%%%%%%%%%%%%%%%%%%%%%%%%
\subsection{ADCT (\textcolor{blue}{Fig.~\ref{fig:appendix_team_figures2}}-(d))}
\textbf{Contributions} The ADCT team introduces a framework for structured pathology report generation, emphasizing data quality and organ-specific diagnostic standardization. Their primary contribution is the development of comprehensive pathology codebooks that map visual features to structured clinical labels, such as tumor grade and invasion depth. They also implement an edge-based filtering to ensure the high quality of training patches.

\textbf{Method} The proposed method utilizes an attention-based Multiple Instance Learning (ABMIL) framework for slide-level representation. For all organs except prostate, features are extracted using the CONCH encoder at 20$\times$ magnification, while a domain-specific HistoEncoder is employed at 10$\times$ for prostate-specific tasks. The architecture incorporates two distinct reporting strategies: a structured MIL-codebook matching approach for efficient outputs and a full fine-tuning of the Flan-T5 model to translate predicted labels into coherent reports.

%%%%%%%%%%%%%%%%%%%%%%%%%%%%%%%%%%%%%%%%%%%%
\subsection{IUCompPath (\textcolor{blue}{Fig.~\ref{fig:appendix_team_figures2}}-(e))}
\textbf{Contributions} The IUCompPath team adpots the HistGen architecture, which employs a Local-Global Hierarchical (LGH) encoder and a Cross-modal Context (CMC) module, and integrating the pre-trained UNI2 and H-Optimus-1 foundation model to extract slide-level features.

\textbf{Method} The pipeline starts with feature extraction via UNI2 and H-Optimus-1. The features are processed through the LGH encoder, which uses local and global Transformers to integrate fine-grained patch details with broad slide-level context. The CMC module facilitates interaction between visual and textual embeddings to ensure multimodal alignment. The final report is produced by a 3-layer Transformer decoder using beam search, based on the features from LGH and CMC.
%%%%%%%%%%%%%%%%%%%%%%%%%%%%%%%%%%%%%%%%%%%%
\subsection{Team Tiger@REG2025 (\textcolor{blue}{Fig.~\ref{fig:appendix_team_figures2}}-(f))}
\textbf{Contributions} TeamTiger proposes a framework based on the WsiCaption architecture, where they replaced the feature extractor with the UNI2 foundation model. They also demonstrated the feasibility of using general-purpose pathology foundation models for generating reports on diverse Pan-Asia datasets.

\textbf{Method} The pipeline begins by extracting non-overlapping patches from WSIs and generating patch-level embeddings using the pretrained UNI2 model. These embeddings are processed through a Transformer encoder equipped with hierarchical Position-Aware Modules (PAMs) to capture spatial context. The final report is produced by a Transformer decoder.

%%%%%%%%%%%%%%%%%%%%%%%%%%%%%%%%%%%%%%%%%%%%
\subsection{MedInsight - ViseurAI (\textcolor{blue}{Fig.~\ref{fig:appendix_team_figures2}}-(g))}
\textbf{Contributions} The MedInsight-ViseurAI team proposed a transformer-based pipeline for pathology report generation. They utilize the UNI encoder for WSI feature extraction and implement a post-processing stage using Sentence-BERT. This stage refines predictions based on their cosine similarity to ground-truth reports to enhance accuracy.

\textbf{Method} The selected patches using a fine-tuned UNI encoder are projected into memory vectors and fed into a 6-layer Transformer decoder aligned with a BioGPT language model. During the final inference stage, Sentence-BERT compares the generated predictions against training data reports to refine the results based on semantic similarity. 

%%%%%%%%%%%%%%%%%%%%%%%%%%%%%%%%%%%%%%%%%%%%
\subsection{PathoMozhi (\textcolor{blue}{Fig.~\ref{fig:appendix_team_figures2}}-(h))}
\textbf{Contributions}
The PathoMozhi team introduces a Flamingo-style vision-language model tailored for pathology report generation. Their primary contribution is the adaptation of a gated cross-attention mechanism that enables a frozen language model to effectively ingest contextual visual features from WSIs. They also implement an multiple instance learning (MIL) strategy that utilizes a bag of 128 sampled patch embeddings per slide. 

\textbf{Method}
The framework utilizes a frozen BioGPT backbone, interfaced with a vision encoder through learnable gated cross-attention layers after each of the 48 transformer blocks. For visual input, patch-level features are first extracted using a CONCH encoder. However, instead of processing the entire WSI, the model randomly samples a bag of 128 instances for each slide during training. These sampled embeddings are then used as visual context tokens, where the gated cross attention mechanism modulates the flow of histological information into the language model. 
%%%%%%%%%%%%%%%%%%%%%%%%%%%%%%%%%%%%%%%%%%%%
\subsection{MTS\_REG\_2025 (\textcolor{blue}{Fig.~\ref{fig:appendix_team_figures3}}-(j))}
\textbf{Contributions}
The MTS REG 2025 team introduces a Retrieval-Augmented Generation (RAG) approach to produce standardized pathology reports from WSIs. Their primary contributions include an adaptive retrieval mechanism that dynamically adjusts the number of reference cases based on similarity distances and a semantic similarity-based training strategy for the projection layer, which is trained via contrastive learning.

\begin{figure*}[t]
	\centering
    \caption{\textbf{Overview of the model architectures from teams ADCT, IUCompPath, TeamTiger@REG2025, MedInsight-ViseurAI, PathoMozhi, and NW-TIA.} ADCT utilizes a MIL-based processing pipeline with a FLAN-T5 encoder-decoder, while IUCompPath employs a local-global hierarchical encoder with a cross-modal context module. TeamTiger@REG2025 incorporates a position-aware module within a Transformer encoder-decoder, and MedInsight-ViseurAI implements patch selection via a memory vector integrated with BioGPT. The PathoMozhi architecture features 48 interleaved gated cross-attention layers for continuous visual feature injection into a BioGPT-Large backbone, and NW-TIA utilizes a prompt-based approach with auxiliary diagnostic heads for organ and finding prediction through BioBART.}
	\label{fig:appendix_team_figures2}
	\includegraphics[width=.9\textwidth]{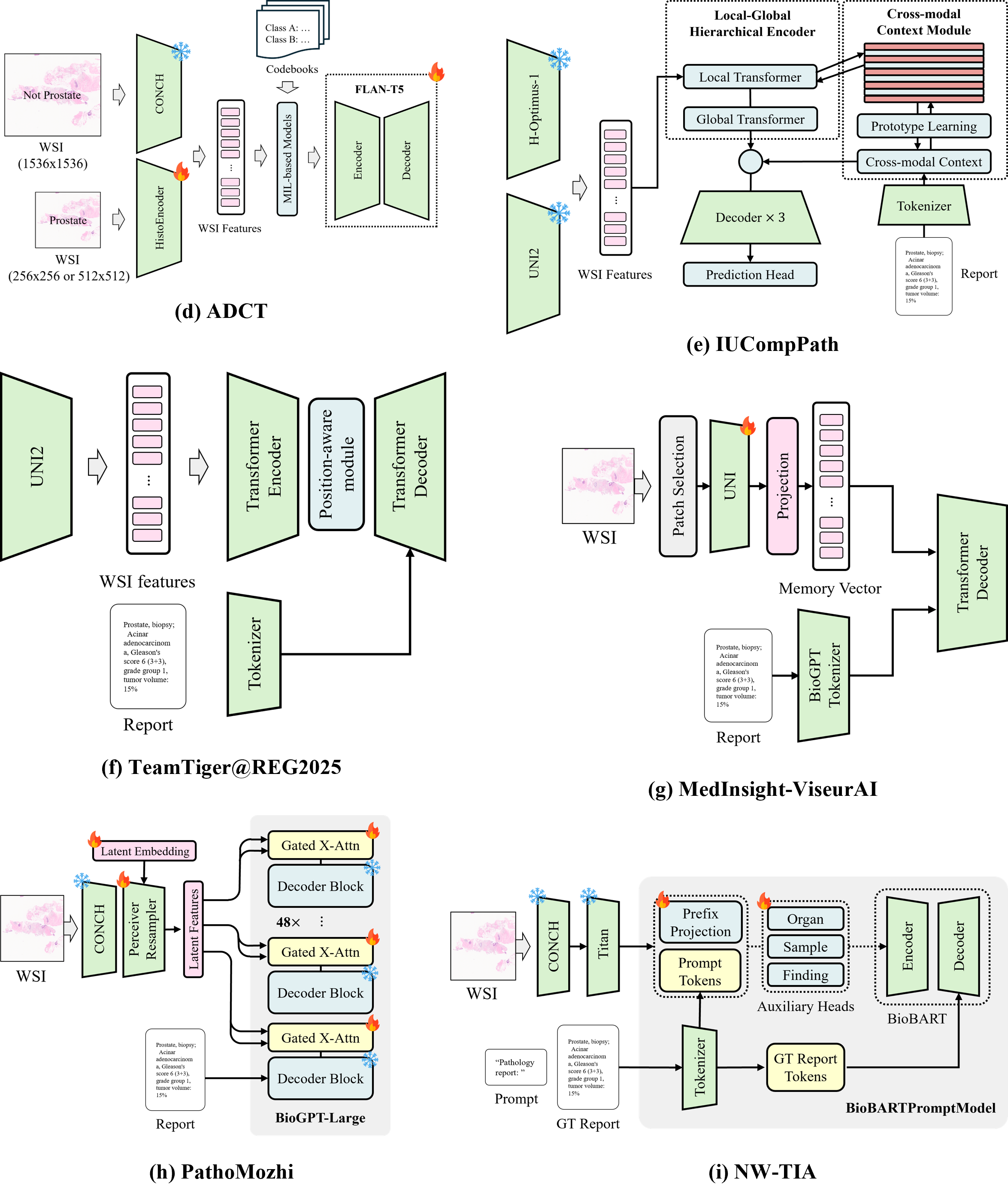}
\end{figure*}

\textbf{Method}
The system utilizes a pretrained Prov-GigaPath encoder to extract slide-level features, which are then transformed into a normalized embedding space via a trained projection head. It uses a “dynamic k-retrieval” process, where retrieval stops if the similarity gap between consecutive cases exceeds a specific threshold, preventing the inclusion of irrelevant data. These retrieved reports are fed into a QEN-3 8B reasoning model, which synthesizes the final report using a sequential majority-voting strategy to select organs, procedures, and histologic types. This approach ensures that the output strictly. adheres to established diagnostic conventions through a structured ensemble of the most similar historical cases. 
%%%%%%%%%%%%%%%%%%%%%%%%%%%%%%%%%%%%%%%%%%%%
\subsection{NW-TIA (\textcolor{blue}{Fig.~\ref{fig:appendix_team_figures2}}-(i))}
\textbf{Contributions}
The NW-TIA team proposes a multi-task, embedding-driven pipeline that leverages WSI-level embeddings for efficient pathology report generation without the need for direct pixel-level supervision. A key contribution is the introduction of auxiliary classification heads for organ type, sample type, and findings, which regularize the training process and enhance the clinical grounding of the generated reports.

\textbf{Method}
The framework utilizes a frozen BioBART backbone conditioned on slide-level embeddings extracted via CONCH and Titan encoder. This team utilizes a multi-task learning strategy; while the decoder generates text, specialized auxiliary MLP heads simultaneously predict the organ, specimen type, and 75 distinct findings from the shared WSI representation. These auxiliary tasks are weighted and backpropagated to force the visual projector to capture essential diagnostic features. During inference, the auxiliary heads are removed, leaving a streamlined generator informed by a clinically-enriched embedding space. 

\begin{figure*}[t]
    \centering
    \caption{\textbf{Overview of architectures from teams MTS\_REG\_2025 and REG\_Path.} MTS\_REG\_2025 employs a contrastive learning retrieval system with Prov-GigaPath and a dynamic $k$-retrieval mechanism for report synthesis via QWEN-3 8B. REG\_Path utilizes a two-stage pipeline for domain alignment and instruction tuning, leveraging the slideChat framework with CONCH encoders for generative report production.}
    \label{fig:appendix_team_figures3}
    \includegraphics[width=.9\textwidth]{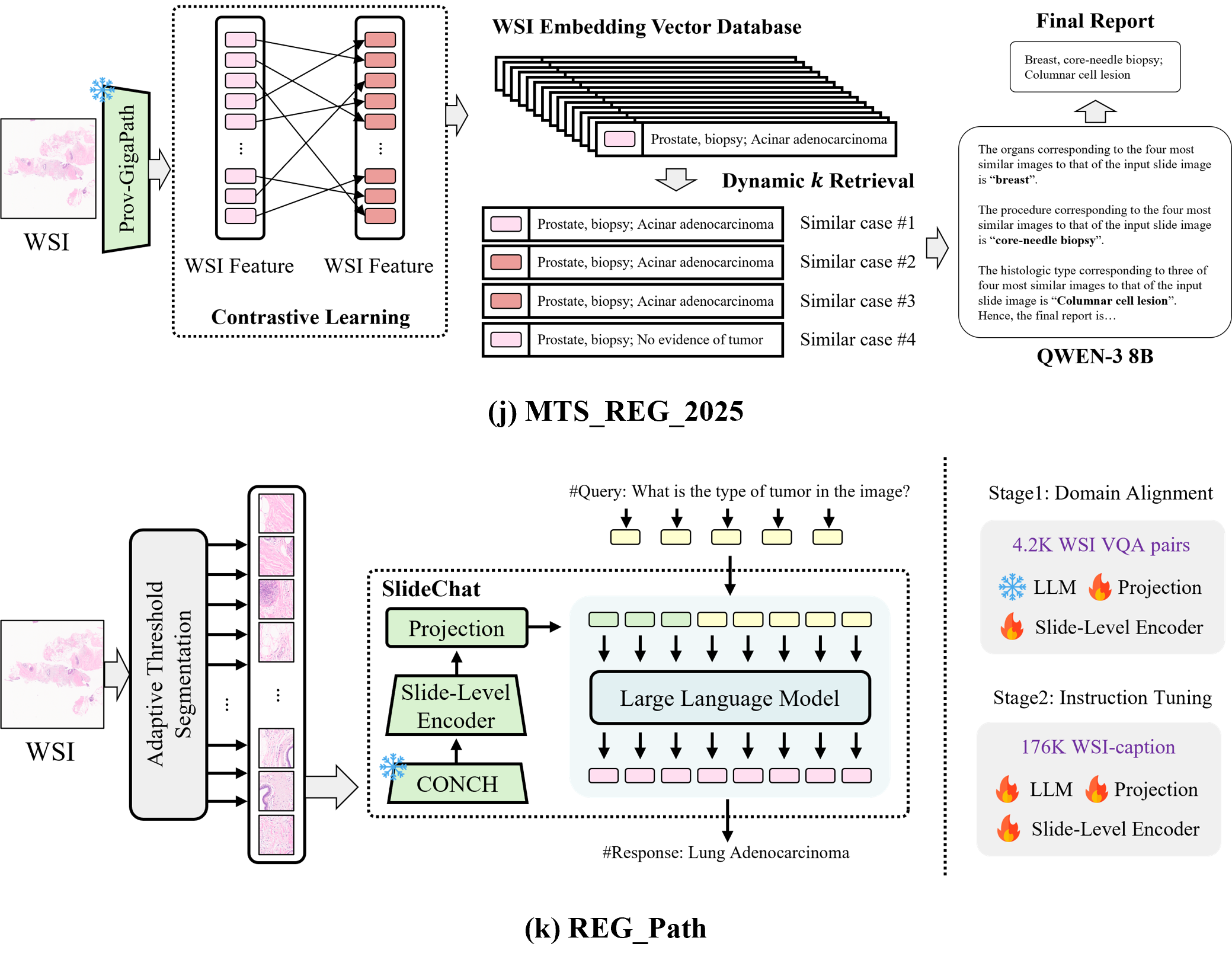}
\end{figure*}

%%%%%%%%%%%%%%%%%%%%%%%%%%%%%%%%%%%%%%%%%%%%
\subsection{REG\_Path (\textcolor{blue}{Fig.~\ref{fig:appendix_team_figures3}}-(k))}
\textbf{Contributions}
The REG Path team proposes a threshold-adaptive preprocessing mechanism that dynamically adjusts segmentation thresholds based on individual slide statistical distributions to ensure more accurate feature extraction. They also introduce a two-stage fine-tuning framework to optimize the model for domain-specific grounding through VQA and clinical coherence in full report generation.

\textbf{Method}
Built upon the SlideChat backbone, the model utilizes a pretrained CONCH encoder to extract patch-level features from WSIs using adatpively tuned segmenation thresholds. The training follows a two-stage pipeline: first, a domain alignment phase using 4.2K WSI-VQA pairs to establish visual-textual grounding, followed by an instruction tuning phase on 176K WSI-caption pairs for end-to-end report generation.

\section{Challenge Organization Details}
In accordance with the BIAS (Biomedical Image Analysis Challenges) statement~\cite{maierhein2020bias} reporting guidelines, we provide additional details regarding the challenge organization that were not fully described in the main manuscript.

\subsection{Challenge Timeline}
The challenge was conducted according to the following timeline:
\begin{enumerate}[\textbullet]
    \item Training data release: May 20, 2025
    \item Registration deadline: July 4, 2025
    \item Debug Phase opens: July 4, 2025
    \item Debug Phase deadline: July 10, 2025
    \item Test Phase 1 opens and Test Phase 1 data release: July 11, 2025
    \item Test Phase 1 deadline: August 1, 2025
    \item Test Phase 2 opens and Test Phase 2 data release: August 2, 2025
    \item Test Phase 2 deadline: August 23, 2025
    \item Announcement of winners: September 9, 2025
\end{enumerate}

\subsection{Participation Policy}
Multiple registrations from the same participant or institution using different accounts were not permitted. This policy was introduced to prevent participants from circumventing the submission limit of two submissions per phase. To ensure fairness and avoid potential conflicts of interest, members affiliated with the organizing institutions did not participate in the challenge.

\subsection{Pre-submission Evaluation}
To balance fairness and accessibility, each participating team was allowed a maximum of two submissions per phase. Instead of providing unrestricted leaderboard access, the official evaluation code was publicly released, allowing participants to locally evaluate their methods prior to submission. The evaluation code is available at \url{https://github.com/hrb0/reg/tree/main/metric}. Additionally, submissions that did not follow the prescribed JSON format specifications (e.g., missing case IDs or mismatched key names) were not evaluated.

\subsection{Award Policy}
Awards and prize funding were provided to the top five teams achieving the highest performance in the challenge. To verify reproducibility, the selected teams were required to submit their source code for independent validation by the organizers. The submitted source code was used internally for reproducibility verification purposes only and was not publicly released. Final awards were granted only to teams that successfully passed the reproducibility verification process. Prize details were provided on the official challenge website (\url{https://reg2025.grand-challenge.org/prizes/}). The top five teams were additionally invited to present their methods during the Satellite Day 2 session at MICCAI 2025.

\subsection{Data Usage and License Policy}
The REG2025 dataset is released under CC BY-NC-SA 4.0. The anonymized WSI–report dataset may be used, reproduced, and shared for non-commercial research, education, method development, benchmarking, and publication with appropriate attribution. Redistributed or adapted materials must follow the same license. Commercial use, re-identification attempts, and clinical use are prohibited, and data use remains subject to Grand Challenge terms and applicable ethics approvals. Dataset is available at \url{https://reg2025.grand-challenge.org/reg2025/}.

\section{Expertise of Annotators}
The annotation and review process involved pathologists with diverse levels of clinical expertise, including three pathologists with more than 20 years of experience, three with 10–20 years of experience, two with 5–10 years of experience, one with less than 5 years of experience, and four non-board-certified pathology trainees.

\clearpage

\begin{table*}[t]
\centering
\caption{\textbf{IMAGINE Lab: Representative histopathological concepts for qualitative analysis.} Organ-specific morphological patterns curated by the IMAGINE Lab team to interpret model predictions across breast, lung, and prostate tissues.}
\label{tbl:concepts}
\begin{tabularx}{\textwidth}{l|X}
\toprule
\textbf{Organ} & \textbf{Concepts} \\
\midrule
\multirow{2}{*}{Breast}
& ``Cribriform pattern'': Tumor glands forming rounded or sieve-like punched-out spaces (cribriform architecture) \\
& ``Papillary structures'': Epithelial fronds or projections supported by fibrovascular cores (papillary architecture) \\
\midrule
\multirow{2}{*}{Lung}
& ``Lepidic growth'': Tumor cells proliferating along intact alveolar walls (lepidic-predominant pattern) \\
& ``Keratin pearls'': Concentric laminated keratinous material within nests of squamous carcinoma cells \\
\midrule
\multirow{2}{*}{Prostate}
& ``Cribriform glands'': Tumor glands forming large rounded sheets with multiple perforated or punched-out lumina \\
& ``Poorly formed glands'': Small, irregular glandular spaces with scant luminal formation (Gleason pattern 4) \\
\bottomrule

\end{tabularx}
\end{table*}

\begin{figure*}[t]
	\centering
    \caption{\textbf{Overview of the dataset construction workflow.} (a) Institution-specific free-text pathology reports are curated into structured and standardized templates with unified terminology. (b) Multi-institutional WSI acquisition and preprocessing pipeline, including de-identification, quality control, expert review, and exclusion of diagnostically equivocal cases. (c) Final dataset preparation, including multi-specimen slide separation and diagnosis-balanced splitting, with each WSI paired to a curated pathology report.}
	\label{fig:appendix_data_workflow}
	\includegraphics[width=\textwidth]{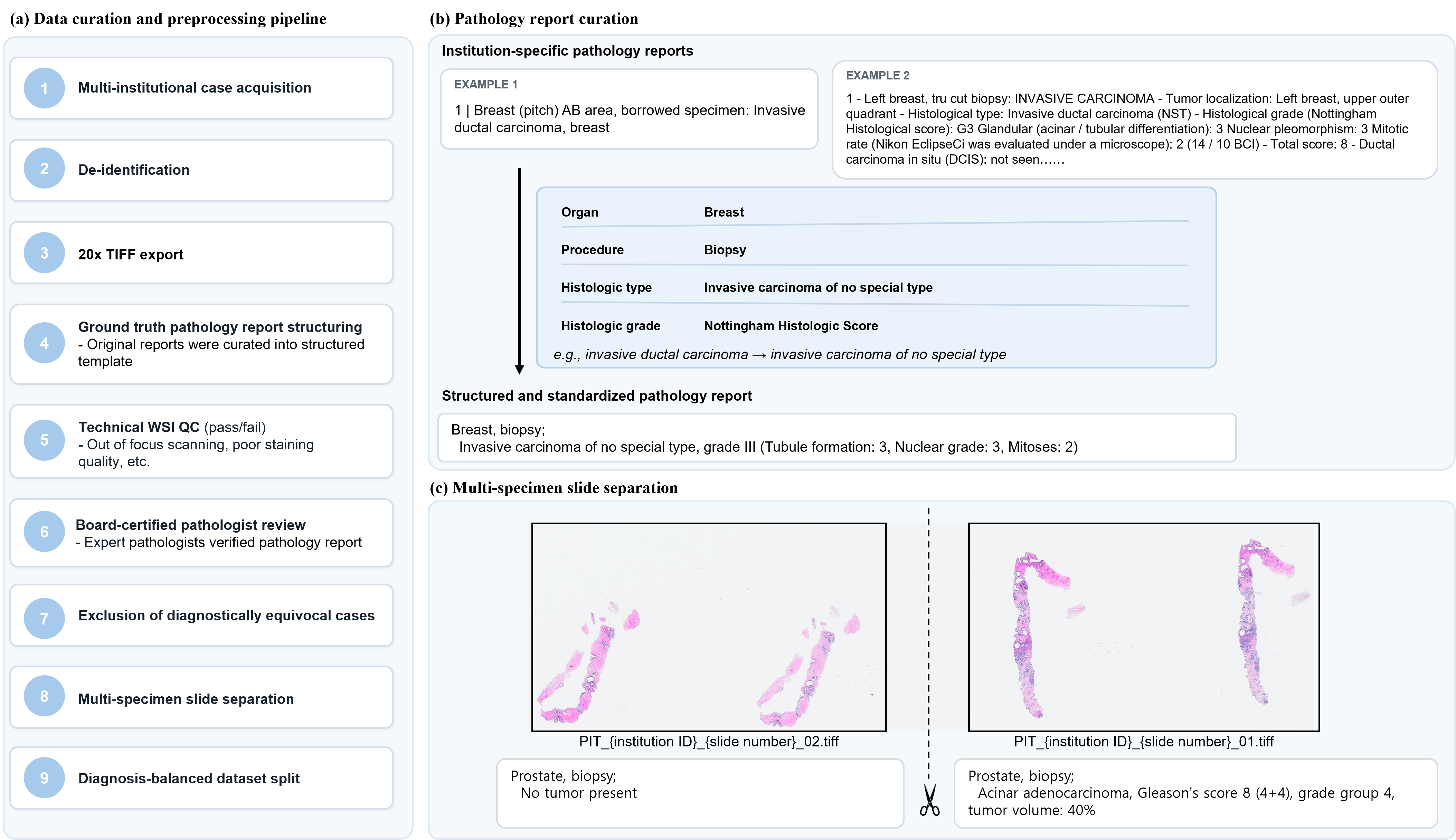}
\end{figure*}

\begin{table*}[t]
\centering
\caption{\textbf{REG2025 challenge organizers.} The challenge consisted of the organizing committee and data providing teams. Detailed information is available on the challenge website (\url{https://reg2025.grand-challenge.org/organizers/}).}
\label{tab:organizers}
\resizebox{\textwidth}{!}{
\begin{tabular}{lll|lll}
\toprule
\textbf{Organizer} & \textbf{Affiliation} & \textbf{Country} & \textbf{Organizer} & \textbf{Affiliation} & \textbf{Country} \\
\hline
Yumi Lee & Ewha Womans University & Korea & Sulen Sarioglu & Memorial Healthcare Group & Turkey \\
Harim Oh & Korea University Anam Hospital & Korea & Serdar Balci & Memorial Healthcare Group & Turkey \\
Junya Fukuoka & Kameda Medical Center & Japan & İlknur Türkmen & Memorial Healthcare Group & Turkey \\
Andrey Bychkov & Kameda Medical Center & Japan & Yuri Tolkach & University Hospital Cologne & Germany \\
Jijgee Munkhdelger & Kameda Medical Center & Japan & Christian Harder & University Hospital Cologne & Germany \\
Rajiv Kumar Kaushal & Tata Memorial Hospital & India & Julian Westerdorf & University Hospital Cologne & Germany \\
Ayushi Sahay & Tata Memorial Hospital & India & Jang-Hwan Choi & Ewha Womans University & Korea \\
Rajni Yadav & AIIMS Delhi & India & Sangjeong Ahn & Korea University Anam Hospital & Korea \\
\bottomrule
\end{tabular}
}
\end{table*}

\clearpage
% \section{Qualitative Results}

% \twocolumn[
%   \begin{@twocolumnfalse}
%     \section{Qualitative Results}
%   \end{@twocolumnfalse}
% ]

%%%%%%%%%%%%%%%%%%%%%% figs/PIT_01_01467_01 %%%%%%%%%%%%%%%%%%%%%
\begin{table*}[!b]
\centering

\caption{\textbf{Attribute-level fidelity in a breast core biopsy.} Among the participating teams, ICGI showed full concordance with the ground truth by correctly generating the principal diagnosis of ductal carcinoma in situ (DCIS) and the key pathologic attributes, including type, nuclear grade, necrosis status, and microcalcification. (\hl{organColor}{\phantom{--}}=organ, \hl{procedureColor}{\phantom{--}}=procedure, \hl{correctColor}{\phantom{--}}=correct, \hl{incorrectColor}{\phantom{--}}=incorrect)}
\label{tab:PIT_01_01467_01}

\scriptsize % \footnotesize
\renewcommand{\arraystretch}{1.6}
\setlength{\tabcolsep}{5pt}
\begin{tabularx}{\textwidth}{|Y|Y|Y|Y|}
\hline

% WSI image
\multicolumn{4}{|l|}{\textbf{Whole Slide Image}} \\ \hline
\multicolumn{4}{|c|}{
    \rule{0pt}{15pt} 
    \includegraphics[width=0.8 \textwidth]{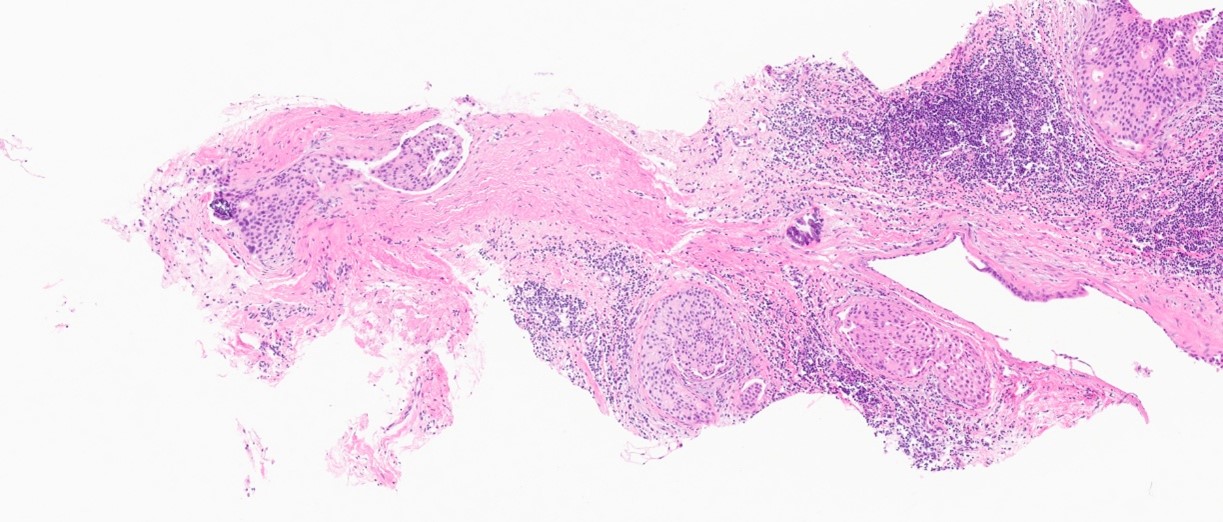} 
    \rule{0pt}{15pt} 
} \\ \hline

\multicolumn{2}{|l|}{\textbf{Ground Truth}} & \textbf{ADCT (0.7115)} & \textbf{ICGI (1.000)} \\ \hline

% Ground Truth
\multicolumn{2}{|p{0.45\textwidth}|}{
    \hl{organColor}{Breast}, \hl{procedureColor}{sono-guided core biopsy;} \newline
    \tab 1. Ductal carcinoma in situ \newline
    \tab - Type: Solid \newline
    \tab - Nuclear grade: Intermediate \newline
    \tab - Necrosis: Absent \newline
    \tab 2. Microcalcification
} & 

    % ADCT
    \hl{correctColor}{Breast, core-needle biopsy;} \newline
    \tab \hl{correctColor}{1. Ductal carcinoma in situ} \newline
    \tab \hl{correctColor}{- Type:} \hl{incorrectColor}{Cribriform} \newline
    \tab \hl{correctColor}{- Nuclear grade: Intermediate} \newline
    \tab \hl{correctColor}{- Necrosis:} \hl{incorrectColor}{Present (Focal)} \newline
    \tab \hl{correctColor}{2. Microcalcification} & 

    % ICGI
    \hl{correctColor}{Breast, sono-guided core biopsy;} \newline
    \tab \hl{correctColor}{1. Ductal carcinoma in situ} \newline
    \tab \hl{correctColor}{- Type: Solid} \newline
    \tab \hl{correctColor}{- Nuclear grade: Intermediate} \newline
    \tab \hl{correctColor}{- Necrosis: Absent} \newline
    \tab \hl{correctColor}{2. Microcalcification} \\ \hline

\textbf{ICL\_PathReport (0.5778)} & \textbf{IMAGINE\_Lab (0.5493)} & \textbf{IUCompPath (0.2639)} & \textbf{MTS\_REG\_2025 (0.2566)} \\ \hline

% ICL_PathReport
\hl{correctColor}{Breast, core-needle biopsy;} \newline
\tab \hl{correctColor}{Ductal carcinoma in situ} \newline
\tab \hl{correctColor}{- Type:} \hl{incorrectColor}{Cribriform, Micropapillary} \newline
\tab \hl{correctColor}{- Nuclear grade:} \hl{incorrectColor}{Low} \newline
\tab \hl{correctColor}{- Necrosis:} \hl{incorrectColor}{Present} & 

% IMAGINE_Lab
\hl{correctColor}{breast, biopsy;} \newline
\tab \hl{correctColor}{ductal carcinoma in situ} \newline
\tab \hl{correctColor}{- type:} \hl{incorrectColor}{cribriform} \newline
\tab \hl{correctColor}{cribriform} \hl{incorrectColor}{high} \newline
\tab \hl{correctColor}{- necrosis:} \hl{incorrectColor}{present (comedo-type)} &

% IUCompPath
\hl{incorrectColor}{Lung, biopsy;} \newline
\tab \hl{incorrectColor}{Non-small cell carcinoma,} \newline
\tab \hl{incorrectColor}{favor adenocarcinoma} & 

% MTS_REG_2025
\hl{incorrectColor}{Lung, biopsy;} \newline
\tab \hl{incorrectColor}{Adenocarcinoma} \\ \hline

\textbf{MedInsight-ViseurAI (0.6456)} & \textbf{REG\_Path (0.4237)} & \textbf{TeamTiger@REG2025 (0.2566)} & \textbf{PathoMozhi (0.5248)} \\ \hline

% MedInsight-ViseurAI
\hl{correctColor}{Breast, sono-guided core biopsy;} \newline
\tab \hl{correctColor}{Ductal carcinoma in situ} \newline
\tab \hl{correctColor}{- Type:} \hl{incorrectColor}{Solid} \newline
\tab \hl{correctColor}{- Nuclear grade:} \hl{incorrectColor}{High} \newline
\tab \hl{correctColor}{- Necrosis:} \hl{incorrectColor}{Present (Comedo-type)} & 

% REG_Path
\hl{correctColor}{Breast, sono-guided core biopsy;} \newline
\tab \hl{incorrectColor}{Papillary neoplasm} & 

% TeamTiger@REG2025
\hl{incorrectColor}{Lung, biopsy;} \newline
\tab \hl{incorrectColor}{Adenocarcinoma} & 

% PathoMozhi
\hl{incorrectColor}{Structured Summary:} \newline
\hl{incorrectColor}{Organ -} \hl{correctColor}{Breast,} \hl{incorrectColor}{Diagnosis -}  \newline
\hl{correctColor}{Ductal carcinoma in situ} 
\hl{correctColor}{- Type: Solid} \hl{incorrectColor}{and cribriform}
\hl{correctColor}{- Nuclear grade: Intermediate} 
\hl{incorrectColor}{- Necrosis: Present (Comedo-type)} \\ \hline
\end{tabularx}

\vspace{5mm}

\end{table*}

%%%%%%%%%%%%%%%%%%%%%% figs/PIT_01_05697_01 %%%%%%%%%%%%%%%%%%%%%
\begin{table*}[htbp]
\centering

\caption{\textbf{Representative whole-slide image and generated reports for a prostate core biopsy.} (A) The \textcolor{red}{red} annotation marks the tumor-involved portion of the biopsied core, corresponding to an estimated tumor volume of ~50\% of the total core length. (B) \textcolor{blue}{Blue} indicates Gleason pattern 3 and \textcolor{orange}{orange} indicates Gleason pattern 4. Pattern 3 is dominant and pattern 4 accounts for ~30\%, resulting in acinar adenocarcinoma, Gleason score 7 (3+4), grade group 2. Across teams, the generated reports were generally concordant for Gleason pattern/score, whereas tumor volume was frequently misestimated (often overcalled as ~80–95\%). (\hl{correctColor}{\phantom{--}}=correct, \hl{incorrectColor}{\phantom{--}}=incorrect)}

\label{tab:PIT_01_05697_01}

\scriptsize % \footnotesize
\renewcommand{\arraystretch}{1.6}
\setlength{\tabcolsep}{5pt}

\begin{tabularx}{\textwidth}{|Y|Y|Y|Y|}
\hline

% WSI image
\multicolumn{4}{|l|}{\textbf{Whole Slide Image}} \\ \hline
\multicolumn{4}{|c|}{
    \rule{0pt}{15pt} 
    \includegraphics[width=0.8 \textwidth]{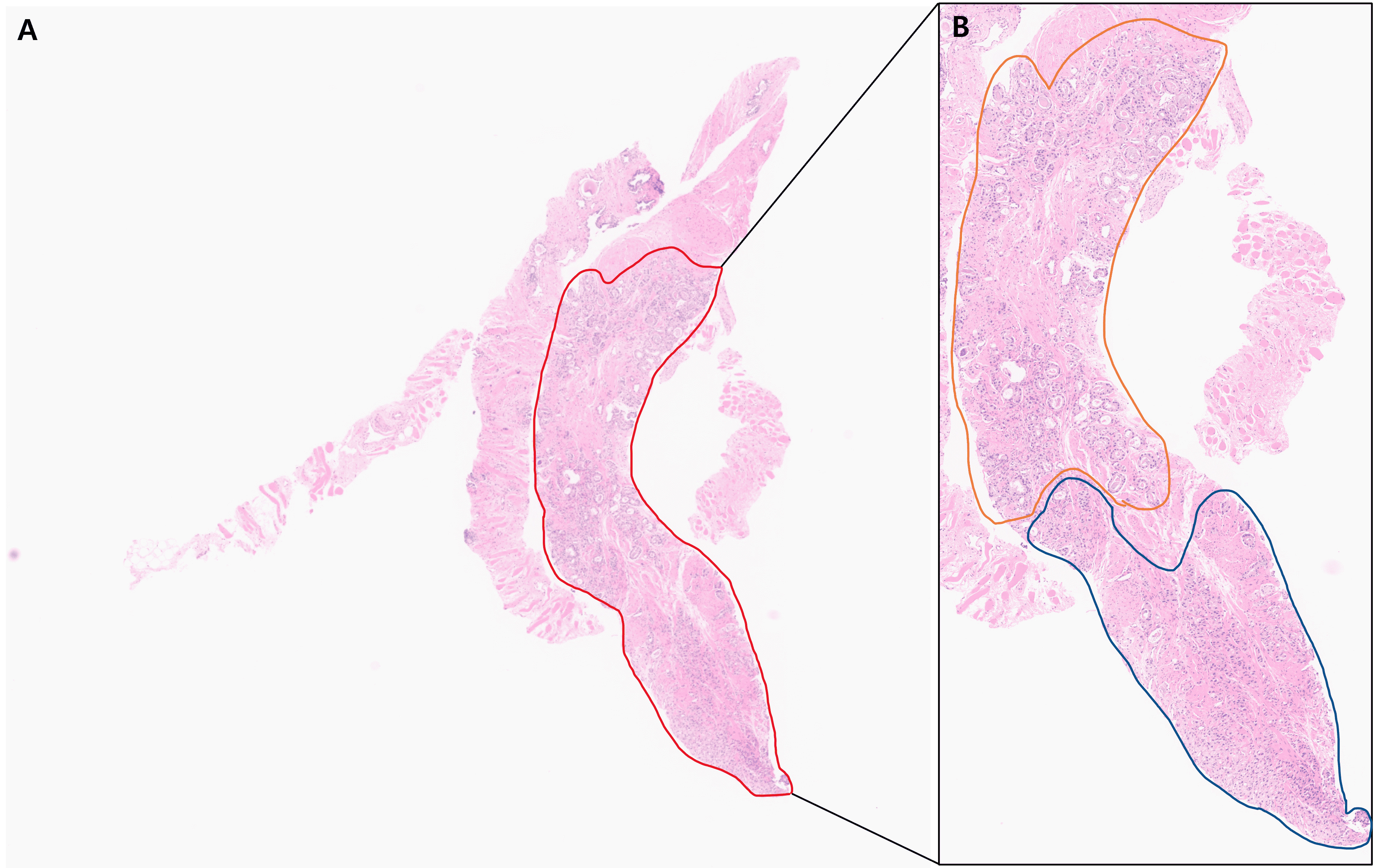} 
    \rule{0pt}{15pt} 
} \\ \hline

\multicolumn{2}{|l|}{\textbf{Ground Truth}} & \textbf{ADCT (0.9329)} & \textbf{ICGI (0.9814)} \\ \hline

    % Ground Truth
    \multicolumn{2}{|p{0.45\textwidth}|}{
        Prostate, biopsy; \newline
        \tab Acinar adenocarcinoma,  \newline
        \tab Gleason's score 7 (3+4), \newline
        \tab grade group 2 (Gleason pattern 4: 30\%), \newline
        \tab tumor volume: 50\%
    } & 

    % ADCT
    \hl{correctColor}{Prostate, biopsy;} \newline
    \hl{correctColor}{Acinar adenocarcinoma,} \newline
    \hl{correctColor}{Gleason's score 7} \hl{incorrectColor}{(3+3),} \newline
    \hl{correctColor}{grade group 2} \newline
    \hl{correctColor}{(Gleason pattern 4: 30\%),} \newline
    \hl{correctColor}{tumor volume:} \hl{incorrectColor}{90\%}\newline
    & 

    % ICGI
    \hl{correctColor}{Prostate, biopsy;} \newline
    \hl{correctColor}{Acinar adenocarcinoma,} \newline
    \hl{correctColor}{Gleason's score 7 (3+4),} \newline
    \hl{correctColor}{grade group 2} \newline
    \hl{correctColor}{(Gleason pattern 4: 30\%), } \newline
    \hl{correctColor}{tumor volume:} \hl{incorrectColor}{90\%}\newline
      \\ \hline

\textbf{ICL\_PathReport (0.6175)} & \textbf{IMAGINE\_Lab (0.7130)} & \textbf{IUCompPath (0.7560)} & \textbf{MTS\_REG\_2025 (0.2817)} \\ \hline

% ICL_PathReport
    \hl{correctColor}{Prostate, biopsy;} \newline
    \hl{correctColor}{Acinar adenocarcinoma} \newline
    \hl{correctColor}{Gleason's score 7,} \newline
    \hl{correctColor}{grade group 2,} \newline
    \hl{correctColor}{tumor volume:} \hl{incorrectColor}{90\%}\newline
    &

% IMAGINE_Lab
\hl{correctColor}{prostate, biopsy;} \newline
\tab \hl{correctColor}{acinar adenocarcinoma,} \newline
\tab \hl{correctColor}{gleason's score 7} \hl{incorrectColor}{(4+3),}\newline 
\tab \hl{correctColor}{grade group} \hl{incorrectColor}{3} \newline
\tab \hl{correctColor}{(Gleason pattern 4:} \hl{incorrectColor}{80\%),} \newline
\tab \hl{correctColor}{tumor volume:} \hl{incorrectColor}{90\%} & 

% IUCompPath
\hl{correctColor}{Prostate, biopsy;} \newline
\tab \hl{correctColor}{Acinar adenocarcinoma,} \newline
\tab \hl{correctColor}{Gleason's score 7} \hl{incorrectColor}{(4+3),}\newline 
\tab \hl{correctColor}{grade group} \hl{incorrectColor}{3} \newline
\tab \hl{correctColor}{(Gleason pattern 4:} \hl{incorrectColor}{60\%),} \newline
\tab \hl{correctColor}{tumor volume:} \hl{incorrectColor}{90\%} &

% MTS_REG_2025
\hl{incorrectColor}{Breast, sono-guided core biopsy;} \newline
\tab \hl{incorrectColor}{1. Invasive carcninoma of} \newline
\tab \hl{incorrectColor}{no special type, grade II} \newline
\tab \hl{incorrectColor}{(Tubule formation: 3,} \newline 
\tab \hl{incorrectColor}{Nuclear grade: 2, Mitoses: 1)} \newline
\tab \hl{incorrectColor}{2. Microcalcification} \\ \hline

\textbf{MedInsight-ViseurAI (0.2877)} & \textbf{REG\_Path (0.7559)} & \textbf{TeamTiger@REG2025 (0.9429)} & \textbf{PathoMozhi (0.6891)} \\ \hline

% MedInsight-ViseurAI
\hl{incorrectColor}{Lung,} \hl{correctColor}{biopsy;}\newline
\tab \hl{incorrectColor}{Chronic granulomatous} \newline
\tab \hl{incorrectColor}{inflammation with necrosis} &

% REG_Path
\hl{correctColor}{Prostate, biopsy;} \newline
\tab \hl{correctColor}{Acinar adenocarcinoma,} \newline
\tab \hl{correctColor}{Gleason's score 7} \hl{incorrectColor}{(4+3),}\newline 
\tab \hl{correctColor}{grade group} \hl{incorrectColor}{3} \newline
\tab \hl{correctColor}{(Gleason pattern 4:} \hl{incorrectColor}{70\%),} \newline
\tab \hl{correctColor}{tumor volume:} \hl{incorrectColor}{90\%} & 

% TeamTiger@REG2025
\hl{correctColor}{Prostate, biopsy;} \newline
\tab \hl{correctColor}{Acinar adenocarcinoma,} \newline
\tab \hl{correctColor}{Gleason's score 7 (3+4),} \newline 
\tab \hl{correctColor}{grade group 2} \newline
\tab \hl{correctColor}{(Gleason pattern 4:} \hl{incorrectColor}{10\%),} \newline
\tab \hl{correctColor}{tumor volume:} \hl{incorrectColor}{90\%} &

% PathoMozhi
\hl{incorrectColor}{Structured Summary:} \newline
\tab \hl{incorrectColor}{Organ -} \hl{correctColor}{Prostate,} \newline
\tab \hl{incorrectColor}{Diagnosis -} \hl{correctColor}{Acinar adenocarcinoma,} \newline 
\tab \hl{correctColor}{Gleason's score 7 (3+4),} \newline 
\tab \hl{correctColor}{grade group 2} \newline
\tab \hl{correctColor}{(Gleason pattern 4: 30\%),} \newline
\tab \hl{correctColor}{tumor volume:} \hl{incorrectColor}{95\%} \\ \hline
\end{tabularx}

\vspace{5mm}

\end{table*}

%%%%%%%%%%%%%%%%%%%%%% figs/PIT_01_04276_01 %%%%%%%%%%%%%%%%%%%%%
\begin{table*}[htbp]
\centering

\caption{\textbf{Representative case of correct identification of colonic malignant lymphoma across teams.} The surface shows normal colonic mucosa, while lymphoma cells infiltrate the underlying tissue. The \textcolor{red}{red} annotation delineates the area involved by lymphoma. Most team-generated reports were concordant for both organ (colon) and diagnosis (malignant lymphoma), with occasional organ-label mismatch (e.g., stomach) despite correct identification of lymphoma. (\hl{correctColor}{\phantom{--}}=correct, \hl{incorrectColor}{\phantom{--}}=incorrect)}
\label{tab:PIT_01_04276_01}

\scriptsize % \footnotesize
\renewcommand{\arraystretch}{1.6}
\setlength{\tabcolsep}{5pt}

\begin{tabularx}{\textwidth}{|Y|Y|Y|Y|}
\hline

% WSI image
\multicolumn{4}{|l|}{\textbf{Whole Slide Image}} \\ \hline
\multicolumn{4}{|c|}{
    \rule{0pt}{15pt} 
    \includegraphics[width=0.7 \textwidth]{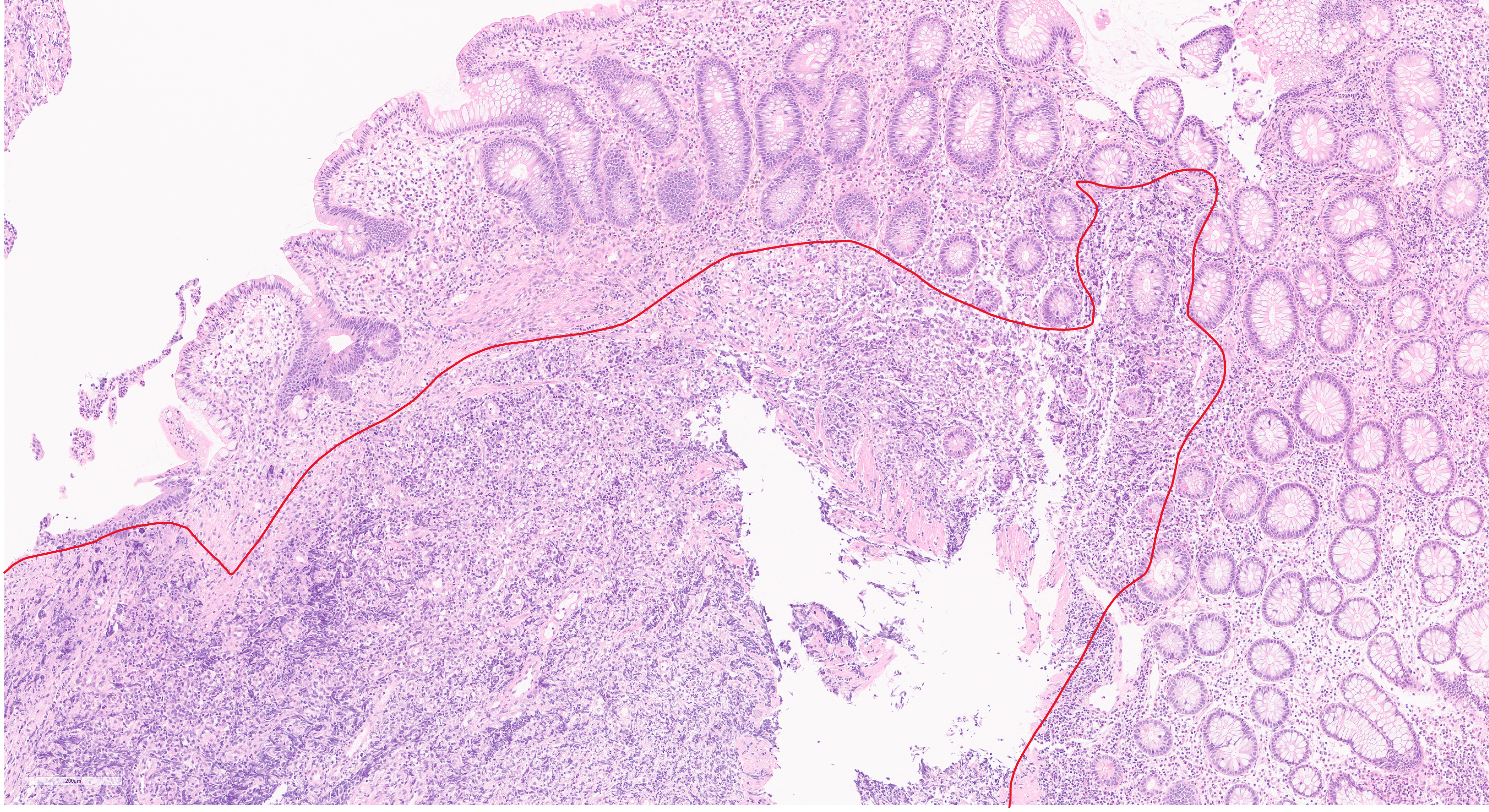} 
    \rule{0pt}{15pt} 
} \\ \hline

\textbf{Ground Truth} & \textbf{ADCT (1.000)} & \textbf{ICGI (1.000)} & \textbf{ICL\_PathReport (1.000)} \\ \hline

% Ground Truth 
Colon, colonoscopic biopsy; \newline
\tab Malignant lymphoma &

% ADCT
\hl{correctColor}{Colon, colonoscopic biopsy;} \newline
\tab \hl{correctColor}{Malignant lymphoma} &

% ICL_PathReport
\hl{correctColor}{Colon, colonoscopic biopsy;} \newline
\tab \hl{correctColor}{Malignant lymphoma} &

% IMAGINE_Lab
\hl{correctColor}{Colon, colonoscopic biopsy;} \newline
\tab \hl{correctColor}{Malignant lymphoma} \\ \hline

\textbf{IMAGINE\_Lab (1.000)} & \textbf{IUCompPath (1.000)} & \textbf{MTS\_REG\_2025 (0.450)} & \textbf{MedInsight-ViseurAI (1.000)} \\ \hline

% IMAGINE_Lab
\hl{correctColor}{Colon, colonoscopic biopsy;} \newline
\tab \hl{correctColor}{Malignant lymphoma} &

% IUCompPath
\hl{correctColor}{Colon, colonoscopic biopsy;} \newline
\tab \hl{correctColor}{Malignant lymphoma} &

% MTS_REG_2025
\hl{incorrectColor}{Stomach,} \hl{correctColor}{endoscopic biopsy;} \newline
\tab \hl{correctColor}{Malignant lymphoma} &

% MedInsight-ViseurAI
\hl{correctColor}{Colon, colonoscopic biopsy;} \newline
\tab \hl{correctColor}{Malignant lymphoma} \\ \hline

\textbf{REG\_Path (1.000)} & \textbf{TeamTiger@REG2025 (1.000)} & \textbf{PathoMozhi (1.000)} & \textbf{NW-TIA (1.000)} \\ \hline

% REG_Path
\hl{correctColor}{Colon, colonoscopic biopsy;} \newline
\tab \hl{correctColor}{Malignant lymphoma} &

% TeamTiger@REG2025
\hl{correctColor}{Colon, colonoscopic biopsy;} \newline
\tab \hl{correctColor}{Malignant lymphoma} &

% PathoMozhi
\hl{correctColor}{Colon, colonoscopic biopsy;} \newline
\tab \hl{correctColor}{Malignant lymphoma} &

% NW-TIA
\hl{correctColor}{Colon, colonoscopic biopsy;} \newline
\tab \hl{correctColor}{Malignant lymphoma} \\ \hline
\end{tabularx}

\vspace{5mm}

\end{table*}

%%%%%%%%%%%%%%%%%%%%%% figs/PIT_01_00193_01 %%%%%%%%%%%%%%%%%%%%%
\begin{table*}[htbp]
\centering

\caption{\textbf{Metaplastic carcinoma mimicking lung squamous cell carcinoma in generated reports.} The \textcolor{red}{red} circles highlight squamous differentiation, supporting the ground truth diagnosis of metaplastic carcinoma. \textbf{ICGI} correctly identified metaplastic carcinoma. Some teams produced “invasive carcinoma of no special type”, which was categorized as indeterminate (organ correct but subtype not specified). Several other teams generated lung squamous cell carcinoma by the squamous morphology—morphologically plausible, yet contextually incorrect. (\hl{correctColor}{\phantom{--}}=correct, \hl{incorrectColor}{\phantom{--}}=incorrect, \hl{ IndeterminateColor}{\phantom{--}}=indeterminate)}
\label{tab:PIT_01_00193_01}

\scriptsize % \footnotesize
\renewcommand{\arraystretch}{1.6}
\setlength{\tabcolsep}{5pt}

\begin{tabularx}{\textwidth}{|Y|Y|Y|Y|}
\hline

% WSI image
\multicolumn{4}{|l|}{\textbf{Whole Slide Image}} \\ \hline
\multicolumn{4}{|c|}{
    \rule{0pt}{15pt} 
    \includegraphics[width=0.7 \textwidth]{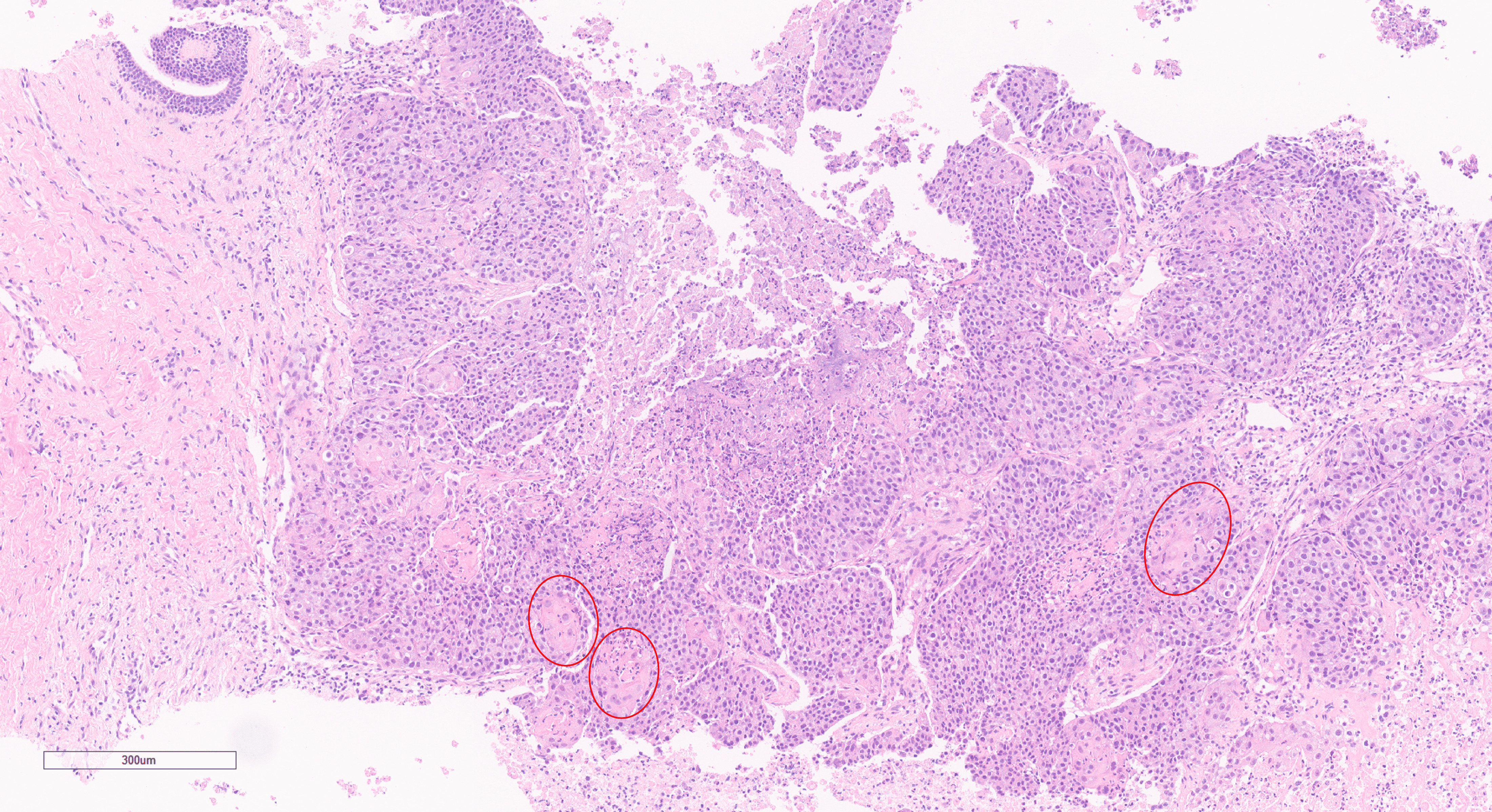} 
    \rule{0pt}{15pt} 
} \\ \hline

\multicolumn{2}{|l|}{\textbf{Ground Truth}} & \textbf{ADCT (0.3373)} & \textbf{ICGI (1.000)} \\ \hline

    % Ground Truth
    \multicolumn{2}{|p{0.45\textwidth}|}{
        Breast, core-needle biopsy; \newline
        \tab Metaplastic carcinoma
    } & 

    % ADCT
    \hl{correctColor}{Breast,} \hl{IndeterminateColor}{sono-guided} \hl{correctColor}{core biospy;} \newline
    \hl{IndeterminateColor}{Invasive carcinoma of no} \newline
    \hl{IndeterminateColor}{special type, grade II}
    \hl{IndeterminateColor}{(Tubule formation: 3,} \newline
    \hl{IndeterminateColor}{Nuclear grade: 2,} \newline
    \hl{IndeterminateColor}{Mitoses: 1)} \newline
    & 

    % ICGI
    \hl{correctColor}{Breast, core-needle biopsy;} \newline
    \tab \hl{correctColor}{Metaplastic carcinoma}  \\ \hline

\textbf{ICL\_PathReport (0.2980)} & \textbf{IMAGINE\_Lab (0.3071)} & \textbf{IUCompPath (0.2845)} & \textbf{MTS\_REG\_2025 (0.3207)} \\ \hline

% ICL_PathReport
    \hl{incorrectColor}{Lung, biopsy;} \newline
    \hl{incorrectColor}{Non-small cell carcinoma,} \newline
    \hl{incorrectColor}{not otherwise specified} \newline
    & 

% IMAGINE_Lab
\hl{correctColor}{lung, biopsy;} \newline
\tab \hl{correctColor}{squamous cell carcinoma}\newline 
& 

% IUCompPath
\hl{incorrectColor}{Lung, biopsy;} \newline
\tab \hl{incorrectColor}{Non-small cell carcinoma,} \newline
\tab \hl{incorrectColor}{favor adenocarcinoma}\newline 
 &

% MTS_REG_2025
\hl{incorrectColor}{Lung, biopsy;} \newline
\tab \hl{incorrectColor}{Squamous cell carcinoma} \\ \hline

\textbf{MedInsight-ViseurAI (0.4090)} & \textbf{REG\_Path (0.3207)} & \textbf{TeamTiger@REG2025 (0.2982)} & \textbf{PathoMozhi (0.3000)} \\ \hline

% MedInsight-ViseurAI
\hl{correctColor}{Breast, core-needle biopsy;}\newline
\tab \hl{IndeterminateColor}{Invasive carcinoma of no} \newline
\tab \hl{IndeterminateColor}{special type, grade II} \newline
\tab \hl{IndeterminateColor}{(Tubule formation: 3,} \newline
\tab \hl{IndeterminateColor}{Nuclear grade: 2,} \newline 
\tab \hl{IndeterminateColor}{Mitoses: 1)} &

% REG_Path
\hl{incorrectColor}{Lung, biopsy;} \newline
\tab \hl{incorrectColor}{Squamous cell carcinoma} & 

% TeamTiger@REG2025
\hl{incorrectColor}{Lung, biopsy;} \newline
\tab \hl{incorrectColor}{Non-small cell carcinoma,} \newline
\tab \hl{incorrectColor}{favor squamous cell carcinoma} &

% PathoMozhi
\hl{incorrectColor}{Structured Summary:} \newline
\hl{incorrectColor}{Organ -} \hl{correctColor}{Breast,} \newline
\hl{incorrectColor}{Diagnosis -} \hl{IndeterminateColor}{Invasive carcinoma} \newline 
\hl{IndeterminateColor}{of no special type, grade III} \newline
\hl{IndeterminateColor}{(Tubule formation: 3,} \newline
\hl{IndeterminateColor}{Nuclear grade: 3 and Mitoses: 2)}\\ \hline

\end{tabularx}

\vspace{5mm}

\end{table*}

%%%%%%%%%%%%%%%%%%%%%% figs/PIT_01_09080_01 %%%%%%%%%%%%%%%%%%%%%
\begin{table*}[htbp]
\centering

\caption{\textbf{Rare benign entity-limited recognition across teams.} The ground truth diagnosis is fundic gland polyp, a minor category of benign polypoid lesion (only three cases were included in the training set). \textbf{MTS\_REG\_2025} correctly generated “fundic gland polyp”. In contrast, most other teams generated “chronic gastritis,” reflecting substitution of a common, nonspecific diagnosis rather than recognizing the specific polyp entity. (\hl{correctColor}{\phantom{--}}=correct, \hl{incorrectColor}{\phantom{--}}=incorrect)}
\label{tab:PIT_01_09080_01}

\scriptsize % \footnotesize
\renewcommand{\arraystretch}{1.6}
\setlength{\tabcolsep}{5pt}

\begin{tabularx}{\textwidth}{|Y|Y|Y|Y|}
\hline

% WSI image
\multicolumn{4}{|l|}{\textbf{Whole Slide Image}} \\ \hline
\multicolumn{4}{|c|}{
    \rule{0pt}{15pt} 
    \includegraphics[width=0.7 \textwidth]{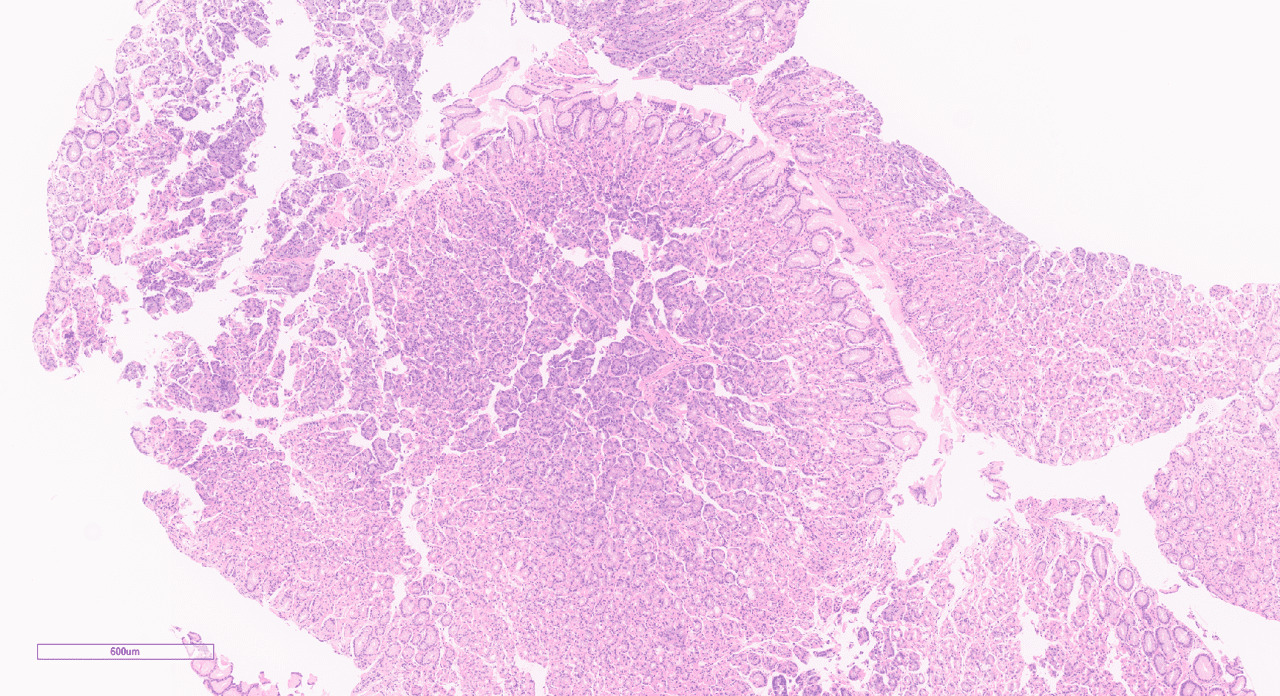} 
    \rule{0pt}{15pt} 
} \\ \hline

\textbf{Ground Truth} & \textbf{ADCT (0.4858)} & \textbf{ICGI (0.5602)} & \textbf{ICL\_PathReport (0.5602)} \\ \hline

% Ground Truth 
Stomach, endoscopic biopsy; \newline
\tab Fundic gland polyp &

% ADCT
\hl{correctColor}{Stomach, endoscopic biopsy;} \newline
\tab \hl{incorrectColor}{Adenocarcinoma, poorly} \newline 
\hl{incorrectColor}{differentiated with poorly}
\hl{incorrectColor}{cohesive carcinoma component} &

% ICGI
\hl{correctColor}{Stomach, endoscopic biopsy;} \newline
\tab \hl{incorrectColor}{Chronic gastritis} &

% ICL_PathReport
\hl{correctColor}{Stomach, endoscopic biopsy;} \newline
\tab \hl{incorrectColor}{Chronic gastritis} \\ \hline

\textbf{IMAGINE\_Lab (0.5602)} & \textbf{IUCompPath (0.5602)} & \textbf{MTS\_REG\_2025 (1.000)} & \textbf{MedInsight-ViseurAI (0.5602)} \\ \hline

% IMAGINE_Lab
\hl{correctColor}{Stomach, endoscopic biopsy;} \newline
\tab \hl{incorrectColor}{Chronic gastritis} &

% IUCompPath
\hl{correctColor}{Stomach, endoscopic biopsy;} \newline
\tab \hl{incorrectColor}{Chronic gastritis} &

% MTS_REG_2025
\hl{correctColor}{Stomach, endoscopic biopsy;} \newline
\tab \hl{correctColor}{Fundic gland polyp} &

% MedInsight-ViseurAI
\hl{correctColor}{Stomach, endoscopic biopsy;} \newline
\tab \hl{incorrectColor}{Chronic gastritis}  \\ \hline

\textbf{REG\_Path (0.5602)} & \textbf{TeamTiger@REG2025 (0.5602)} & \textbf{PathoMozhi (0.5602)} & \textbf{NW-TIA (0.5602)} \\ \hline

% REG_Path
\hl{correctColor}{Stomach, endoscopic biopsy;} \newline
\tab \hl{incorrectColor}{Chronic gastritis} &

% TeamTiger@REG2025
\hl{correctColor}{Stomach, endoscopic biopsy;} \newline
\tab \hl{incorrectColor}{Chronic gastritis} &

% PathoMozhi
\hl{correctColor}{Stomach, endoscopic biopsy;} \newline
\hl{incorrectColor}{Chronic gastritis with lymphoid} \hl{incorrectColor}{aggregates} &

% NW-TIA
\hl{correctColor}{Stomach, endoscopic biopsy;} \newline
\tab \hl{incorrectColor}{Chronic gastritis}  \\ \hline

\end{tabularx}

\vspace{5mm}

\end{table*}

%%%%%%%%%%%%%%%%%%%%%% figs/PIT_01_09100_01 %%%%%%%%%%%%%%%%%%%%%
\begin{table*}[htbp]
\centering

\caption{\textbf{Stomach endoscopic biopsy: malignant lymphoma.} Nine teams correctly generated malignant lymphoma. Two teams instead generated “small cell carcinoma”. Given the phenotypic overlap on H\&E, the model’s output is consistent with \textit{High-Fidelity Histomorphology-Driven Phenotype Recognition.} (\hl{correctColor}{\phantom{--}}=correct, \hl{incorrectColor}{\phantom{--}}=incorrect)}
\label{tab:PIT_01_09100_01}

\scriptsize % \footnotesize
\renewcommand{\arraystretch}{1.6}
\setlength{\tabcolsep}{5pt}

\begin{tabularx}{\textwidth}{|Y|Y|Y|Y|}
\hline

% WSI image
\multicolumn{4}{|l|}{\textbf{Whole Slide Image}} \\ \hline
\multicolumn{4}{|c|}{
    \rule{0pt}{15pt} 
    \includegraphics[width=0.7 \textwidth]{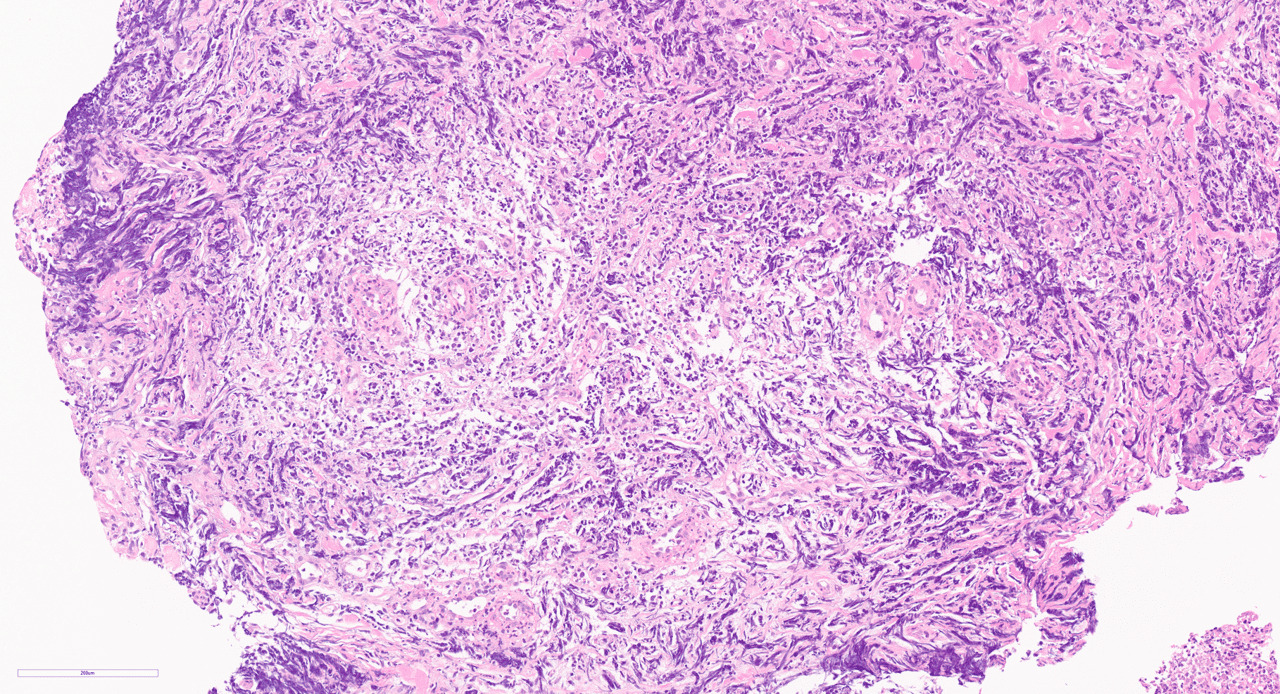} 
    \rule{0pt}{15pt} 
} \\ \hline

\textbf{Ground Truth} & \textbf{ADCT (0.5323)} & \textbf{ICGI (1.000)} & \textbf{ICL\_PathReport (1.000)} \\ \hline

% Ground Truth 
Stomach, endoscopic biopsy;\newline
\tab Malignant lymphoma &

% ADCT
\hl{correctColor}{Stomach, endoscopic biopsy;} \newline
\tab \hl{incorrectColor}{No significant pathologic} \newline 
\tab \hl{incorrectColor}{findings} &

% ICGI
\hl{correctColor}{Stomach, endoscopic biopsy;} \newline
\tab \hl{correctColor}{Malignant lymphoma} &

% ICL_PathReport
\hl{correctColor}{Stomach, endoscopic biopsy;} \newline
\tab \hl{correctColor}{Malignant lymphoma} \\ \hline

\textbf{IMAGINE\_Lab (1.000)} & \textbf{IUCompPath (1.000)} & \textbf{MTS\_REG\_2025 (0.5639)} & \textbf{MedInsight-ViseurAI (1.000)} \\ \hline

% IMAGINE_Lab
\hl{correctColor}{Stomach, endoscopic biopsy;} \newline
\tab \hl{correctColor}{Malignant lymphoma} &

% IUCompPath
\hl{correctColor}{Stomach, endoscopic biopsy;} \newline
\tab \hl{correctColor}{Malignant lymphoma} &

% MTS_REG_2025
\hl{correctColor}{Stomach, endoscopic biopsy;} \newline
\tab \hl{incorrectColor}{Small cell carcinoma} &

% MedInsight-ViseurAI
\hl{correctColor}{Stomach, endoscopic biopsy;} \newline
\tab \hl{correctColor}{Malignant lymphoma}  \\ \hline

\textbf{REG\_Path (1.000)} & \textbf{TeamTiger@REG2025 (1.000)} & \textbf{PathoMozhi (0.5639)} & \textbf{NW-TIA (1.000)} \\ \hline

% REG_Path
\hl{correctColor}{Stomach, endoscopic biopsy;} \newline
\tab \hl{correctColor}{Malignant lymphoma} &

% TeamTiger@REG2025
\hl{correctColor}{Stomach, endoscopic biopsy;} \newline
\tab \hl{correctColor}{Malignant lymphoma} &

% PathoMozhi
\hl{correctColor}{Stomach, endoscopic biopsy;} \newline
\hl{incorrectColor}{Small cell carcinoma} &

% NW-TIA
\hl{correctColor}{Stomach, endoscopic biopsy;} \newline
\tab \hl{correctColor}{Malignant lymphoma}  \\ \hline

\end{tabularx}

\vspace{5mm}

\end{table*}

%%%%%%%%%%%%%%%%%%%%%% figs/PIT_01_03514_01 %%%%%%%%%%%%%%%%%%%%%
\begin{table*}[htbp]
\centering

\caption{\textbf{Preservation of colorectal-type gland-forming morphology despite incorrect site prediction.}  Notably, despite the site mismatch, these outputs consistently captured a colorectal-type, gland-forming phenotype concordant with the suspected primary, illustrating the models’ ability to encode and reproduce diagnostically relevant histomorphologic features. (\textcolor{red}{Red} annotation: colorectal-type tumor portion, \hl{correctColor}{\phantom{--}}=correct, \hl{incorrectColor}{\phantom{--}}=incorrect, \hl{IndeterminateColor}{\phantom{--}}=indeterminate) }
\label{tab:PIT_01_03514_01}

\scriptsize % \footnotesize
\renewcommand{\arraystretch}{1.6}
\setlength{\tabcolsep}{5pt}

\begin{tabularx}{\textwidth}{|Y|Y|Y|Y|}
\hline

% WSI image
\multicolumn{4}{|l|}{\textbf{Whole Slide Image}} \\ \hline
\multicolumn{4}{|c|}{
    \rule{0pt}{15pt} 
    \includegraphics[width=0.6 \textwidth]{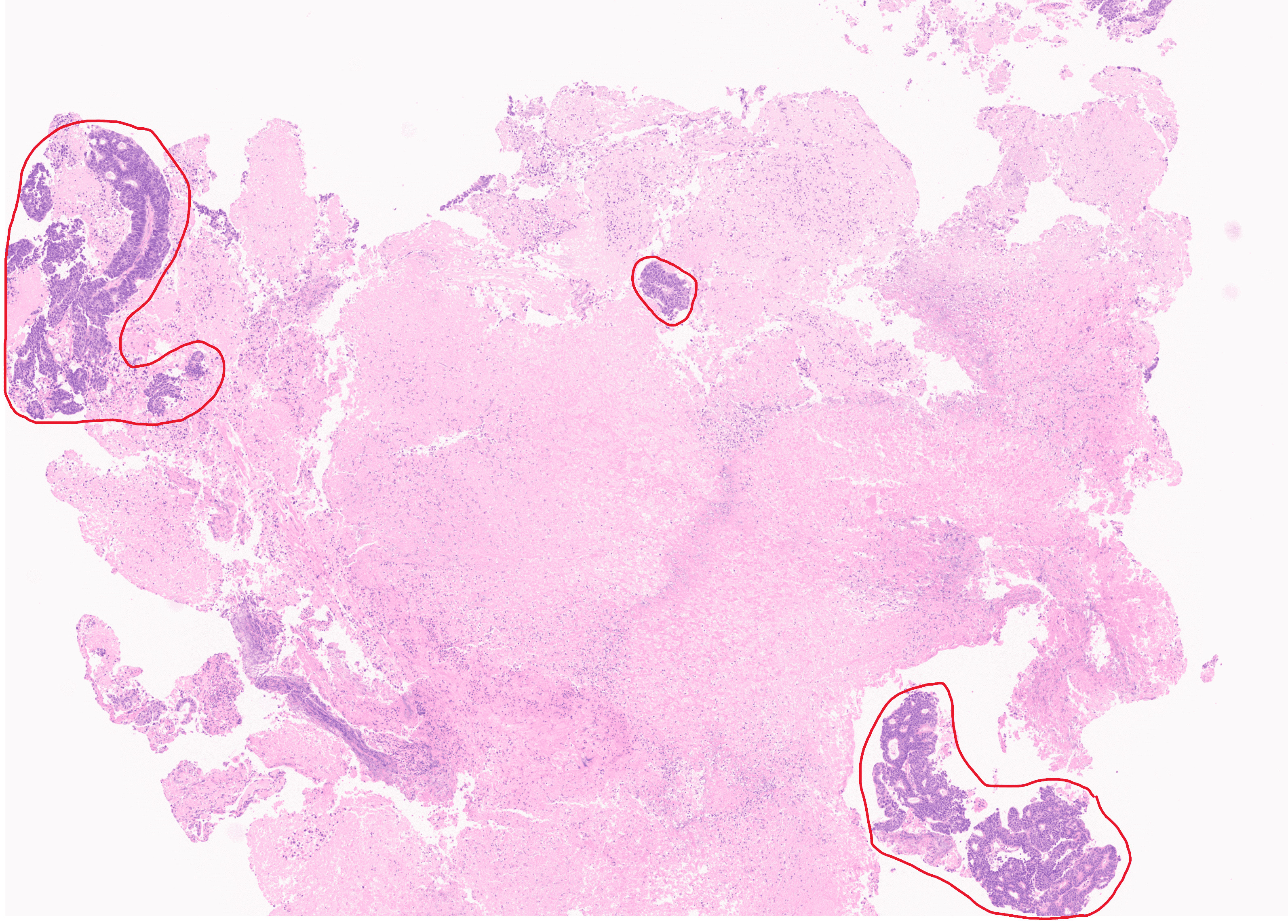} 
    \rule{0pt}{15pt} 
} \\ \hline

\textbf{Ground Truth} & \textbf{ADCT (0.5330)} & \textbf{ICGI (0.3817)} & \textbf{ICL\_PathReport (0.3817)} \\ \hline

% Ground Truth 
Uterine cervix, colposcopic biopsy; \newline
\tab Adenocarcinoma,\newline
\tab favor colorectal primary &

% ADCT
\hl{correctColor}{Uterine cervix, colposcopic biopsy;} \newline
\tab \hl{incorrectColor}{Endocervical adenocarcinoma,} \newline 
\tab \hl{incorrectColor}{HPV-associated, usual type} &

% ICGI
\hl{IndeterminateColor}{Rectum, colonoscopic biopsy;} \newline
\tab \hl{IndeterminateColor}{Adenocarcinoma,} 
\tab \hl{IndeterminateColor}{moderately differentiated} &

% ICL_PathReport
\hl{IndeterminateColor}{Rectum, colonoscopic biopsy;} \newline
\tab \hl{IndeterminateColor}{Adenocarcinoma,} 
\tab \hl{IndeterminateColor}{moderately differentiated} \\ \hline

\textbf{IMAGINE\_Lab (0.3796)} & \textbf{IUCompPath (0.3817)} & \textbf{MTS\_REG\_2025 (0.2741)} & \textbf{MedInsight-ViseurAI (0.3796)} \\ \hline

% IMAGINE_Lab
\hl{IndeterminateColor}{Colon, colonoscopic biopsy;} \newline
\tab \hl{IndeterminateColor}{Adenocarcinoma,} 
\tab \hl{IndeterminateColor}{moderately differentiated} &

% IUCompPath
\hl{IndeterminateColor}{Rectum, colonoscopic biopsy;} \newline
\tab \hl{IndeterminateColor}{Adenocarcinoma,} 
\tab \hl{IndeterminateColor}{moderately differentiated} &

% MTS_REG_2025
\hl{incorrectColor}{Lung, biopsy;} \newline
\tab \hl{incorrectColor}{Squamous cell carcinoma} &

% MedInsight-ViseurAI
\hl{IndeterminateColor}{Colon, colonoscopic biopsy;} \newline
\tab \hl{IndeterminateColor}{Adenocarcinoma,} 
\tab \hl{IndeterminateColor}{moderately differentiated}  \\ \hline

\textbf{REG\_Path (0.3796)} & \textbf{TeamTiger@REG2025 (0.3470)} & \textbf{PathoMozhi (0.3796)} & \textbf{NW-TIA (0.3796)} \\ \hline

% REG_Path
\hl{IndeterminateColor}{Colon, colonoscopic biopsy;} \newline
\tab \hl{IndeterminateColor}{Adenocarcinoma,} 
\tab \hl{IndeterminateColor}{moderately differentiated} &

% TeamTiger@REG2025
\hl{incorrectColor}{Lung, biopsy;} \newline
\tab \hl{incorrectColor}{Adenocarcinoma} &

% PathoMozhi
\hl{IndeterminateColor}{Colon, colonoscopic biopsy;} \newline
\tab \hl{IndeterminateColor}{Adenocarcinoma,} 
\tab \hl{IndeterminateColor}{moderately differentiated} &

% NW-TIA
\hl{IndeterminateColor}{Colon, colonoscopic biopsy;} \newline
\tab \hl{IndeterminateColor}{Adenocarcinoma,} 
\tab \hl{IndeterminateColor}{moderately differentiated}  \\ \hline

\end{tabularx}

\vspace{5mm}

\end{table*}

\clearpage

%%%%%%%%%%%%%%%%%%%%%% figs/PIT_01_00111_01 %%%%%%%%%%%%%%%%%%%%%
\begin{table*}[htbp]
\centering

\caption{\textbf{Fine-grained histologic grading (Nottingham) reproduced in generated pathology reports.} The Nottingham histologic grade is determined by summing three component scores—tubule (gland) formation, nuclear pleomorphism (variation), and mitotic count—each scored from 1 to 3 (total score 3–5 = Grade I, 6–7 = Grade II, 8–9 = Grade III). Tubule formation is graded based on the proportion of glandular structures (<10\% = score 3, 10–75\% = score 2, >75\% = score 1). In this case, the component score were tubule formation=3 (glandular structure <10\%), nuclear grade=3 (marked nuclear variation), and mitotic count = 2 (representative mitotic figures: \textcolor{red}{red} circle), yielding a total score of 8 (Grade III). \textbf{ICGI} and \textbf{ICL\_PathReport} correctly reproduced the diagnosis, the overall Nottingham grade, and the individual component scores. (\hl{correctColor}{\phantom{--}}=correct, \hl{incorrectColor}{\phantom{--}}=incorrect)}
\label{tab:PIT_01_00111_01}

\scriptsize % \footnotesize
\renewcommand{\arraystretch}{1.6}
\setlength{\tabcolsep}{5pt}

\begin{tabularx}{\textwidth}{|Y|Y|Y|Y|}
\hline

% WSI image
\multicolumn{4}{|l|}{\textbf{Whole Slide Image}} \\ \hline
\multicolumn{4}{|c|}{
    \rule{0pt}{15pt} 
    \includegraphics[width=0.8 \textwidth]{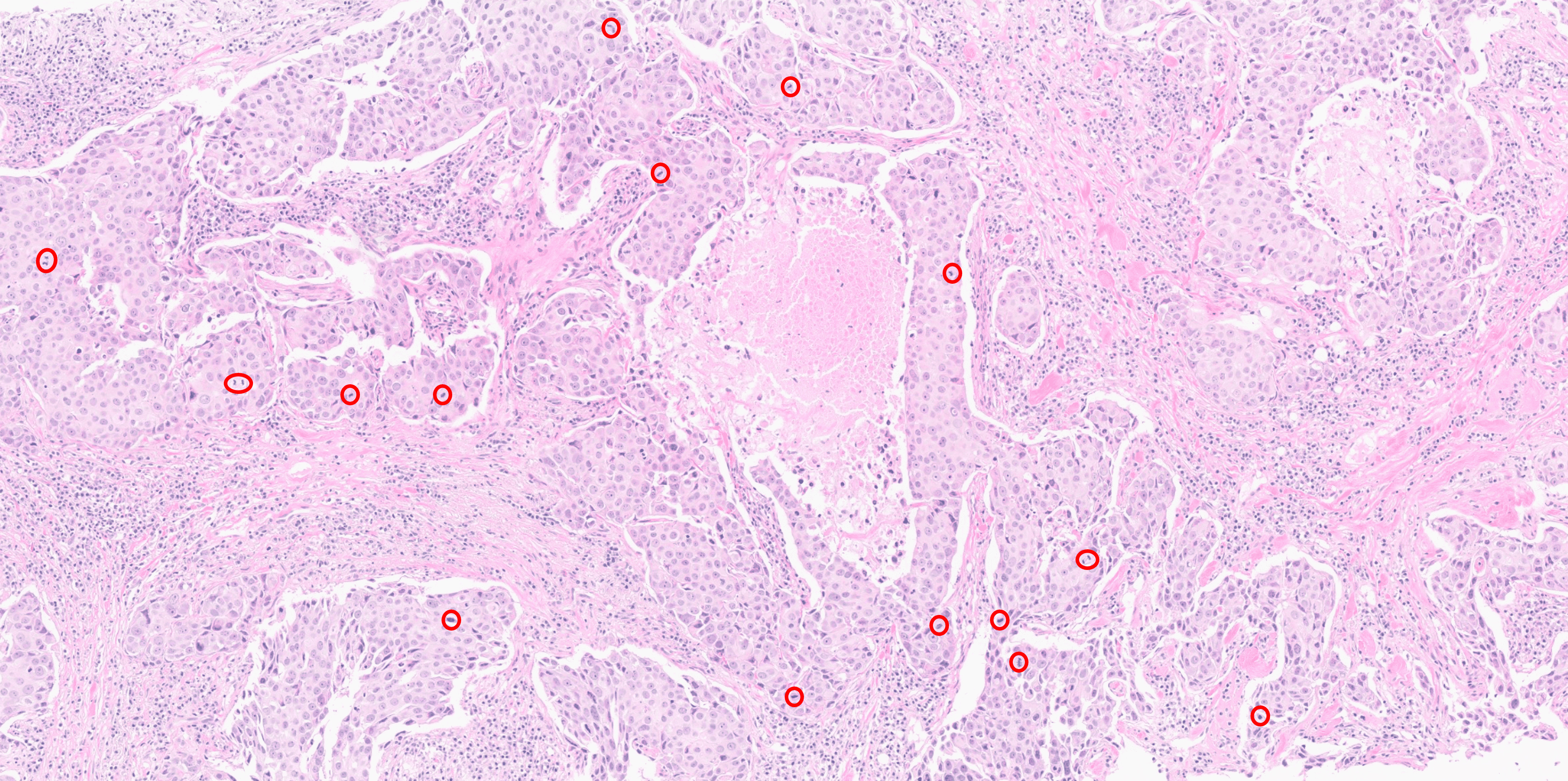} 
    \rule{0pt}{15pt} 
} \\ \hline

\multicolumn{2}{|l|}{\textbf{Ground Truth}} & \textbf{ADCT (0.8141)} & \textbf{ICGI (1.000)} \\ \hline

    % Ground Truth
    \multicolumn{2}{|p{0.45\textwidth}|}{
    Breast, sono-guided core biopsy; \newline
    \tab Invasive carcinoma of no special type, grade III \newline 
    \tab (Tubule formation: 3, Nuclear grade: 3, Mitoses: 2)
    } &

% ADCT
\hl{correctColor}{Breast, sono-guided core biopsy;} \newline
\tab \hl{correctColor}{Invasive carcinoma of no} \newline 
\tab \hl{correctColor}{special type, grade} \hl{incorrectColor}{I} \newline
\tab \hl{correctColor}{(Tubule formation: 3,} \newline
\tab \hl{correctColor}{Nuclear grade:} \hl{incorrectColor}{1,} \newline 
\tab \hl{correctColor}{Mitoses:} \hl{incorrectColor}{1)} &

% ICGI
\hl{correctColor}{Breast, sono-guided core biopsy;} \newline
\tab \hl{correctColor}{Invasive carcinoma of no} \newline 
\tab \hl{correctColor}{special type, grade III} \newline
\tab \hl{correctColor}{(Tubule formation: 3,} 
\tab \hl{correctColor}{Nuclear grade: 3,}
\tab \hl{correctColor}{Mitoses: 2)} \\ \hline

\textbf{ICL\_PathReport (1.000)} & \textbf{IMAGINE\_Lab (0.2525)} & \textbf{IUCompPath (0.2627)} & \textbf{MTS\_REG\_2025 (0.2667)} \\ \hline

% ICL_PathReport
\hl{correctColor}{Breast, sono-guided core biopsy;} \newline
\tab \hl{correctColor}{Invasive carcinoma of no} \newline 
\tab \hl{correctColor}{special type, grade III} \newline
\tab \hl{correctColor}{(Tubule formation: 3,} 
\tab \hl{correctColor}{Nuclear grade: 3,}
\tab \hl{correctColor}{Mitoses: 2)}  &

% IMAGINE_Lab
\hl{incorrectColor}{lung, biopsy;} \newline
\tab \hl{incorrectColor}{adenocarcinoma} &

% IUCompPath
\hl{incorrectColor}{Lung, biopsy;} \newline
\tab \hl{incorrectColor}{Non-small cell carcinoma,} \newline
\tab \hl{incorrectColor}{favor adenocarcinoma} &

% MTS_REG_2025
\hl{incorrectColor}{Lung, biopsy;} \newline
\tab \hl{incorrectColor}{Squamous cell carcinoma} \\ \hline

\textbf{MedInsight-ViseurAI (0.8148)} & \textbf{REG\_Path (0.2627)} & \textbf{TeamTiger@REG2025 (0.2627)} & \textbf{PathoMozhi (0.7060)} \\ \hline

 % MedInsight-ViseurAI
\hl{correctColor}{Breast, sono-guided core biopsy;} \newline
\tab \hl{correctColor}{Invasive carcinoma of no} \newline 
\tab \hl{correctColor}{special type, grade} \hl{incorrectColor}{II} \newline
\tab \hl{correctColor}{(Tubule formation: 3,} \newline
\tab \hl{correctColor}{Nuclear grade:} \hl{incorrectColor}{2,} \newline
\tab \hl{correctColor}{Mitoses:} \hl{incorrectColor}{1)} &

% REG_Path
\hl{incorrectColor}{Lung, biopsy;} \newline
\tab \hl{incorrectColor}{Non-small cell carcinoma,} \newline
\tab \hl{incorrectColor}{favor adenocarcinoma} &

% TeamTiger@REG2025
\hl{incorrectColor}{Lung, biopsy;} \newline
\tab \hl{incorrectColor}{Non-small cell carcinoma,} \newline
\tab \hl{incorrectColor}{favor adenocarcinoma} &

% PathoMozhi
\hl{incorrectColor}{Structured Summary:} \newline
\hl{incorrectColor}{Organ -} \hl{correctColor}{Breast,} \newline
\hl{incorrectColor}{Diagnosis -} \hl{correctColor}{Invasive carcinoma} \newline
\hl{correctColor}{of no special type, grade III} \newline 
\hl{correctColor}{(Tubule formation: 3,} \newline
\hl{correctColor}{Nuclear grade: 3 and} \newline
\hl{correctColor}{Mitoses: 2)} \\ \hline

\end{tabularx}

\vspace{5mm}

\end{table*}

%%%%%%%%%%%%%%%%%%%%%%%%%%%%%%% END %%%%%%%%%%%%%%%%%%%%%%%%%%%%%%%



\end{document}